\documentclass[journal]{IEEEtran}
\usepackage{amssymb}
\usepackage{amsmath,graphicx}
\usepackage{mathrsfs}
\usepackage{subfigure,epstopdf}
\usepackage{algorithm2e}
\RestyleAlgo{ruled}
\usepackage{extarrows}
\usepackage[compress]{cite}

\newtheorem{Problem}{\bf Problem}
\newtheorem{Definition}{\bf Definition}
\newtheorem{Lemma}{\bf Lemma}
\newtheorem{Theorem}{\bf Theorem}
\newtheorem{Remark}{\bf Remark}
\newenvironment{Proof}{\noindent{\em Proof:\/}}{\hfill $\blacksquare$\par}
\newtheorem{Proposition}{\bf Proposition}

\newcommand{\R}{\mathbb{R}}
\newcommand{\RR}{\mathcal{R}}
\newcommand{\n}{\mathscr{N}}
\newcommand{\D}{\mathcal{D}}
\newcommand{\G}{\mathcal{G}}
\newcommand{\I}{\mathcal{I}}
\newcommand{\N}{\mathcal{N}}
\newcommand{\M}{\mathcal{M}}
\newcommand{\U}{\mathcal{U}}
\newcommand{\C}{\mathcal{C}}
\newcommand{\1}{\mathbf{1}}
\newcommand{\A}{\mathcal{A}}
\newcommand{\B}{\mathcal{B}}

\renewcommand{\L}{\mathcal{L}}

\renewcommand{\P}{\mathcal{P}}

\title{\LARGE \bf
Coordinated Defense Allocation in Reach-Avoid Scenarios with Efficient Online Optimization
}
\author{Junwei Liu, Zikai Ouyang, Jiahui Yang, Hua Chen, Haibo Lu and Wei Zhang
\thanks{Junwei Liu and Hua Chen are with the School of System Design and Intelligent Manufacturing, Southern University of Science and Technology, Shenzhen 518055, China {\tt\small \{liujw,chenh6\}@sustech.edu.cn}
}
\thanks{Zikai Ouyang and Wei Zhang are with the School of System Design and Intelligent Manufacturing, Southern University of Science and Technology, Shenzhen 518055, China, and also with the Peng Cheng Laboratory, Shenzhen 518000, China {\tt\small ouyang2022@mail.sustech.edu.cn}, {\tt\small zhangw3@sustech.edu.cn}}
\thanks{Jiahui Yang is with the Department of Mechanical and Energy Engineering, Southern University of Science and Technology, Shenzhen 518055, China {\tt\small 11911213@mail.sustech.edu.cn}}
\thanks{Haibo Lu is with the Peng Cheng Laboratory, Shenzhen 518000, China {\tt\small luhb@pcl.ac.cn}}
}
\begin{document}
\maketitle

\begin{abstract}
In this paper, we present a dual-layer online optimization strategy for defender robots operating in multiplayer reach-avoid games within general convex environments. Our goal is to intercept as many attacker robots as possible without prior knowledge of their strategies. To balance optimality and efficiency, our approach alternates between coordinating defender coalitions against individual attackers and allocating coalitions to attackers based on predicted single-attack coordination outcomes. We develop an online convex programming technique for single-attack defense coordination, which not only allows adaptability to joint states but also identifies the maximal region of initial joint states that guarantees successful attack interception. Our defense allocation algorithm utilizes a hierarchical iterative method to approximate integer linear programs with a monotonicity constraint, reducing computational burden while ensuring enhanced defense performance over time. Extensive simulations conducted in 2D and 3D environments validate the efficacy of our approach in comparison to state-of-the-art approaches, and show its applicability in wheeled mobile robots and quadcopters.
\end{abstract}

\begin{IEEEkeywords}
Multirobot systems, coordination, task allocation, optimization and optimal control.
\end{IEEEkeywords}

\section{Introduction}

Effective task execution in many multirobot systems necessitates both cooperation among team members and competition against adversarial agents. Multiplayer adversarial games, characterized by conflicting objectives, serve as an ideal platform to investigate the delicate interplay between cooperation and competition. Techniques derived from multiplayer adversarial games have been applied to a wide range of real-world applications, including search and rescue \cite{vidal2002probabilistic}, autonomous vehicles \cite{bansal2017hamilton}, security \cite{robin2016multi}, among others. Nevertheless, solving multiplayer adversarial games optimally and efficiently remains a formidable challenge due to the high dimensionality of the solution space, the need for fast decision-making, and the uncertainty induced by unknown opposing strategies. Although deep multi-agent reinforcement learning \cite{lowe2017multi,rashid2020monotonic,zhang2021multi} offers powerful tools, these approaches currently encounter difficulties in managing non-stationarity during training, restricting their applicability in real-world adversarial scenarios.

This paper delves into a class of multiplayer adversarial games, known as multiplayer reach-avoid games, wherein two teams strive to attack or defend a target set without awareness of their opponent's strategies. The reach-avoid game exhibits close connections with other types of adversarial games, such as pursuit-evasion \cite{chung2011search,bakolas2012relay,alexopoulos2015cooperative,de2021decentralized,zhang2022game}, capture-the-flag \cite{huang2011differential,garcia2018capture,jaderberg2019human}, and perimeter defense \cite{shishika2018local,guerrero2020perimeter,velhal2022decentralized}, among others. In a typical multiplayer reach-avoid game, the attacker team tries to reach the target set with as many robots as possible while avoiding capture, whereas the defender team seeks to prevent such incursions by capturing attackers outside the target set. In this work, we focus on the defensive perspective of multiplayer reach-avoid games within convex environments, proposing an effective strategy for the defender team to minimize the number of successful attackers entering the target set, irrespective of the attacker team's decisions. 

A conventional approach to devising defense strategies for multirobot systems employs the reach-avoid differential game framework. This framework casts the problem as a zero-sum game with two opposing players, aiming to determine whether the joint state can be guided to a target set without entering an avoid set \cite{margellos2011hamilton}, \cite{fisac2015reach}. Unlike the traditional Hamilton-Jacobi reachability problem \cite{mitchell2005time}, reach-avoid differential games introduce additional challenges due to the state constraints imposed by the avoid set. Within the reach-avoid differential game framework, defense strategies can be generated by solving the Hamilton-Jacobi-Isaacs (HJI) equation or a related variant offline \cite{bacsar1998dynamic}. However, the HJI method and its numerical approximation \cite{landry2018reach}, \cite{zhou2018efficient} struggle to decouple the inherent computational complexities, leading to poor scalability in large-scale multirobot systems.

To overcome these limitations, our work focuses on the decomposition method, which partitions the problem into two distinct yet interrelated subproblems: single-attack defense coordination and defense allocation. The former involves developing a coordination strategy for a coalition of defenders to prevent an individual attacker from reaching the target set. If successful, we say the defender coalition wins against the attacker. The latter, on the other hand, seeks to optimally assign defender coalitions to attackers by evaluating the predicted outcomes of selected single-attack coordination scenarios. A defense strategy can be obtained by solving these subproblems in an alternating manner. In contrast to the HJI method, the decomposition method offers a more tractable solution by leveraging problem structure and improving computational efficiency. We introduce online optimization techniques for both subproblems, ensuring the real-time adaptability and scalability of our approach for dynamic interactions.

As the foundation of our framework, we present an online convex programming-based defense coordination algorithm for any defender coalition against a single attacker. Our algorithm adapts to every joint state of the defender coalition and attacker, and identifies the maximum defense-winning region of initial joint states, regardless of any admissible strategy employed by the attacker. The underlying principle behind our approach leverages the concept of the safe-reachable set to devise objective functions. The safe-reachable set represents the collection of positions an attacker can reach without being captured by any defender given the current joint state. We then translate the single-attack defense coordination problem into an online robust maximization problem, where the robustness accounts for the worst-case scenario the attacker could employ. Solving this online robust maximization problem yields the corresponding defense coordination strategy. 

Building upon the single-attack defense coordination results, we develop an allocation algorithm that ensures a monotonic improvement in defense performance. This algorithm incorporates predicted outcomes from single-attack defense coordination scenarios to formulate the problem as integer linear programs (ILPs) that rely on the joint state. We employ a heuristic method to tackle the ILPs and incorporate a monotonicity constraint. This constraint guarantees that the expected number of attackers unable to reach the target set is monotonically non-decreasing over time, facilitating a sustained improvement in defense performance as the game progresses. At the core of our algorithm is a hierarchical iterative method that approximately solves the ILPs, with each hierarchy defined by the concepts of active defense set and irreducibility to balance optimality and efficiency. By aggregating the suboptimal solutions obtained across all hierarchies, we derive a solution that substantially diminishes the number of selected pairs of defender coalitions and attackers within the ILPs, while preserving a certain degree of optimality. 

\subsection{Related Work}

In this subsection, we provide an overview of the current state-of-the-art in single-attack defense coordination and defense allocation, highlighting the prevailing issues and unresolved challenges within each domain.

One approach to single-attack defense coordination involves constructing the HJI solution analytically \cite{garcia2020optimal}, \cite{fu2023justification}. Garcia et al. \cite{garcia2020optimal} present the derivation of the optimal defense strategy for two defenders against a single attacker in a 3D scenario with a point target set and zero capture radii. Another approach employs mixed pursuit-defense strategies, where some defenders are designated as pursuers focusing on directly capturing the attacker \cite{pan2012pursuit}, \cite{deng2020multi}. In \cite{deng2020multi}, it is shown that the attacker is guaranteed capture outside a circular target set within a bounded 2D convex domain for a set of initial joint states. This is achieved by assigning one defender to perform defense, while the remaining defenders implement the area-minimization strategy based on the attacker's Voronoi cell, as developed in \cite{huang2011guaranteed,pierson2016intercepting}. 

Recently, a series of geometric-based works by Yan et al. has been proposed for deriving defense coordination strategies \cite{yan2018reach,yan2019task,yan2020guarding,yan2022matching}. Central to these approaches is the set that characterizes the attacker's locations which can be safely reached through admissible strategies. This concept was first introduced in \cite{isaacs1999differential} as the ``safe region" and later reformulated for cooperative pursuit in \cite{zhou2016cooperative} as the ``safe-reachable set". We adopt the term safe-reachable set to emphasize the combination of safety and reachability. In cases where the capture radii are zero and the maximum speeds of defenders are all greater than that of the attacker, the safe-reachable set reduces to the Apollonius circle/sphere. Building on the Apollonius circle (resp. sphere), an analytical defense strategy is given for a set of initial joint states involving two defenders in a 2D (or 3D) convex domain with a line segment (resp. plane) shaped target set \cite{yan2018reach} (resp. \cite{yan2020guarding}), ensuring the attacker cannot enter the target set. This defense coordination technique has been extended to accommodate an arbitrary number of defenders in both specific 2D and 3D scenarios \cite{yan2019task}, \cite{yan2022matching}. Notably, the approach presented in \cite{yan2022matching} is capable of handling cases involving non-zero capture radii.

Despite the progress made in single-attack defense coordination strategies, the existing approaches have mainly concentrated on specific cases with particular target sets or scenarios with zero capture radii. As a result, a comprehensive solution for general convex target sets in convex environments and non-zero capture radii remains an open challenge. Moreover, the defense strategies developed thus far are applicable only to a portion of the joint space, which may not adequately address real-world situations in which the attacker deviates from the optimal strategy. These limitations underscore the need for more adaptable defense coordination methods that can be applied to a broader range of scenarios.

In the context of defense allocation, the problem can be reformulated into joint state-dependent ILPs. Distinct from classical multirobot allocation problems \cite{gerkey2004formal,korsah2013comprehensive}, the objective function here hinges on the estimated outcomes involving all defenders against a single attacker, which generally suffers from computational burdens when attempting a direct solution. In \cite{chen2016multiplayer}, the authors rely on the HJI method to exclusively examine pairwise interactions between defenders and attackers. While this approach enables polynomial-time solutions through the Ford-Fulkerson algorithm \cite{ford1956maximal} or its improvement, the Hopcroft-Karp algorithm \cite{hopcroft1973n}, it oversimplifies the problem by reducing the level of cooperation among defenders. The study presented in \cite{garcia2020multiple} concentrates on the assignment of performance improvement under the assumption that all attackers can be captured, which restricts the applicability of their approach. By the results of the geometric-based approach to single-attack defense coordination, the ILP in \cite{yan2019task} is resolved exclusively at the initial time, with the resulting assignment maintained throughout the entire process. This approach can guarantee the capture of a certain number of attackers, whereas it lacks robustness in response to the dynamic interactions among agents. To overcome this limitation, the approach introduced in \cite{yan2019task} is further refined in \cite{yan2022matching} by approximately solving the ILPs with a polynomial-time method. Nevertheless, this solution entails employing all possible combinations of single-attack defense coordination, which can be time-consuming in practical applications.

From the preceding discussion, it is evident that existing defense allocation methods fail to tackle the time-consuming issue associated with identifying the estimated outcomes of selected pairs of defender coalitions and individual attackers, particularly when the estimated outcomes cannot be readily provided offline. Moreover, when the optimal solution of the ILP remains unattainable due to efficiency constraints, current approaches fall short in offering guidance to enhance cooperative defense performance. Consequently, there is a demand for innovative solutions on defense allocation that balance optimality and computational efficiency.

\subsection{Contributions}

Our contributions are summarized as fourfold:

\begin{enumerate}
    \item Our defense strategy extends the applicability over previous methods to convex domains that contain a general convex target set in both 2D and 3D spaces, and it accommodates scenarios involving defenders with nonzero capture radii.
    \item Our single-attack defense coordination algorithm, which is implemented through online convex programming, is designed to yield the maximum defense-winning region. The algorithm is applicable to all joint states, rather than being confined to a limited portion of the joint space as seen in existing strategies. This allows for a higher chance of winning when facing non-optimal attack strategies.
    \item We propose a defense allocation algorithm that effectively reduces the computational load associated with determining the predicted outcomes of selected defender coalition-attacker pairs. This efficiency is a major advancement, as existing methods often struggle with time-consuming computations that hinder their practical use. Moreover, our algorithm is designed to incrementally improve defense performance, ensuring continual enhancement over time. This is particularly valuable in situations where the optimal allocation is unattainable due to efficiency constraints.
    \item We provide a comprehensive evaluation of our algorithm through both simulations and real-world physics engine validations in 2D and 3D scenarios. This rigorous testing showcases the superiority of our algorithm in comparison to existing state-of-the-art approaches, demonstrating improvements in both optimality and efficiency. 
\end{enumerate}

\subsection{Organization and Notation}

The remainder of this paper is organized as follows. In Section \ref{sec:2}, we formulate the problem of cooperative defense in multiplayer reach-avoid games and decompose it into two subproblems: single-attack defense coordination and defense allocation. In Section \ref{sec:3}, we present a defense strategy for a coalition against a single attacker. Section \ref{sec:4} reformulates the defense allocation problem as an ILP and presents a heuristic method to approximate the solution. Finally, Section \ref{sec:5} presents the simulation results that benchmark the effectiveness of our approach.

We use the following notation throughout the paper. Let $\R^n$ be the $n$-dimensional Euclidean space, and $||\cdot||$ be the Euclidean norm. Given a vector $v$, the normalizer $\n(v)$ is defined as $\n(v)=\frac{v}{||v||}$ if $v\neq 0$ and $0$ otherwise. In the scalar case, $\n$ is equivalent to the sign function. The indicator function of a set $S$ is denoted by $\1_{S}(\cdot)$ and defined as $\1_{S}(x)=1$ if $x\in S$ and $0$ otherwise. The cardinality of a set $S$ is denoted by $|S|$. Given a set of vectors $v_1,\ldots,v_n$, $(v_1,\ldots,v_n)$ is used to denote the column vector consisting of the entries of $v_i$ for $i=1,\ldots,n$ in sequence. The Cartesian product of $N$ copies of a set $S$ is denoted by $S^N$.

\section{Problem Formulation}\label{sec:2}

\begin{figure}[tp!]
	\centering
	\includegraphics[width=0.9\linewidth]{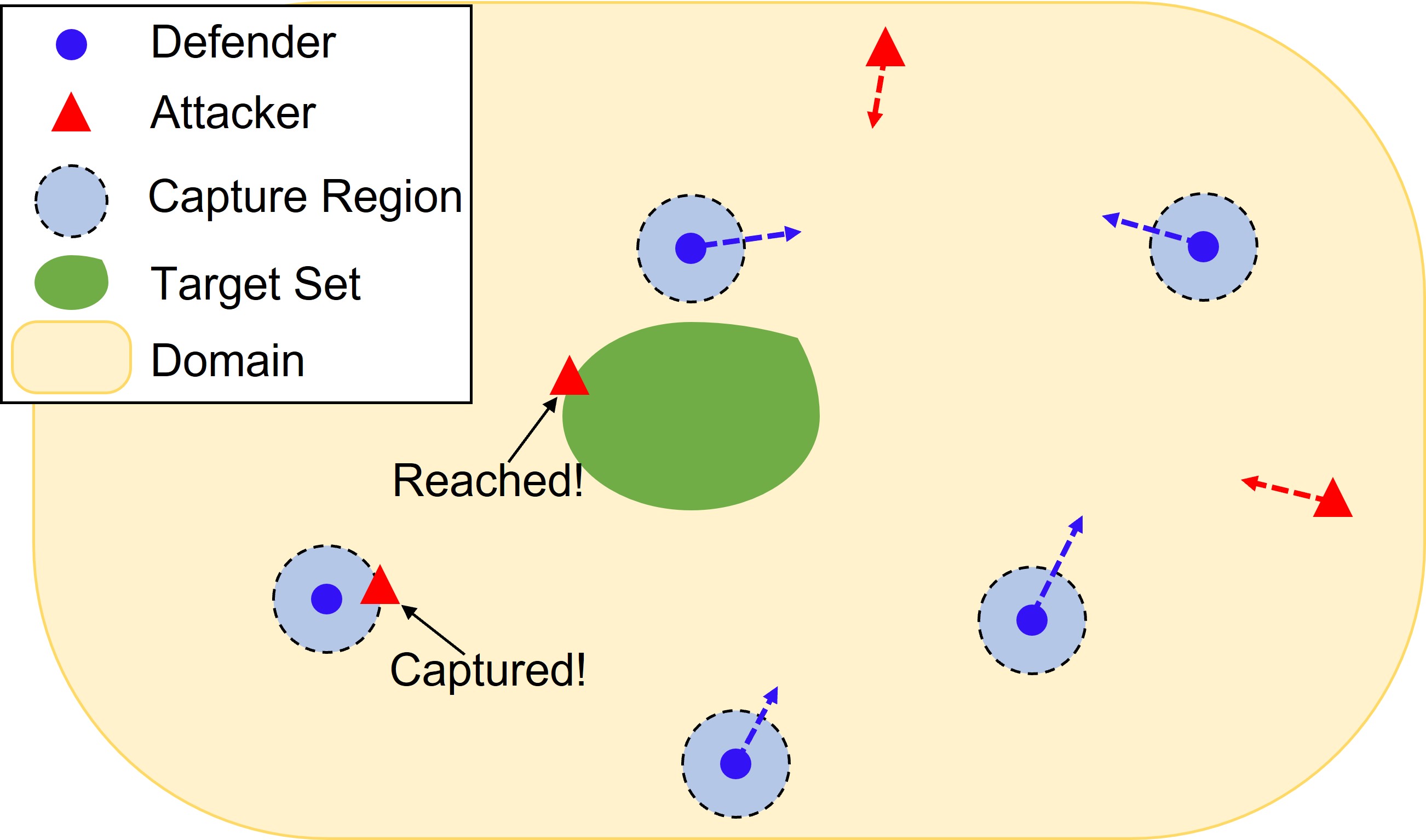}
	\caption{\footnotesize {A 2D instance of a multiplayer reach-avoid game.}}
	\label{fig:radg}
\end{figure}

Consider a multiplayer reach-avoid game involving a team of defender robots (called \textit{defenders}) competing against a team of attacker robots (called \textit{attackers}) within a confined \textit{domain} containing a designated \textit{target set}. The primary objective of the attacker team is to deploy as many robots as possible to reach the target set while avoiding capture by any defender, whereas the defender team strives to prevent the attackers from entering the target set. An illustrative depiction of this game is provided in Fig. \ref{fig:radg}. This paper primarily focuses on the defense perspective, wherein we aim to develop a strategy for the defending team to minimize the total number of attackers that successfully reach the target set, without prior knowledge of the attackers' strategy.

We study scenarios where the domain and target set, denoted as $\D$ and $\G$ respectively, are bounded, closed, and convex. Concretely, we assume that $\D$ and $\G$ are defined by the zero sublevel sets of smooth convex functions $d_k$ and $g_l$, where $k$ and $j$ belong to the index sets $\I_D$ and $\I_G$, respectively. Thus, these sets can be expressed as follows:
\begin{align}
    \D=&\{q\in \R^{n}\mid \max_{k\in \I_D}d_k(q)\leq 0\}\label{con:d}\\
    \G=&\{q\in \R^{n}\mid \max_{l\in \I_G}g_l(q)\leq 0\}.\label{con:t}
\end{align}
In this context, we model the robots as particles, with defenders indexed by $\N=\{1,\ldots,N\}$ and attackers indexed by $\M=\{1,\ldots,M\}$. Besides, the robots are assumed to move within the domain $\D$ with single-integrator dynamics:
\begin{align}
    \dot{p}^d_i=&u^d_i,~\forall i\in \N\label{dyn:d}\\
	\dot{p}^a_j=&u^a_j,~\forall j\in \M\label{dyn:a}
\end{align}
where $p^d_i$ and $u^d_i$ (resp. $p^a_j$ and $u^a_j$) represent the position and velocity of defender $i$ (resp. attacker $j$). 

We define $x=(p^d_1,\ldots,p^d_N,p^a_1,\ldots,p^a_M)$ as the \textit{joint state} of the system \eqref{dyn:d}-\eqref{dyn:a}. Thus, the joint state satisfies the constraint:
\begin{equation}\label{con:state}
x\in \D^{N+M}.
\end{equation}
Additionally, we define the \textit{control inputs} of defender $i$ and attacker $j$ as $u^d_i$ and $u^a_j$, respectively. These control inputs are subject to the following constraints:
\begin{align}
    u^d_i\in \U^d_i=& \{v^d_i\in \R^n\mid ||v^d_i||\leq v^d_{i,\max}\},~\forall i\in \N\label{con:d_input}\\
	u^a_j\in \U^a_j=& \{v^a_j\in \R^n\mid ||v^a_j||\leq v^a_{j,\max}\},~\forall j\in \M.\label{con:a_input}
\end{align}
Here, $v^d_{i,\max}$ and $v^a_{j,\max}$ are the maximum speeds of defender $i$ and attacker $j$, respectively. To prevent attackers from easily outrunning defenders, we impose a constraint on the maximum speed ratio, requiring that defenders possess equal or greater speed capabilities than attackers:
\begin{equation}\label{con:ratio}
\gamma_{ij}=\frac{v^d_{i,\max}}{v^a_{j,\max}}\geq1,~\forall (i,j)\in \N\times\M
\end{equation}
where $\gamma_{ij}$ denotes the ratio of defender $i$'s maximum speed relative to that of attacker $j$.

In the multiplayer reach-avoid game, interactions between defenders and attackers are determined by their relative positions. Specifically, attacker $j$ is said to be \textit{captured} by defender $i$ if the distance between their positions is less than a positive constant $r_i$:
\begin{equation}\label{con:capture}
    ||p^a_j-p^d_i||<r_i
\end{equation}
where $r_i$ represents the $i$th \textit{capture radius}. This leads to the \textit{capture region} of defender $i$ defined as $\RR_i=\{q\in \D\mid ||q-p^d_i||<r_i\}$. 

To simplify our analysis, we assume that attackers that have been captured or have reached the target set become stationary, while defenders can proceed with their tasks after capturing an attacker. This assumption allows us to focus on \textit{active} attackers, which are defined as those attackers who have neither been captured nor reached the target set. The active attackers at the joint state $x$ are indexed by $\M_a(x)=\{j\in \M\mid \rho_j(x_j)=1\}$, where $x_j=(p^d_1,\ldots,p^d_N,p^a_j)$ and $\rho_j$ is defined as:
\begin{equation}\label{rho_j}
	\rho_j(x_j)=
	\begin{cases}
	    1, & \text{attacker}~j~\text{is active} \\
        0, & \text{otherwise}.
	\end{cases}
\end{equation}
Keeping this in mind, we can modify the input constraint for attacker $j$ as follows:
\begin{equation}\label{con:a_input'}
    u^a_j\in \tilde{\U}^a_j(x_j)=
    \begin{cases}
	    \{0\},&\rho_j(x_j)=0\\
	    \U^a_j,&\rho_j(x_j)=1,
	\end{cases}
~\forall j\in \M.
\end{equation}

In this game, we adopt a state feedback information pattern wherein each team makes decisions according to the current joint state without accessing the control inputs of the opposing team. Under this information pattern, an \textit{admissible} strategy for defender $i$ (resp. attacker $j$) is defined as a mapping from the joint state $x$ to the input constraint set $\U^d_i$ (resp. $\tilde{\U}^a_j(x_j)$) such that $u^d_i=\pi^d_i(x)$ (resp. $u^a_j=\pi^a_j(x)$). We define an admissible defense strategy as the collection of all defenders' admissible strategies, denoted by $\pi^d=(\pi^d_1,\ldots,\pi^d_N)$. Likewise, an admissible attack strategy, denoted by $\pi^a=(\pi^a_1,\ldots,\pi^a_M)$, is defined as the collection of all attackers' admissible strategies.

The outcome of the multiplayer reach-avoid game is evaluated in terms of a payoff function, which quantifies the total number of attackers reaching the target set up to the terminal time $t_f$. We consider the terminal time $t_f$ to be free, meaning that the game continues until all attackers have been captured or have reached the target set. Given an initial joint state $x_0$ and a pair of admissible defense and attack strategies $\pi^d$ and $\pi^a$, the payoff function is formally defined as:
\begin{equation}\label{payoff}
    J(x_0,\pi^d,\pi^a)=\sum_{j\in \M}\1_{\G}(p^a_j(t_f)).
\end{equation}
The goal of the defender team is to minimize the value of the payoff function $J$ by choosing appropriate defense strategies, regardless of any admissible attack strategy.

\begin{Problem}[Cooperative Defense in Multiplayer Reach-Avoid Games]\label{pb}
Given a domain $\D$ and a target set $\G$ as defined by \eqref{con:d} and \eqref{con:t}, along with an initial joint state $x_0$ of the system \eqref{dyn:d}-\eqref{dyn:a}, we seek to find an admissible defense strategy $\pi^d$ that minimizes the payoff function $J$ against any admissible attack strategy $\pi^a$. This corresponds to solving the following minimax optimal control problem:
\begin{equation*}
	\begin{split}
        &\min_{\pi^d}\max_{\pi^a}~J(x_0,\pi^d,\pi^a)\\
        &~\mathrm{s.t.}~~x(0)=x_0,~\eqref{dyn:d},~\eqref{dyn:a},~\eqref{con:state},~\eqref{con:d_input},~\eqref{con:ratio},~\eqref{con:a_input'}.
	\end{split}
\end{equation*}
\end{Problem}

Solving Problem \ref{pb} poses significant challenges due to several factors, such as the high dimensionality of the joint state, the discreteness of the payoff function, and the presence of both cooperative and competitive interactions among agents. A widely used approach to tackling these challenges is to employ the differential game formulation, which involves solving the HJI equation or its invariant. However, as the dimension of the joint state increases, the computational complexity of this method becomes intractable, particularly in multi-agent scenarios.

To tackle the aforementioned challenges, we decompose Problem \ref{pb} into two interconnected subproblems:

\begin{itemize}
    \item \textbf{Single-Attack Defense Coordination Problem:}
    Given a coalition of defenders and a single attacker, design an admissible strategy for the coalition to prevent the attacker from reaching the target set, irrespective of any admissible strategy employed by the attacker. In this context, a \textit{coalition} refers to a group of defenders collaboratively working to protect against a specific attacker.
    \item \textbf{Multi-Attack Defense Allocation Problem:} Building upon the solution to the Single-Attack Defense Coordination Problem, design an allocation algorithm that assigns coalitions to active attackers, with the objective of minimizing the payoff function \eqref{payoff}. Recall that, active attackers are those that have not been captured or have not yet reached the target set.
\end{itemize}

The Single-Attack Defense Coordination Problem is solved in a continuous-time manner, wherein the outcome of the multiplayer reach-avoid game is binary: the attacker either succeeds in reaching the target set or is captured outside the target set. We say that the coalition \textit{wins} if the attacker can never enter the target set; otherwise, the coalition is considered \textit{defeated}. Consequently, the problem of single-attack defense coordination is to find a winning strategy for the coalition against the attacker. Conversely, the Multi-Attack Defense Allocation Problem is handled in a discrete-time framework, where each time instant corresponds to an implementation of the allocation. We refer to these time instants as \textit{allocation times}. At each allocation time, the allocation algorithm reevaluates the assignment of active attackers to coalitions based on the current joint state. 

Integrating the solutions of these two subproblems results in a holistic solution to Problem \ref{pb}. Given the current joint state, we initially identify the set of active attackers $\M_a$ by checking their relative distances to the defenders up to the last time instant. Subsequently, the solution to the Multi-Attack Defense Allocation Problem dictates the assignment of active attackers to coalitions. For each specified coalition-attacker pair, we then apply the solution to the Single-Attack Defense Coordination Problem to compute the control input for each defender in the corresponding coalition. This dual-layer scheme continues to update in real-time, effectively decomposing the high-dimensional, complex minimax optimal control problem into smaller, more tractable problems. 

\section{Defense Coordination Against A Single Attacker}\label{sec:3}

In this section, we present a defense strategy for a coalition in multiplayer reach-avoid games involving a single attacker. Our approach begins with a formal definition of the safe-reachable set and a derivation of its sublevel set representation. Leveraging the minimum squared distance between the safe-reachable set and the target set, we construct an objective function that transcribes the single-attack defense coordination problem into its online robust maximization. The solution to this robust maximization leads to an online convex programming-based defense strategy, which guarantees the coalition's victory for a specific region of initial joint states. To increase the chances of winning against non-optimal attack strategies, we further propose a dual-mode switching defense strategy to fit with all initial joint states.

\subsection{Safe-Reachable Set}

In the multiplayer reach-avoid game with a single attacker, the binary outcome of an initial joint state is determined by the reach-avoid set \cite{margellos2011hamilton}. This set comprises joint states from which the attacker can reach the target set without being captured by any of the defenders. By solving an HJI equation numerically, the reach-avoid set can be represented as a sublevel set of the HJI solution \cite{mitchell2007toolbox}. However, the backward computation associated with this method suffers from the curse of dimensionality, hindering its application to multi-agent systems \cite{zhou2018efficient}. In contrast, the safe-reachable set seeks to collect the attacker's reachable locations that remain capture-free given the current joint state \cite{zhou2016cooperative}. This set can be computed using forward computation, which only concerns the attacker's position state and does not involve solving for the target set reachability, making it more computationally efficient. We will demonstrate that under our problem settings, the binary outcome of an initial joint state can also be characterized by the safe-reachable set.

A formal definition of the safe-reachable set for a coalition-attacker pair is given below. Consider a coalition $\C$ and an attacker $j$ with dynamics described by \eqref{dyn:d} and \eqref{dyn:a}, respectively, and denote their joint state as $x^\C_j=(p^d_i,i\in \C;p^a_j)$. In light of \eqref{con:capture}, the capture-free condition for attacker $j$ can be expressed as follows:
\begin{equation}\label{con:capture_f}
    \max_{i\in \C}s_i(p^d_i,p^a_j)\leq 0
\end{equation}
where $s_i(p^d_i,p^a_j)=r_i-||p^d_i-p^a_j||$. To account for the dependence of agent trajectories on their initial states and strategies, we introduce the notations $\chi^a_j(\cdot;p^a_j(0),\pi^a_j)$ and $\chi^d_i(\cdot;p^d_i(0),\pi^d_i)$, which respectively denote the trajectory of attacker $j$ over time starting from $p^a_j(0)$ using the attack strategy $\pi^a_j$, and the trajectory of defender $i$ over time starting from $p^d_i(0)$ using the defense strategy $\pi^d_i$.

\begin{Definition}[Safe-Reachable Set]\label{def:srs}
Given a joint state $x^\C_j$ of $(\C,j)$, the \textit{safe-reachable set} (SRS) $\Omega^\C_j(x^\C_j)$ for attacker $j$ induced by the coalition $\C$ consists of all positions $q\in \D$ satisfying the following properties:
\begin{enumerate}
    \item (Reachability) There exists a time instant $\tau$ and an admissible attack strategy $\pi^a_j$ over the interval $[0,\tau]$ such that the attacker $j$ can reach position $q$ at time $\tau$:
    \begin{equation*}
    	q=\chi^a_j(\tau;p^a_j,\pi^a_j);
	\end{equation*}
    \item (Safety) For any admissible defense strategy $\pi^d_i$ used by the defenders $i\in \C$, the capture-free condition \eqref{con:capture_f} holds for all $t\in [0,\tau]$:
    \begin{equation*}
    	\max_{i\in \C}s_i(\chi_i^d(t;p^d_i,\pi^d_i),\chi^a_j(t;p^a_j,\pi^a_j))\leq 0,~\forall t\in [0,\tau].
	\end{equation*}
\end{enumerate}
\end{Definition}

The SRS offers a predictive way to evaluate the capture-free condition \eqref{con:capture_f} by incorporating the dynamic and input constraints of both opponents. When the SRS is nonempty, it implies that the attacker is currently situated within the SRS, and the capture-free condition is satisfied for the current joint state, as per Definition \ref{def:srs}. Moreover, the size of the SRS can be used to anticipate the attacker's ability to move freely without compromising safety. A larger SRS indicates that the attacker has more options for safe movement, regardless of any defense strategies that may be employed. 

Similar to the reach-avoid set, the SRS admits a sublevel set representation. Consider a point $q$ inside the SRS that attacker $j$ can reach at time $\tau$. Simultaneously, defender $i$ can arrive at point $q^d_i=p^d_i+v^d_{i,\max}\frac{q-p^d_i}{||q-p^d_i||}\tau$ using the constant strategy $v^d_{i,\max}\frac{q-p^d_i}{||q-p^d_i||}$. According to Definition \ref{def:srs}, $q$ adheres to the capture-free condition $\max_{i\in \C}s_i(q^d_i,q)=\max_{i\in \C}(r_i-\bigl\lvert ||q-p^d_i||-v^d_{i,\max}\tau \bigr\rvert)\leq 0$. On the other hand, since attacker $j$ can safely reach $q$, the time $\tau$ must fulfill $\tau\in [\tau_{\min},\frac{||q-p^d_i||}{v^{d}_{i,\max}})$, with $\tau_{\min}=\frac{||q-p^a_j||}{v^{a}_{j,\max}}$ representing the minimum time for attacker $j$ to reach $q$. This leads to $\max_{i\in \C}(r_i-\bigl\lvert ||q-p^d_i||-v^d_{i,\max}\tau_{\min} \bigr\rvert)=\max_{i\in \C}(r_i+\gamma_{ij}||q-p^a_j||-||q-p^d_i||)\leq 0$, which can be rewritten as $\max_{i\in \C}c_{ij}(q,x_{ij})\leq 0$ with
\begin{equation}\label{ineq:c_ij}
	c_{ij}(q,x_{ij})=(\gamma_{ij}||q-p^a_j||+r_i)^2-||q-p^d_i||^2
\end{equation}
and $x_{ij}=(p^d_i,p^a_j)$. Moreover, satisfying inequality \eqref{ineq:c_ij} for every $i\in \C$ is sufficient to ensure that $q$ lies within the SRS. 

\begin{Proposition}[Sublevel Set Representation]\label{prop:ssr}
The SRS defined in Definition \ref{def:srs} is closed and convex, taking the form 
\begin{equation}\label{set:srs}
	\Omega^\C_j(x^\C_j)=\bigcap_{i\in \C}\Big\lbrace q\in \D\mid c_{ij}(q,x_{ij})\leq 0 \Big\rbrace
\end{equation}
where $c_{ij}$ is given by equation \eqref{ineq:c_ij}.
\end{Proposition}

\begin{Proof}
We have shown that $\max_{i\in \C}c_{ij}(q,x_{ij})\leq 0$ for each $q\in\Omega^\C_j(x^\C_j)$, and the remaining is detailed in Appendix \ref{proof:ssr}.
\end{Proof}

\begin{figure}[tp!] 
	\centering
	\includegraphics[width=0.48\linewidth]{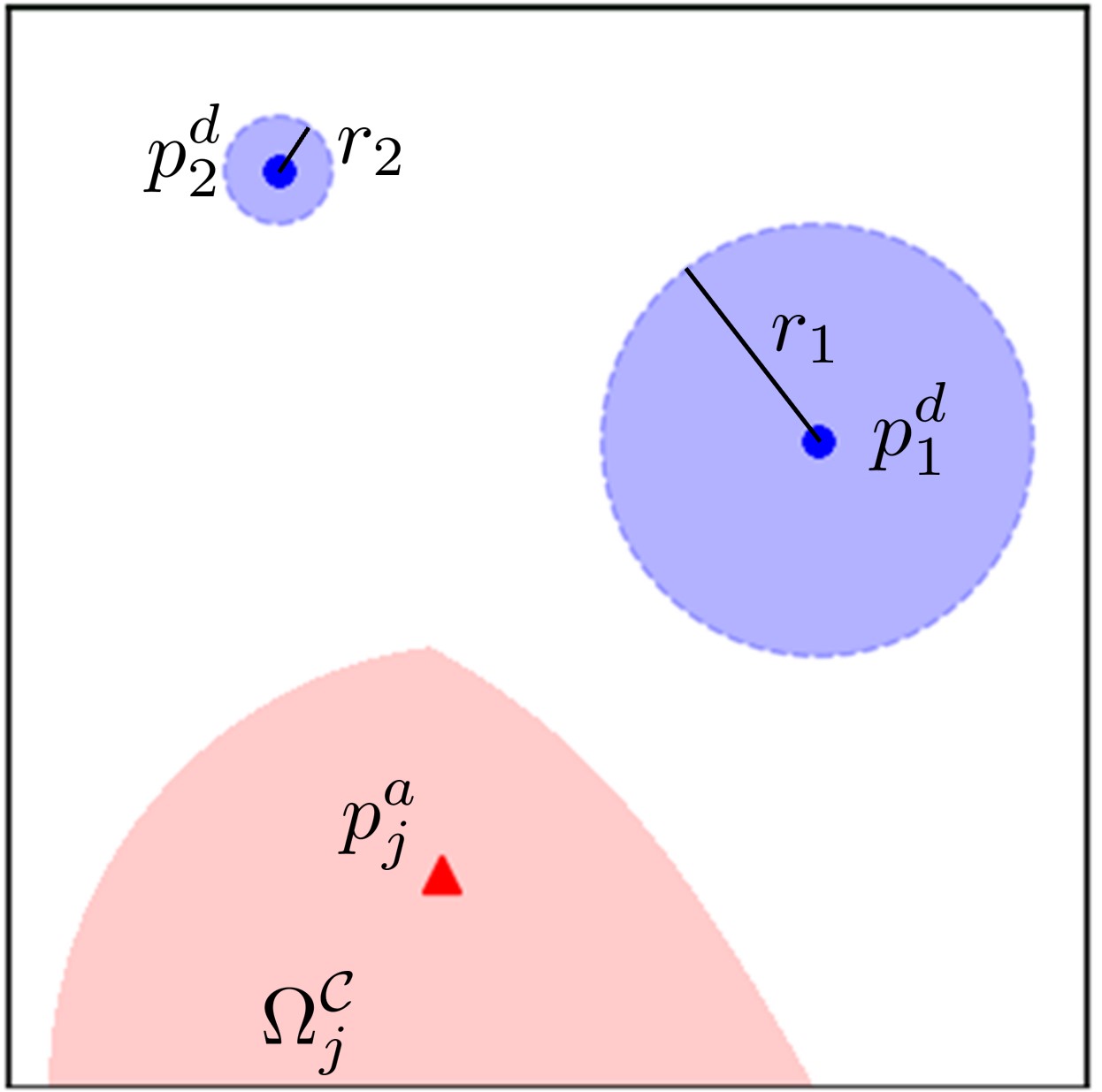}
	\includegraphics[width=0.48\linewidth]{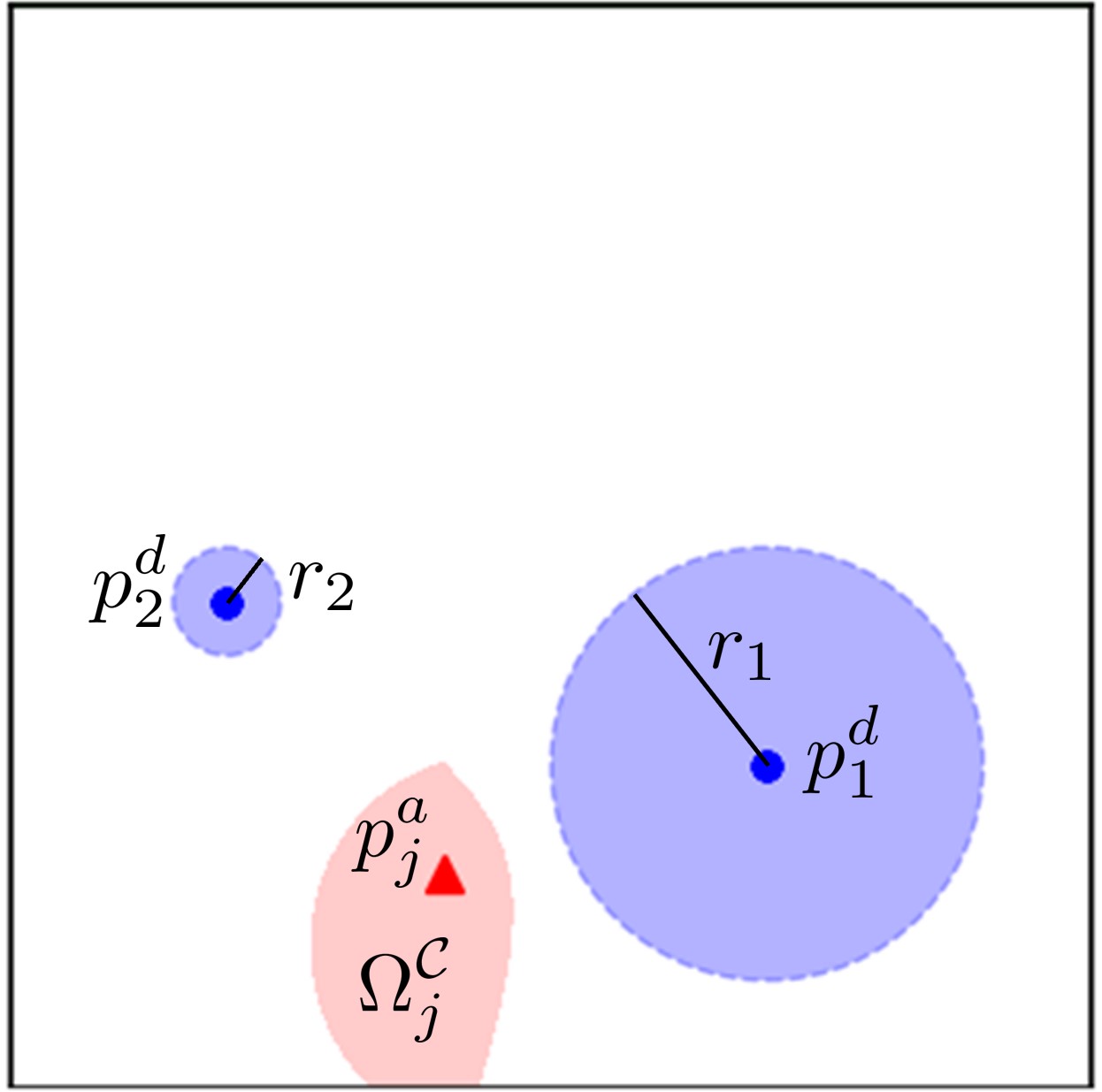}
	\caption{\footnotesize {Visualization of the SRS under different joint states in a 2D domain. The blue disks represent the defenders' capture regions, and the red region illustrates the SRS for a specific attacker. The parameters are chosen as $\gamma_{1j}=1$, $\gamma_{2j}=2$, $r_1=2$ and $r_2=0.5$. As displayed from left to right, the SRS varies with joint states, and its size decreases as the distances between the attacker and the defenders diminish while accounting for the capture radii of the defenders.}
	\label{fig:srs}}
\end{figure}

The SRS serves as a generalization of both the Voronoi diagram and Apollonius diagram by accommodating nonzero capture radii. As evident from equation \eqref{set:srs}, when all the capture radii equal zero, the SRS simplifies to the Voronoi cell if $\gamma_{ij}=1$ and to the Apollonius circle/sphere if $\gamma_{ij}>1$. Unlike the Voronoi diagram and Apollonius diagram, the SRS partitions the space based on the attacker's capacity to move freely without being captured, factoring in the defenders' capture radii, as depicted in Fig. \ref{fig:srs}. This distinctive characteristic of the SRS enables a more realistic representation of the interaction between the attacker and defenders, enhancing defense coordination with unpredictable attacker movements.

\subsection{Single-Attack Objective Function}

Having the SRS concept, we next develop an objective function for the single-attack defense coordination problem. Given that the SRS pertains to the attacker's forward reachability, it is natural to define an objective function that connects the SRS and the target set in terms of their minimum distance. With this in mind, we propose the minimum squared distance between the SRS and the target set as the objective function, which we refer to as the \textit{single-attack objective function}:
\begin{equation}\label{fun:saof}
	\Phi^{\C}_j(x^\C_j)=\min_{\genfrac{}{}{0pt}{2}{q\in \Omega^{\C}_j(x^{\C}_j)}{\tilde{q}\in \G}}||q-\tilde{q}||^2.
\end{equation}
Since both the SRS and the target set are convex and admit sublevel set representations, according to Proposition \ref{prop:ssr} and the assumption on the target set, the value of $\Phi^\C_j$ can be determined by solving the following parametric convex program:
\begin{equation}\label{co:d}
	\begin{split}
        &\min_{(q,\tilde{q})\in \R^{2n}}~~||q-\tilde{q}||^2\\
		&~~~~\mathrm{s.t.}~~c_{ij}(q,x_{ij})\leq 0,~\forall i\in \C\\
		&~~~~~~~~~~d_k(q)\leq 0,~\forall k\in \I_D\\
		&~~~~~~~~~~g_l(\tilde{q})\leq 0,~\forall l\in \I_G
	\end{split}
\end{equation}
where the joint state $x^\C_j$ serves as the parametric vector.

The single-attack objective function provides a quantitative measure for the binary outcome of the single-attacker multi-defender reach-avoid game. It is essential to note that the value of $\Phi^\C_j$ is always nonnegative for a nonempty SRS, and positive if the intersection between the SRS and the target set is empty. This suggests that retaining the positivity of $\Phi^\C_j$ is critical in preventing the attacker from reaching the target set. Consequently, a winning guarantee condition for the coalition can be derived in terms of the single-attack objective function. 

\begin{Lemma}[Defense-Winning Condition]\label{lem:dwc}
For a given coalition-attacker pair $(\C,j)$ at the initial joint state, the coalition $\C$ is guaranteed to win against attacker $j$ if and only if there exists an admissible defense strategy such that, for any admissible attack strategy, the single-attack objective function remains positive throughout the duration:
\begin{equation*}
    \Phi^\C_j(x^\C_j(t))>0,~\forall t\in[0,t_f).
\end{equation*}
\end{Lemma}

\begin{Proof}
For sufficiency, if $\Phi^\C_j(x^\C_j(t))$ remains positive for all $t\in[0,t_f)$, then the SRS cannot intersect the target set during this time interval. Since the attacker's position is always contained within the non-empty SRS, the attacker can never enter the target set. This implies that the coalition is guaranteed to win the game. For necessity, assume for the sake of contradiction that the coalition wins while $\Phi^\C_j(x^\C_j(T))=0$ for some $T\in [0,t_f)$. According to Definition \ref{def:srs}, there is an admissible attack strategy starting from time $T$ that drives attacker $j$ to the target set within a finite time. This, however, contradicts the assumption that the coalition wins.
\end{Proof}

\begin{Remark}[Online Robust Maximization]
Lemma \ref{lem:dwc} confirms that if the strict zero superlevel set of $\Phi^\C_j$ is \textit{robustly positively invariant} \cite{blanchini1999set} under an admissible defense strategy, then the coalition can guarantee victory against any admissible attack strategy. The notion of robustness means that the coalition must maximize the single-attack objective function $\Phi^\C_j$ for the worst-case attack strategy. As a consequence, the optimal defense coordination problem can be converted into an online robust maximization of the single-attack objective function for the coalition at each joint state.
\end{Remark}

\subsection{Defense-Winning Strategy}\label{sec:gbdws}

We proceed to derive a defense-winning strategy for the coalition $\C$ by performing online robust maximization of the single-attack objective function $\Phi^\C_j$ when it has a positive value. In light of Pontryagin's Maximum Principle for differential games (cf. Theorem 8.2 in \cite{nocedal2006numerical}), robustly maximizing $\Phi^\C_j$ for the coalition $\C$ amounts to robustly maximizing its time derivative $\dot{\Phi}^\C_j$ with respect to the inputs $u^d_i$, $i\in \C$. Note that, the time derivative of $\Phi^\C_j$ along the joint state trajectory is given by
\begin{equation}\label{eq:deriv_d}
	\dot{\Phi}^\C_j=\sum_{i\in \C}\frac{\partial \Phi^\C_j}{\partial p^d_i}u^d_i+\frac{\partial \Phi^\C_j}{\partial p^a_j}u^a_j.
\end{equation}
To measure the effect of each agent's input on $\dot{\Phi}^\C_j$, we need to determine the gradients of $\Phi^\C_j$ concerning the positions of all agents. Fortunately, by invoking the Karush–Kuhn–Tucker (KKT) conditions (cf. Theorem 12.1 in \cite{nocedal2006numerical}), a linear correspondence can be made almost everywhere between the gradients of $\Phi^\C_j$ and those of $c_{ij}$, $i\in \C$ given in \eqref{ineq:c_ij}. 

\begin{Proposition}[Gradient Correspondence]\label{prop:gc}
Suppose the convex program \eqref{co:d} admits a unique solution for each joint state $x^\C_j$ within a region $\D'\subset \D$. Then for almost every $x^\C_j\in \D'$, there exist nonnegative constants $\lambda_{ij}^*$, $i\in \C$ such that 
\begin{equation}\label{eq:gc}
	\begin{split}
	    \frac{\partial \Phi^\C_j}{\partial p^d_i
	    }=&\lambda_{ij}^*\frac{\partial c_{ij}}{\partial p^d_i}(\xi^\C_j,x_{ij}),~~\forall i\in \C\\
	    \frac{\partial \Phi^\C_j}{\partial p^a_j
	    }=&\sum_{i\in \C}\lambda_{ij}^*\frac{\partial c_{ij}}{\partial p^a_j}(\xi^\C_j,x_{ij})
	\end{split}
\end{equation}
where $\xi^\C_j$ is the first $n$-dimensional component of the unique solution to the convex program \eqref{co:d}. 
\end{Proposition}

\begin{Proof}
See Appendix \ref{proof:gc}.
\end{Proof}

Proposition \ref{prop:gc} provides a tractable procedure for generating a defense strategy that approximately robustly maximizes $\Phi^\C_j$, assuming that \eqref{eq:deri_c} holds almost everywhere. This procedure entails deriving the gradient of the functions $c_{ij}$, $i\in \C$ for each $q\in \Omega^\C_j(x^\C_j)$ and $q\neq p^a_j$ as follows:
\begin{equation}\label{eq:deri_c}
    \begin{split}
	    \frac{\partial c_{ij}}{\partial p^d_i}(q,x_{ij})=&2(q-p^d_i)^T,~i\in \C\\
	    \frac{\partial c_{ij}}{\partial p^a_j}(q,x_{ij})=&-2\zeta_{ij}(q,p^a_j)(q-p^a_j)^T
	\end{split}
\end{equation}
where $\zeta_{ij}(q,p^a_j)=\gamma_{ij}^2+\frac{r_i\gamma_{ij}}{||q-p^a_j||}$. By substituting \eqref{eq:deri_c} with $q=\xi^\C_j$ into \eqref{eq:gc} and noting that the parameter $\lambda_{ij}^*$ is nonnegative, we can obtain an almost robust optimal defense strategy for defender $i$ in coalition $\C$ given by:
\begin{equation}\label{ds:s}
    \pi^d_i(x^\C_j)=v^d_{i,\max}\n(\xi^\C_j-p^d_i),~i\in \C
\end{equation}
for $x^\C_j\in \D'$ such that the convex program \eqref{co:d} has a unique solution. Moreover, it can be shown that the uniqueness of the solution to \eqref{co:d} is guaranteed as long as $\Phi^\C_j$ remains positive.

\begin{Proposition}[Solution Uniqueness]\label{prop:unique}
For each joint state $x^\C_j$ within the region defined by
\begin{equation*}
    \D^\C_j=\Big\{x^\C_j\in \D^{|\C|+1}\mid\Phi^\C_j(x^\C_j)>0\Big\}
\end{equation*}
the convex program \eqref{co:d} has a unique solution.
\end{Proposition}

\begin{Proof}
See Appendix \ref{proof:unique}.
\end{Proof}

We are now ready to present the main result of this section, which asserts that the almost robust optimal defense strategy given by \eqref{ds:s} guarantees a win for the coalition once the joint state $x^\C_j$ lies within the region $\D^\C_j$.

\begin{Theorem}[Defense-Winning Strategy]\label{thm:dws}
For any initial joint state within the region $\D^\C_j$, the coalition $\C$ is ensured to win against attacker $j$ under the defense strategy \eqref{ds:s}, irrespective of any admissible attack strategies.
\end{Theorem}

\begin{Proof}
In light of Proposition \ref{prop:gc}, inserting \eqref{eq:gc} along with \eqref{eq:deri_c} and \eqref{ds:s} into \eqref{eq:deriv_d} reveals that for almost every $x^\C_j\in \D^\C_j$,
\begin{equation*}
    \begin{split}
	    \dot{\Phi}^\C_j=&\sum_{i\in \C}2\lambda_{ij}^*v^d_{i,\max}(\xi^\C_j-p^d_i)^T\n(\xi^\C_j-p^d_i)\\
	    &-\sum_{i\in \C}2\lambda_{ij}^*\zeta_{ij}(\xi^\C_j,p^a_j)(\xi^\C_j-p^a_j)^Tu^a_j\\
	    \geq&\sum_{i\in \C}2v^d_{i,\max}\lambda_{ij}^*(||\xi^\C_j-p^d_i||-\gamma_{ij}||\xi^\C_j-p^a_j||-r_i)\\
	    \geq& 0
	\end{split}
\end{equation*}
where the first and second inequalities hold due to $||u^a_j||\leq v^{a}_{j,\max}$ and $\xi^\C_j\in \Omega^\C_j(x^\C_j)$, respectively. As such, $\Phi^\C_j(x^\C_j(t))$ is non-decreasing over the time interval $[0,t_f)$. This implies that $\Phi^\C_j(x^\C_j(t))\geq\Phi^\C_j(x^\C_j(0))>0$ for all $t\in [0,t_f)$, which in turn completes the proof, as established by Lemma \ref{lem:dwc}.
\end{Proof}

\begin{Remark}[Maximum Defense-Winning Region]
The defense-winning region $\D^\C_j$ is \textit{maximum} in the sense that $\D^\C_j$ contains all initial joint states for which the coalition $\C$ can assuredly win against attacker $j$. This is because if an initial joint state $x^\C_j(0)$ is not in $\D^\C_j$, then it will lead to $\Phi^\C_j(x^\C_j(0))=0$ according to the definition of $\D^\C_j$. This implies that attacker $j$ can safely reach the target set as stated by Lemma \ref{lem:dwc}. Therefore, the outcome of a single-attack multi-defender reach-avoid game can be fully determined by checking whether the current joint state lies within the defense-winning region $\D^\C_j$.
\end{Remark}

\subsection{Dual-Mode Switching Defense Coordination}

We have identified a region $\D^\C_j$ of joint states within which the almost robust optimal defense strategy \eqref{ds:s} ensures victory for the coalition. This defense strategy relies on the solution of the convex program \eqref{co:d}, which typically does not guarantee solution uniqueness for joint states outside $\D^\C_j$. In real-world scenarios, as the attack strategy is unknown, it would be advantageous for the defense strategy to handle any joint state, not just those within the region $\D^\C_j$. By doing so, the chances of winning for the coalition against non-optimal attack strategies are increased. In the remainder of this section, we focus on utilizing the concept of SRS to extend the defense strategy to cover the entire joint state space.

We start by proposing a novel objective function specifically designed for joint states outside the region $\D^\C_j$. The single-attack objective function $\Phi^\C_j$ is not suitable for this purpose, as it cannot distinguish between joint states that are not in $\D^\C_j$, resulting in zero values for all such joint states. Given that an intelligent attacker tends to reach the target set safely and rapidly, and the intersection between the SRS and the target set is nonempty for $x^\C_j\notin \D^\C_j$, the objective function is defined as the minimum squared distance between the attacker's position and the intersection between the SRS and the target set:
\begin{equation}\label{saof}
	\begin{split}
		\bar{\Phi}^\C_j(x^\C_j)=\min_{q\in \Omega^\C_j(x^\C_j)\cap \G}||q-p^a_j||^2.
	\end{split}
\end{equation}
The convexity of the SRS enables the determination of $\bar{\Phi}^\C_j(x^\C_j)$ through the solution of the parametric convex program:
\begin{equation}\label{co:a}
	\begin{split}
		&\min_{q\in \R^{n}}~~||q-p^a_j||^2\\
		&~~\mathrm{s.t.}~~c_{ij}(q,x_{ij})\leq 0,~\forall i\in \C\\
		&~~~~~~~~g_l(q)\leq 0,~\forall l\in \I_G.
	\end{split}
\end{equation}
The uniqueness of the solution to \eqref{co:a} is guaranteed since the projection of $p^a_j$ onto the convex and closed set $\Omega^\C_j(x^\C_j)\cap\G$ is unique \cite{boyd2004convex}.

\begin{algorithm}[tp!]
\caption{Dual-Mode Switching Defense Coordination}\label{alg:dmsdc}
\KwData{Positions of defenders $p^d_i$, $i\in \C$ and attacker $p^a_j$}
\KwResult{Defender inputs $u^d_i$, $i\in \C$}
\For{each defender $i\in \C$}{
    $\Phi^{\C}_j,\xi^\C_j\gets$ Optimal value and solution of \eqref{co:d} with $(\C,j)$\\
    \uIf{$\Phi^{\C}_j>0$}{
        $u^d_i\gets v^d_{i,\max}\n(\xi^\C_j-p^d_i)$
        }
    \Else{
    $\bar{\xi}^\C_j\gets$ Solution to \eqref{co:a} with $(\C,j)$\\
    $u^d_i\gets v^d_{i,\max}\n(\bar{\xi}^\C_j-p^d_i)$
    }
}
\end{algorithm}

The objective function $\bar{\Phi}^\C_j$ is then employed to derive a defense strategy for which $x^{\C}_j\notin \D^C_j$. Based on the definition of $\bar{\Phi}^\C_j$, the goal of the coalition is to robustly maximize $\bar{\Phi}^\C_j$ over time, which can be accomplished by robustly maximizing the time derivative of $\bar{\Phi}^\C_j$. To achieve this, we need to determine the gradient of $\bar{\Phi}^\C_j$ with respect to the position of each defender in the coalition. Analogous to Proposition \ref{prop:gc}, it can be shown that for almost every $x^\C_j\notin \D^\C_j$, there is a set of nonnegative constants $\mu^*_{ij}$, $i\in \N$, such that
\begin{equation}\label{eq:gc_a}
	\begin{split}
	    \frac{\partial \bar{\Phi}^\C_j}{\partial p^d_i}=&\mu^*_{ij}\frac{\partial c_{ij}}{\partial p^d_i}(\bar{\xi}^\C_j,x_{ij}),~\forall i\in \C\\
	    \frac{\partial \bar{\Phi}^\C_j}{\partial p^a_j}=&\sum_{i\in \C}\mu^*_{ij}\frac{\partial c_{ij}}{\partial p^a_j}(\bar{\xi}^\C_j,x_{ij})-2(\bar{\xi}^\C_j-p^a_j)^T
	\end{split}
\end{equation}
where $\bar{\xi}^\C_j$ is the unique solution to the convex program \eqref{co:a}. The proof of \eqref{eq:gc_a}, owing to its similarity to that of Proposition \ref{prop:gc}, can be found in Appendix \ref{proof:gc}. Considering the almost everywhere holding of \eqref{eq:gc_a}, substituting \eqref{eq:deri_c} with $q=\bar{\xi}^\C_j$ into \eqref{eq:gc_a} yields an almost robust optimal defense strategy for $x^{\C}_j\notin \D^C_j$:
\begin{equation}\label{ds_}
    \pi^d_i(x^\C_j)=v^d_{i,\max}\n(\bar{\xi}^\C_j-p^d_i),~\forall i\in \C.
\end{equation}
This defense strategy, together with the defense strategy \eqref{ds:s} for $x^{\C}_j\in \D^C_j$, constitutes a dual-mode switching defense strategy applicable to the entire joint state space. The proposed algorithm for coordinating the defense against a single attacker is outlined in Algorithm \ref{alg:dmsdc}.

\begin{figure}[tp!]
	\centering
	\includegraphics[width=0.48\linewidth]{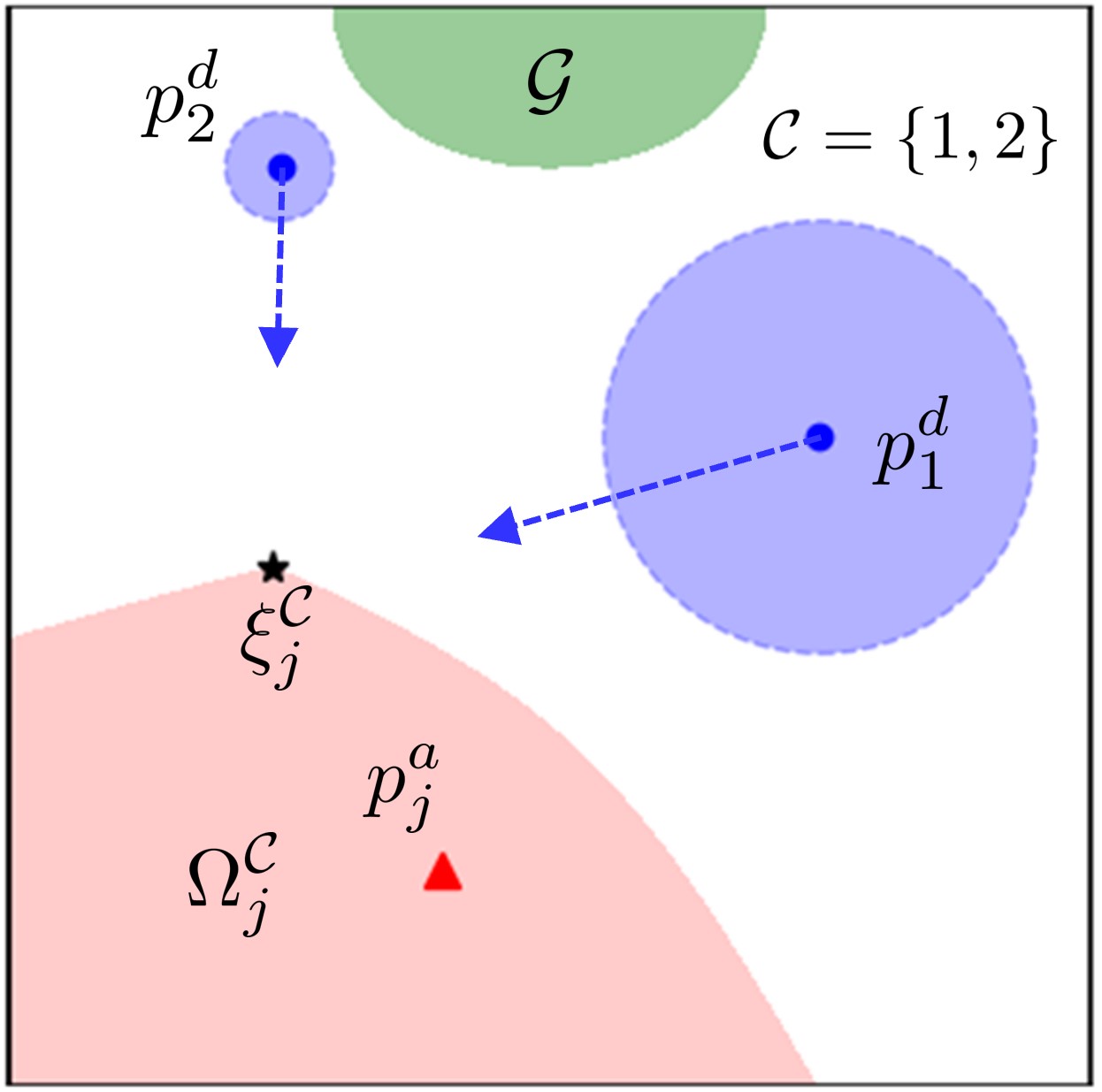}
	\includegraphics[width=0.48\linewidth]{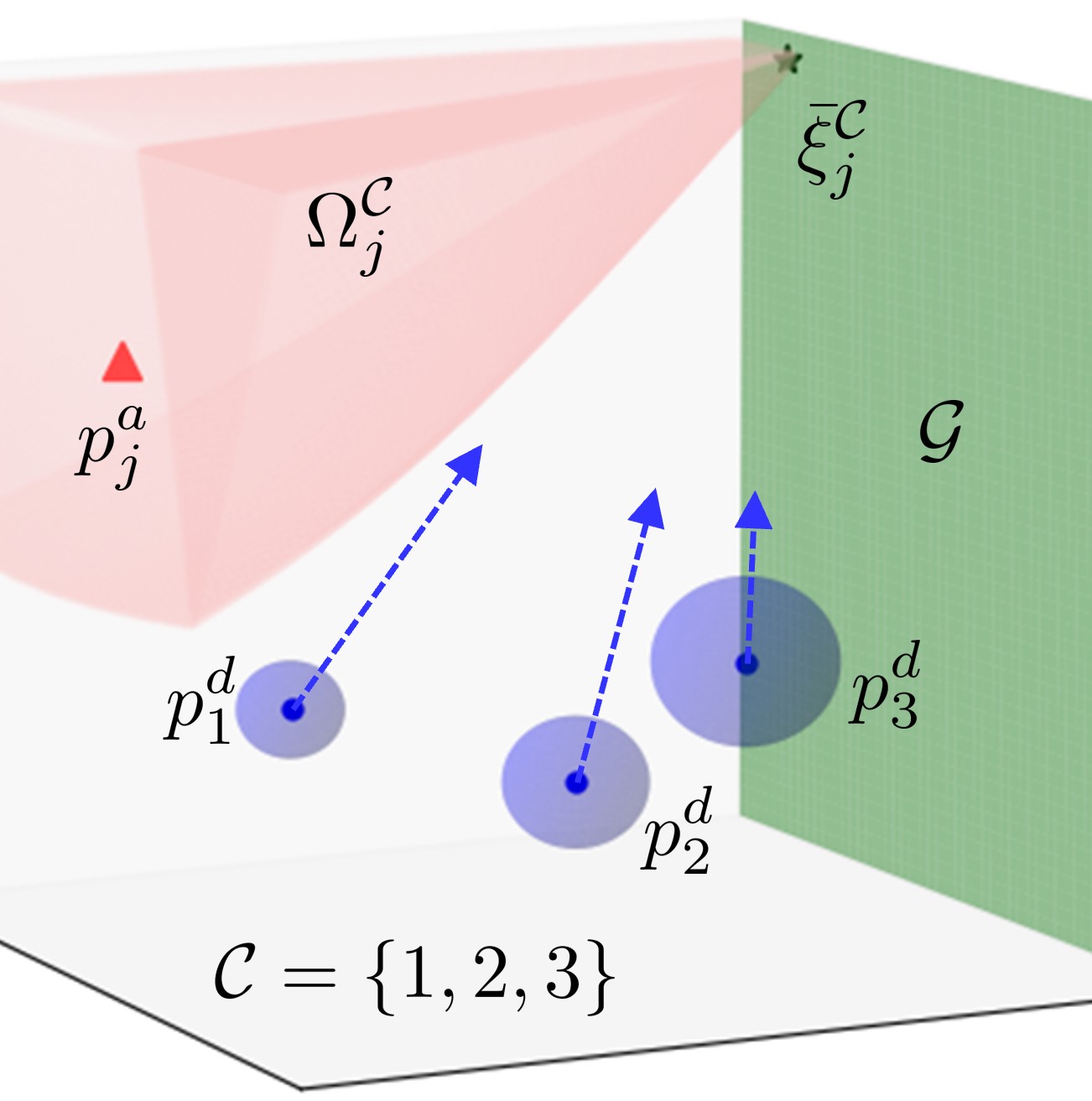}
	\caption{\footnotesize {Defense coordination in both 2D and 3D cases. On the left, the almost-optimal waypoint $\xi^\C_j$ represents the point within the SRS $\Omega^\C_j$ that is closest to the target set $\G$. On the right, the almost-optimal waypoint $\bar{\xi}^\C_j$ signifies the point within the intersection of the SRS $\Omega^\C_j$ and the target set $\G$ that is closest to the attacker. Both almost-optimal waypoints offer intuitive reference points for the coalition to intercept the attacker in various scenarios.}
	\label{fig:odc}}
\end{figure}

\begin{Remark}[Almost-Optimal Waypoint for Defense Coordination]
Algorithm \ref{alg:dmsdc} suggests that the $n$-dimensional vector defined by
\begin{equation}\label{owp}
	\begin{split}
		\tilde{\xi}^{\C}_j=
		\begin{cases}
		\xi^{\C}_j, &x^{\C}_j\in \D^{\C}_j\\
		\bar{\xi}^{\C}_j,& \mathrm{otherwise}
		\end{cases}	
    \end{split}
\end{equation}
which is obtained by solving the convex program \eqref{co:d} or \eqref{co:a}, serves as an almost-optimal waypoint for the coalition. This waypoint provides a reference point for the defenders in the coalition to coordinate their movements, depending on whether the joint state lies within the region $\D^{\C}_j$ or not, as illustrated in Fig. \ref{fig:odc}. By following the almost-optimal waypoint, the coalition can effectively adapt their defense strategy to defend against a single attacker.
\end{Remark}

\begin{Remark}[Individual Attack Strategy]
As a byproduct of our analysis, we can also derive a strategy for the single attacker. Contrary to the defenders' goal, the attacker seeks to minimize $\Phi^\C_j$ robustly 
 for $x^\C_j\in \D^\C_j$, and $\bar{\Phi}^\C_j$ robustly for $x^\C_j\notin \D^\C_j$. Employing a similar approach to that used in the derivation of the defense strategy, we arrive at the following attack strategy:
\begin{equation}\label{as}
    \pi^a_j(x^\C_j)=v^a_{j,\max}\n(\tilde{\xi}^{\C}_j-p^a_j)
\end{equation}
where $\tilde{\xi}^{\C}_j$, defined as \eqref{owp}, also serves as an almost-optimal waypoint for the attacker. The detailed derivation is omitted in this paper, as our primary focus lies on the defense perspective.
\end{Remark}

\section{Defense Allocation for Multiple Attackers}\label{sec:4}

In this section, we delve into the dynamic problem of allocating coalitions to multiple attackers. Building upon our previous results on defense coordination for single-attack scenarios, we reformulate the defense allocation problem as a joint state-dependent integer linear program (ILP), but solving this ILP in real-time may become computationally infeasible as the agent size grows. To address this issue, we propose an alternative solution that integrates two key components: the Hierarchical Integer Linear Programming (HILP) method and the Monotonic Defense Enhancement Allocation (MDEA) algorithm. The HILP method iteratively constructs a hierarchy based on the concepts of active defense sets and irreducibility, leading to a computationally efficient suboptimal solution. The MDEA algorithm then incorporates this suboptimal solution while enforcing a monotonicity constraint, ensuring that the expected number of attackers that fail to reach the target set remains non-decreasing over time.

\subsection{Integer Linear Program Formulation}

Following the classical taxonomy of multirobot task allocation proposed by Gerkey et al. \cite{gerkey2004formal}, the defense allocation problem can be categorized as a single-task robot, multirobot task, and instantaneous assignment problem. Under the current joint state, each defender can defend against at most one attacker at a time, each attacker can be simultaneously guarded by multiple defenders, and the assignment only concerns the present moment. To represent the assignment, a binary matrix called the assignment matrix $\A=(a_{ij})_{N\times M}$ is used, where the entry $a_{ij}=1$ if attacker $j$ is assigned to defender $i$, and 0 otherwise. In addition, since all defenders are single-task, the assignment matrix must respect the \textit{conflict-free} constraint, i.e., the row sum of the assignment matrix should not exceed 1.

To better model the effect of each coalition-attacker pair on the defense performance, we introduce an alternative matrix representation. Let $\C_k$, $k=1,\ldots,\tilde{N}$ be the list of all different coalitions in which $\tilde{N}=2^{N}-1$. We define the \textit{coalition assignment matrix} to be $\Theta=(\theta_{kj})_{\tilde{N}\times M}$ such that
\begin{equation}\label{con:binary}
	\theta_{kj}\in \{0,1\},~\forall k=1,\ldots,\tilde{N},~j=1,\ldots,M
\end{equation}
and $\theta_{kj}=1$ if attacker $j$ is allocated to coalition $\C_k$. The coalition assignment matrix $\Theta$ inherently induces an assignment matrix $\A$ through the linear transformation:
\begin{equation}\label{mtx:inc}
    \A=\B\Theta
\end{equation}
where $\B=(b_{ik})_{N\times \tilde{N}}$ denotes the \textit{incidence matrix} representing the binary relationship between defenders and coalitions. Specifically, $b_{ik}=1$ if $i\in \C_k$ and 0 otherwise. To maintain the single-task property, the conflict-free constraint is defined through the induced assignment matrix as
\begin{equation}\label{con:cf}
	\sum_{j=1}^{M}\sum_{k=1}^{\tilde{N}}b_{ik}\theta_{kj}\leq 1,~\forall i=1,\ldots,N.
\end{equation}
Additionally, the \textit{redundancy-free} constraint is imposed on $\Theta$:
\begin{equation}\label{con:rf}
    \sum_{k=1}^{\tilde{N}}\theta_{kj}\leq 1,~\forall j=1,\ldots,M
\end{equation}
ensuring that each attacker is assigned to at most one coalition.

We now present an objective function incorporating decision variables as elements of the coalition assignment matrix. Recalling Theorem \ref{thm:dws}, if $\Phi^{\C_k}_j>0$ at the joint state $x^{\C_k}_j$, an admissible defense strategy for $\C_k$ exists that prevents attacker $j$ from reaching the target set. In light of this, we designate $(\C_k,j)$ as a \textit{feasible} coalition-attacker pair at $x^{\C_k}_j$ if $\Phi^{\C_k}_j(x^{\C_k}_j)>0$. In line with the payoff function given in \eqref{payoff}, the \textit{reward} for assigning attacker $j$ to coalition $\C_k$, denoted by $w_{kj}$, is set as 1 if $(\C_k,j)$ is feasible and attacker $j$ is active, and 0 otherwise, i.e.,
\begin{equation*}
	w_{kj}(x_j)=\rho_j(x_j)\n(\Phi^{\C_k}_j(x^{\C_k}_j))
\end{equation*}
with $\rho_j$ given in \eqref{rho_j}. Subsequently, the objective function, depending on the choice of a coalition assignment matrix $\Theta$ at a given joint state $x$, is defined as the total sum of the rewards of all assigned coalition-attacker pairs:
\begin{equation}\label{fun:maof}
	\Gamma(\Theta,x)=\sum_{k=1}^{\tilde{N}}\sum_{j=1}^{M}w_{kj}(x_j)\theta_{kj}.
\end{equation}
We refer to $\Gamma$ as the \textit{multi-attack objective function}, which signifies the expected number of attackers that will be unable to access the target set from the current joint state. The reason behind this objective function is that it enables the assessment of defense performance in multi-attack scenarios.

\begin{Lemma}[Defense Performance Assessment]\label{lem:oe}
Let $\pi^d$ be an admissible defense strategy associated with time-dependent coalition assignment matrices $\Theta(\cdot)$ that consistently satisfy constraints \eqref{con:binary}, \eqref{con:cf}, and \eqref{con:rf}. Suppose there exists a time instant $t_0\geq0$, such that for all admissible attack strategies, 
\begin{equation}\label{ineq:gamma}
	\Gamma(\Theta(t),x(t))+N_c(t)\geq M_c
\end{equation}
holds over the time interval $[t_0,t_f)$, where $N_c(t)$ is the number of captured attackers up to time $t$, and $M_c$ is a positive integer. Then, the payoff function $J$ under the defense strategy $\pi^d$ is bounded by $M-M_c$, i.e., 
\begin{equation*}
	J(x(0),\pi^d,\pi^a)\leq M-M_c
\end{equation*}
for any admissible attack strategy $\pi^a$.
\end{Lemma}

\begin{Proof}
Let $N_e(t)$ denote the number of attackers that have entered the target set up to time $t$. At time $t$, each attacker must be either captured, active, or have entered the target set. Hence, we have $M=N_e(t)+M_a(t)+N_c(t)$. Since the multi-attack objective function $\Gamma$ only concerns active attackers, it follows that $\Gamma(\Theta(t),x(t))\leq M_a(t)$. Combining this with inequality \eqref{ineq:gamma} yields $M_c\leq\Gamma(\Theta(t),x(t))+N_c(t)\leq M_a(t)+N_c(t)$ for all $t\in [t_0,t_f)$, implying that $N_e(t)\leq M-M_c$ for all $t\in [t_0,t_f)$. Consequently, based on the definition of the payoff function, we obtain $J(x(0),\pi^d,\pi^a)=N_e(t_f)\leq M-M_c$.
\end{Proof}

\begin{Remark}[Online Maximization and Monotonicity Constraint]
Lemma \ref{lem:oe} implies that one way to minimize the payoff function $J$, against any admissible attack strategies, is to maximize the multi-attack objective function $\Gamma$ with respect to the coalition assignment matrix $\Theta$ at each joint state $x$. This leads to the following integer linear program (ILP) for the multi-attack defense allocation problem along the joint state trajectory:
\begin{equation}\label{ilp}
    \begin{split}
	    &\max_{\Theta}~\Gamma(\Theta,x)\\
	    &~\mathrm{s.t.}~~\eqref{con:binary},\eqref{con:cf},\eqref{con:rf}.
    \end{split}
\end{equation}
Furthermore, to ensure overall defense performance, Lemma \ref{lem:oe} highlights the significance of maintaining a robustly positively invariant superlevel set for the function $\Gamma + N_c$ with a level $M_c > 0$. Here, the value $\Gamma(\Theta(t),x(t))+N_c(t)$ represents the expected number of attackers that fail to reach the target set at time $t$. A sufficient condition for preserving the superlevel set property is to impose the monotonicity constraint, given by:
\begin{equation}\label{con:m}
    \begin{split}
        &\Gamma(\Theta(t_l),x(t_l))+N_c(t_l)\\
        \geq&\Gamma(\Theta(t_{l-1}),x(t_{l-1}))+N_c(t_{l-1}),~\forall l=1,\ldots,L
    \end{split}
\end{equation}
where $t_l$, $l=0,\ldots,L$ are the allocation times with $t_0=0$. It should be noted that the optimal solution to the ILP \eqref{ilp} automatically satisfies the monotonicity constraint \eqref{con:m}.
\end{Remark}

Although solving the ILP \eqref{ilp} provides an optimal solution to the defense allocation problem, it entails a substantial computational burden, particularly as the number of agents increases. This is mainly due to the vast amount of possible coalition-attack pairs, which not only leads to a large-scale ILP but more importantly, requires considerable computational effort to determine the multi-attack objective function. In the case of $N$ defenders and $M_a$ active attackers, there are $\tilde{N}M_a$ possible coalition-attacker pairs, implying that the same amount of convex programs of the form \eqref{co:d} need to be calculated to specify the corresponding rewards. 
To alleviate the computational burden, we propose a heuristic algorithm for the ILP \eqref{ilp} in Section \ref{sec:4.3} to generate a suboptimal solution, which will then be fed into an allocation algorithm to ensure the monotonicity constraint \eqref{con:m} in Section \ref{sec:4.4}. In particular, prior to detailing the heuristic algorithm, we will derive two critical components in the next section, which together construct a hierarchy that facilitates the generation of a suboptimal solution for the ILP \eqref{ilp}.

\subsection{Active-Defense Set and Irreducibility}

The optimal solution to the ILP \eqref{ilp} tends to favor feasible coalition-attacker pairs comprising smaller coalitions. On one hand, the multi-attack objective function is influenced only by feasible coalition-attacker pairs, as the reward for an infeasible pair is zero. On the other hand, smaller coalitions are less likely to violate the conflict-free constraint \eqref{con:cf}, and thus more likely to contribute positively to the multi-attack objective function. As a result, the key to striking a balance between optimality and efficiency in solving the ILP \eqref{ilp} is to identify as many feasible sub-pairs of coalition-attacker pairs with small amounts of defenders as possible while keeping the computational cost low. Here, a \textit{sub-pair} of a coalition-attacker pair refers to a subset of the coalition that is also paired with the same attacker. To this end, we will focus on seeking feasible sub-pairs of a coalition-attacker pair from the aspects of optimality and efficiency in the subsequent analysis.
 
As a point of departure, we revisit the convex program \eqref{co:d} subject to a feasible coalition-attacker pair $(\C,j)$. As shown in Theorem \ref{thm:dws}, the vector $\xi^{\C}_j$ extracted from the solution to \eqref{co:d} establishes an almost-optimal waypoint for the coalition $\C$, which must lie on the boundary of the SRS $\Omega^{\C}_j$. Moreover, owing to the convexity of the SRS, some of the inequality constraints $c_{ij}(q,x_{ij})\leq 0$, $i\in \C$ must be active at $q=\xi^{\C}_j$. As such, we can define an index set of defenders through an active set in the convex program \eqref{co:d} that incorporates the almost-optimal waypoint $\xi^{\C}_j$.

\begin{Definition}[Active-Defense Set]\label{def:ads}
Suppose the coalition-attacker pair $(\C,j)$ is feasible at the joint state $x^\C_j$. The \textit{active-defense set} (ADS) $\Lambda^\C_j(x^\C_j)$ for $(\C,j)$ is the active set of the inequality constraints $c_{ij}(q,x_{ij})\leq 0$, $i\in\C$ in \eqref{co:d} at the almost-optimal waypoint $\xi^\C_j$, i.e., 
\begin{equation}\label{ads}
	\Lambda^\C_j(x^\C_j)=\left\lbrace i\in \C\mid c_{ij}(\xi^\C_j,x_{ij})=0\right\rbrace.
\end{equation}
\end{Definition}

\begin{figure}[tp!] 
	\centering
	\includegraphics[width=0.9\linewidth]{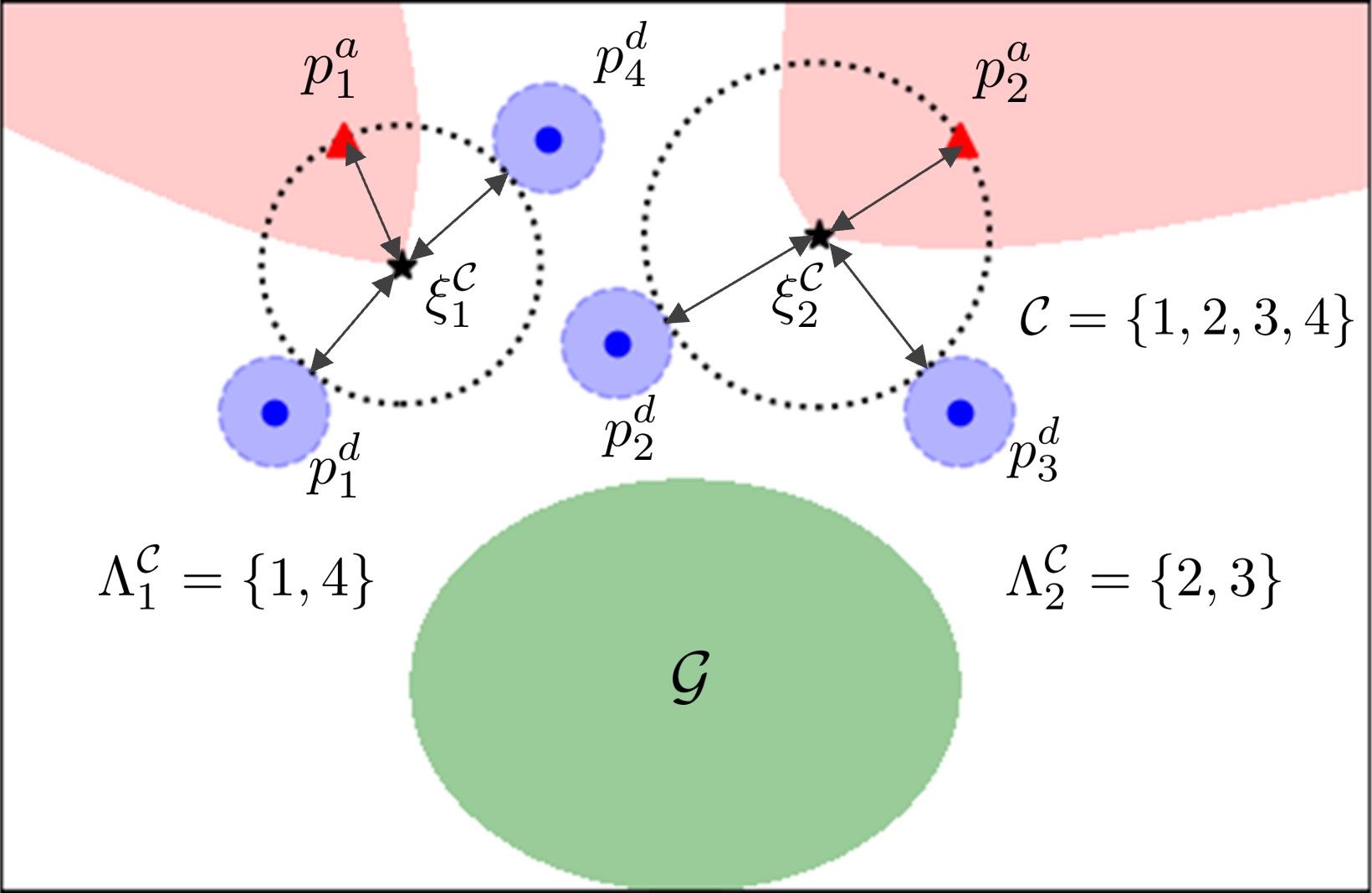}
	\caption{\footnotesize {Geometric interpretation of the ADS in a 2D domain for the case of $\gamma_{ij}=1$. In this case, the ADS $\Lambda^\C_j$ is the set of defender indices such that the capture region for each defender $i\in \Lambda^\C_j$ (illustrated as a blue disk) is tangent to the dashed circle of radius $||\xi^\C_j-p^a_j||$ centered at the almost-optimal waypoint $\xi^\C_j$.}
	\label{fig:ads}}
\end{figure}

The construction of the ADS does not necessitate the computation of any additional convex programs. This is because the feasibility of the coalition-attacker pair has already been verified using a convex program of the form \eqref{co:d}, and the almost-optimal waypoint can be determined as a byproduct of this computation. It is worth noting that the ADS concept indeed provides a proximity-based method for obtaining a specific sub-pair, which is associated with both locations of the attacker and the target set, as illustrated in Fig. \ref{fig:ads}. Furthermore, it can be shown that such a sub-pair inherits the feasibility of the original coalition-attacker pair.

\begin{Proposition}[Feasibility]\label{prop:f_ads}
If the coalition-attacker pair $(\C,j)$ is feasible at the joint state $x^{\C}_j$, then the sub-pair $(\Lambda^{\C}_j,j)$ is also feasible at the corresponding joint state.
\end{Proposition}

\begin{Proof}
See Appendix \ref{proof:f_ads}.
\end{Proof}

We turn to look at how to generate feasible sub-pairs of a sufficiently small scale. While the ADS allows for feasible sub-pair generation without incurring additional computational effort, the resulting sub-pair size may not be minimum with respect to feasibility. In other words, the sub-pair may include redundant defenders that could be removed without affecting the feasibility of the sub-pair. To precisely quantify the redundancy, we introduce the following definition.

\begin{Definition}[Irreducibility]\label{def:irre}
A feasible coalition-attacker pair $(\C,j)$ is said to be \textit{irreducible} at a joint state $x^{\C}_j$ if there is no proper subset $\C'$ of $\C$ such that $(\C',j)$ is feasible.
\end{Definition}

Given a feasible coalition-attacker pair $(\C,j)$, we can develop an iterative procedure to generate all irreducible sub-pairs. The process commences by evaluating size-1 subsets of $\C$. If $|\C|=1$, $(\C,j)$ is inherently deemed irreducible and the process terminates. Otherwise, we check each size-1 subset $\C'$ for irreducibility by confirming whether $\Phi^{\mathcal{C}'}_j>0$. Then, we remove the element of $\C'$ from $\C$ for which $(\C',j)$ is irreducible, forming a new set $\mathcal{C}_r^{(1)}$. If $|\C_r^{(1)}|\leq 1$, all irreducible sub-pairs have been identified, and the process ends. In case further iterations are required, we repeat with all size-$l$ subsets of $\C_r^{(l-1)}$ for $l=2,\ldots,|\C|$, where $\C_r^{(l)}$ is analogously defined by removing elements from $\C_r^{(l-1)}$. The procedure continues until either $|\C_r^{(l)}|\leq l$ or $|\C|=l$ at the iteration $l$. Moreover, the number of iterations can be reduced by noting that the cardinality of the coalition for any irreducible sub-pair is upper bounded by the dimension of the domain.

\begin{Proposition}[Upper Bounded Cardinality]\label{prop:ub}
In an $n$-dimensional domain, if the coalition-attacker pair $(\C,j)$ is irreducible, then $\C$ has cardinality no greater than $n$.
\end{Proposition}

\begin{Proof}
See Appendix \ref{proof:ub}.
\end{Proof}

\begin{algorithm}[tp!]
\caption{Irreducible Sub-Pairs of $(\C,j)$}\label{alg:irre}
\KwData{Positions $p^d_i\in \R^n$, $i\in \C$ and $p^a_j\in \R^n$}
\KwResult{The set consisting of irreducible sub-pairs $\P$}
\textbf{Initialization:} $\P\gets\emptyset$, $\C_r\gets \C$\\
\For{$i=1$ \KwTo $n$}{
    \uIf{$|\C_r|<i$}{
        \textbf{break}
    }
    \uElseIf{$|\C|=i$}{
        Put $(\C,j)$ in $\P$\\
        \textbf{break}
    }
    \Else{
        \ForAll{$\C'\subset \C_r$ with $|\C'|=i$}{
            $\Phi^{\C'}_j\gets$ Optimal value of \eqref{co:d} s.t. $(\C',j)$\\
            \If{$\Phi^{\C'}_j>0$}{ 
                Put $(\C',j)$ in $\P$\\
                Remove the elements of $\C'$ from $\C_r$
            }
        }
    }
}
\end{algorithm}

We summarize our findings of identifying irreducible sub-pairs of a feasible coalition-attacker pair in Algorithm \ref{alg:irre}. It should be emphasized that when solving the ILP \eqref{ilp}, we can focus on irreducible coalition-attacker pairs rather than all possible pairs. This is because replacing a coalition-attacker pair with one of its irreducible sub-pairs will not affect the optimal value of the multi-attack objective function. In general, Algorithm \ref{alg:irre} can lower the number of convex programs needed to determine the multi-attack objective function. However, directly using the concept of irreducibility could lead to heavy computational loads. For instance, computing all irreducible sub-pairs in the worst-case scenario requires up to compute $\tilde{N}$ convex programs for a coalition-attacker pair $(\N,j)$. To improve computational efficiency, a more effective approach would be to combine the concept of irreducibility with that of the ADS.

\subsection{Hierarchical Integer Linear Programming}\label{sec:4.3}

We now exhibit the integration of the ADS and irreducibility concepts to establish the first level of the hierarchy for the ILP \eqref{ilp}. Our approach initiates by checking coalition-attack pairs of the form $(\N,j)$ and only focusing on feasible ones. The proximity-based sub-pair $(\Lambda^{\N}_j,j)$ is then taken into account, where $\Lambda^{\N}_j$ is the ADS for the feasible pair $(\N,j)$. Finally, irreducible sub-pairs of $(\Lambda^{\N}_j,j)$ are selected to refine the multi-attack objective function $\Gamma$. As a result, we assign the highest priority to coalition-attacker pairs $(\C,j)$ that are irreducible and have $\C$ as a subset of $\Lambda^{\N}_j$. Compared to directly prioritizing all irreducible sub-pairs of $(\N,j)$, this method reduces the potential computational challenges while offering a considerable amount of feasible sub-pairs.

After specifying the first hierarchy level, we can derive a suboptimal solution to the ILP \eqref{ilp}. Let $\Xi_1$ denote the first priority set that contains all coalition-attacker pairs with the highest priority. In the case where $\Xi_1$ is empty, no feasible coalition-attacker pairs exist, and the ILP solution simply corresponds to a zero coalition assignment matrix. To confine the selection of coalition-attacker pairs solely to those within $\Xi_1$, we impose the following constraint on the decision variables:
\begin{equation}\label{con:binary_1}
	\theta_{kj}=0,~\forall (\C_k,j)\notin \Xi_1.
\end{equation}
By resolving the ILP \eqref{ilp} with the additional constraint \eqref{con:binary_1}, we attain a suboptimal solution denoted as $\Theta_1$, which provides a lower bound on the optimal value of the multi-attack objective function. In order to evaluate $\Theta_1$, we define two index sets $\N_r^{(1)}$ and $\M_r^{(1)}$ in which $\N_r^{(1)}$ contains the indices of defenders that remain unassigned to any attacker based on $\Theta_1$, and $j\in \M_r^{(1)}$ if $\Theta_1$ omits attacker $j$ and $(\N,j)$ is feasible. If either $\N_r^{(1)}$ or $\M_r^{(1)}$ is empty, there are no more available coalition-attacker pairs and further steps are not needed.

The above procedure can be iteratively applied, if possible, to improve the suboptimal solution. At iteration $l\geq 2$, the $l$-th priority set $\Xi_l$ is formed by identifying all irreducible coalition-attacker pairs $(\C,j)$ for which $\C$ is contained in the ADS for a feasible $(\N^{(l-1)}_r,j)$ and $j\in \M_r^{(l-1)}$. The decision variables are then restricted to $\Xi_l$ through the binary constraint:
\begin{equation}\label{con:binary_iter}
	\begin{split}
	    &\theta_{kj}=0,~\forall (\C_k,j)\notin \Xi_l
    \end{split}
\end{equation}
resulting in a subproblem of the original ILP \eqref{ilp}:
\begin{equation}\label{ilp_iter}
    \begin{split}
	    &\max~\Gamma(\Theta,x)\\
	    &~\mathrm{s.t.}~\eqref{con:binary},\eqref{con:cf},\eqref{con:rf},\eqref{con:binary_iter}.
    \end{split}
\end{equation}
Having a solution to the subproblem \eqref{ilp_iter}, represented by $\Theta_l$, the suboptimal solution to the ILP \eqref{ilp} is updated by adding $\Theta_l$ to the previous suboptimal solution, leading to
\begin{equation*}
    \Theta_{\text{HILP}}=\sum_{k=1}^{l}\Theta_k.
\end{equation*}
The procedure terminates at the $l$th iteration when $\Xi_l$ is empty or one of the index sets $\N_r^{(l)}$ and $\M_r^{(l)}$ is empty, where $\N_r^{(l)}$ and $\M_r^{(l)}$ are obtained by excluding the indices related to $\tilde{\Theta}_l$ from $\N_r^{(l-1)}$ and $\M_r^{(l-1)}$, respectively.

\begin{algorithm}[tp!]
\caption{Hierarchical Integer Linear Programming}
\label{alg:hilp}
\KwData{Positions $p^d_i$, $i\in \N$ and $p^a_j$, $j\in \M_a$}
\KwResult{Coalition assignment matrix $\Theta_{\text{HILP}}$}
Initialize $\Theta_{\text{HILP}}\gets 0_{\tilde{N}\times M}$, $\N_r\gets \N$
$\M_r\gets \M_a$\\
\While{$\N_r\neq\emptyset$ and $\M_r\neq\emptyset$}{
    $\Xi\gets \emptyset$\\
    \ForAll{$j\in \M_r$}{
        $\Phi^{\N_r}_j\gets$ Optimal value of \eqref{co:d} s.t. $(\N_r,j)$\\
        \eIf{$\Phi^{\N_r}_j>0$}{
            $\Lambda^{\N_r}_j\gets$ ADS for $(\N_r,j)$\\
            $\P\gets$ Result of Algorithm \ref{alg:irre} s.t. $(\Lambda^{\N_r}_j,j)$\\
            Put all elements of $\P$ in $\Xi$
        }{
            Remove $j$ from $\M_r$
        }
    }
    \If{$\Xi=\emptyset$}{
        \textbf{break}
    }
    $\Theta\leftarrow$ A solution to the ILP \eqref{ilp_iter} with \eqref{con:binary_iter} s.t. $\Xi$\\
    $\Theta_{\text{HILP}}\gets\Theta_{\text{HILP}}+\Theta$\\
    Remove all $i$ with $(\B\Theta)(i,:)\neq 0_{1\times M}$ from $\N_r$\\
    Remove all $j$ with $\Theta(:,j)\neq 0_{\tilde{N}\times 1}$ from $\M_r$
}
\end{algorithm}

In summary, our approach to solving the ILP \eqref{ilp} is outlined in Algorithm \ref{alg:hilp}, which strikes a balance between optimality and efficiency, as demonstrated by the subsequent proposition.

\begin{Proposition}[Optimality-Efficiency Tradeoff]\label{prop:oet}
Algorithm \ref{alg:hilp} possesses two key properties:

1) (Optimality) If $\M_r$ is empty after the first iteration, then the suboptimal solution $\Theta_{\text{HILP}}$ is an optimal solution to the ILP \eqref{ilp}.

2) (Efficiency) If the cardinality of the ADS for each $(\N_r,j)$ in each iteration is at most $n$, then the total number of calculations for the convex program of the form \eqref{co:d} is less than $2^{n-1}M_a(1+M_a)$.
\end{Proposition}

\begin{Proof}
See Appendix \ref{proof:oet}.
\end{Proof}

The assumptions made in Proposition \ref{prop:oet} for Algorithm \ref{alg:hilp} are satisfied in many practical scenarios. For instance, if the ADSs $\Lambda^{\N_r}_j$ for the coalition-attacker pair $(\N,j)$, $j\in \M_a$ are mutually disjoint, then the set $\M_r$ becomes empty after the first iteration and the suboptimal solution $\tilde{\Theta}$ is optimal for the ILP \eqref{ilp}. Such a scenario commonly arises when the attackers are sparsely distributed relative to the defenders. Furthermore, empirical results indicate that the cardinality of the ADSs typically does not exceed the number of defenders. Compared to directly solving the ILP, where the number of calculations for the convex program of the form \eqref{co:d} grows exponentially with the number of defenders, the calculation times for our algorithm do not empirically depend on the number of defenders and are only quadratically dependent on the number of active attackers. Consequently, Algorithm \ref{alg:hilp} offers a viable alternative for tackling the ILP in large-scale scenarios with many defenders, especially when conventional approaches become computationally infeasible.

\subsection{Monotonic Defense Enhancement Allocation}\label{sec:4.4}

Once acquiring the suboptimal solution produced by Algorithm \ref{alg:hilp}, we adjust it by incorporating the monotonicity constraint \eqref{con:m}. Specifically, we update the coalition assignment matrix at each time step by equating it to the solution derived from Algorithm \ref{alg:hilp} if the monotonicity constraint \eqref{con:m} is met; otherwise, we retain the updated coalition assignment matrix from the previous step. Note that the number of active attackers may change between two consecutive time steps. As such, we need to set the columns of the previous coalition assignment matrix to zero for those corresponding to attackers captured at the current time step before executing the update. This approach offers the advantage of maintaining the validity of all inequalities in \eqref{ilp} without requiring additional computation of the ILP, thereby reducing the computational effort.

While the above approach effectively integrates the monotonicity constraint \eqref{con:m}, it may give rise to undesired oscillatory behaviors in defense allocation. Such oscillations occur when a defender's assignment repeatedly switches between two distinct attackers, potentially impairing the system's performance. These oscillations stem from the fact that Algorithm \ref{alg:hilp} might produce non-unique solutions, leading to the equality in \eqref{con:m} holding. To tackle this issue, we implement a stricter version of the monotonicity constraint \eqref{con:m}, updating the coalition assignment matrix only when the inequality constraint \eqref{con:m} is strictly fulfilled. This ensures that we stick to the previous coalition assignment matrix when the equality in \eqref{con:m} holds, thus avoiding oscillations while still enforcing the monotonicity constraint \eqref{con:m}. 

\begin{algorithm}[tp!]
\caption{Monotonic Defense Enhancement Allocation}\label{alg:mdea}
\KwData{Positions $p^d_i$, $i\in \N$ and $p^a_j$, $j\in \M_a$, solution of Algorithm \ref{alg:hilp} $\Theta_{\text{HILP}}$, previous coalition assignment matrix $\tilde{\Theta}^p$, previous index set of active attackers $\M_a^p$}
\KwResult{Coalition assignment matrix $\Theta_{\text{MDEA}}$}
\If{$\M_a=\emptyset$}{
    \Return $\Theta_{\text{MDEA}}\gets 0_{\tilde{N}\times M}$
}
\textbf{Part I: Monotonicity constraint enforcement}\\
$\Gamma\gets$ Number of nonzero column vectors in $\Theta_{\text{HILP}}$\\
$\Gamma^p\gets$ Number of nonzero column vectors in $\tilde{\Theta}^p$\\
\eIf{$\Gamma>\Gamma^p-|\M_a^p|+|\M_a|$}{
    $\Theta_{\text{MDEA}}\gets \Theta_{\text{HILP}}$}{
    \If{$|\M_a^p|\neq |\M_a|$}{
        \ForAll{$j\in \M_a^p\setminus \M_a$}{
        $\tilde{\Theta}^p(:,j)\gets 0_{\tilde{N}\times 1}$
        }
    }
    $\Theta_{\text{MDEA}}\gets \tilde{\Theta}^p$
}
$\tilde{\Theta}^p\gets\Theta_{\text{MDEA}}$, $\M_a^p\gets\M_a$\\
\textbf{Part II: Greedy assignment of unassigned defenders}\\
$\A\gets \B\Theta_{\text{MDEA}}$\\
Put all $i\in \N$ with $\A(i,:)=0_{1\times M}$ in $\N_r$\\
Put all $j\in \M_a$ with $\A(:,j)=0_{N\times 1}$ in $\M_r$\\
\If{$\M_r\neq \emptyset$ and $\N_r\neq \emptyset$}{
    \ForAll{$i\in \N_r$}{
        $j_{\min}\gets \arg\min_{j\in\M_r} ||p^d_i-p^a_j||$\\
        $(i,j_{\min})$th element of $\A\gets 1$
    }
}
$\Theta_{\text{MDEA}}\gets \B^{\dagger}\A$ with $\B^{\dagger}$ as the pseudoinverse of $\B$\\
\end{algorithm}

Our approach to the multi-attack defense allocation problem is summarized in Algorithm \ref{alg:mdea}, which consists of two components. The first component imposes the monotonicity constraint \eqref{con:m} on the solution obtained from Algorithm \ref{alg:hilp}. For the initial step, $\tilde{\Theta}^p$ and $\M^p_a$ are respectively set to $\Theta_{\text{HILP}}$ and $\M^p_a=\M$. The second part uses a greedy algorithm that allocates each unassigned defender to the nearest unassigned active attacker. This is inspired by the observation that a closer the defender-attacker pair results in a reduced SRS. To achieve this, we first convert the coalition assignment matrix into the corresponding assignment matrix in terms of the linear equation \eqref{mtx:inc}. We then update the assignment matrix using the greedy algorithm and revert it to the coalition assignment matrix through the pseudoinverse of the incidence matrix. The greedy algorithm is computationally efficient and provides a valuable supplement to the first component of Algorithm \ref{alg:mdea}.

We are now able to encode the solution to the multi-attack defense allocation problem into an admissible defense strategy for the multiplayer reach-avoid game. Given the output of Algorithm \ref{alg:mdea}, $\Theta_{\text{MDEA}}=(\theta_{kj})$, we define the assignment rule as follows:
\begin{equation}\label{asgn}
	\begin{split}
		\sigma(i)=
		\begin{cases}
		j, &i\in \C_k~\&~\theta_{kj}=1\\
		\mathrm{null},& \text{otherwise}
		\end{cases}
	\end{split}
\end{equation}
where $\mathrm{null}$ denotes the case of an unassigned defender. For each active attacker $j\in \M_a$, the coalition against attacker $j$ induced by $\sigma$ is given by:
\begin{equation*}
	\tilde{\C}_j=\{i\in \N\mid\sigma(i)=j\}.
\end{equation*}
Combining Algorithm \ref{alg:dmsdc} with the assignment rule $\sigma$ as described in \eqref{asgn}, we obtain an admissible defense strategy of the form:
\begin{equation}\label{ds:m}
	\tilde{\pi}^d_i(x)=v^d_{i,\max}\n(\xi_{\sigma(i)}-p^d_i),~i\in \N
\end{equation}
where $\xi_j=\tilde{\xi}^{\tilde{\C}_j}_j$ is the almost-optimal waypoint defined in \eqref{owp}. If a defender is unassigned, its control input is set to zero for the sake of energy saving.

\begin{Theorem}[Performance-Guaranteed Defense Strategy]\label{thm:pgds}
For any initial joint state $x_0$, the payoff function $J$ under the defense strategy $\tilde{\pi}^d$ given by \eqref{ds:m}, and any admissible attack strategy satisfies the following inequality:
\begin{equation}\label{ineq:payoff}
	J(x_0,\tilde{\pi}^d,\pi^a)\leq M-\Gamma(\Theta_{\text{MDEA}}^0,x_0)-N_c(0)
\end{equation}
where $\Gamma$ is the multi-attack objective function defined in \eqref{fun:maof}, $\Theta_{\text{MDEA}}^0$ is the initial solution of Algorithm \ref{alg:mdea}, and $N_c(0)$ is the number of attackers captured at the initial joint state. Moreover, the function $\Gamma+N_c$ is monotonically non-decreasing along the system trajectory.
\end{Theorem}

\begin{Proof}
The enforcement of the monotonicity constraint \eqref{con:m} in Algorithm \ref{alg:mdea} ensures that the sequence $\Gamma(\Theta_{\text{MDEA}}(t_l),x(t_l))+N_c(t_l)\}_{l=0}^L$ is monotonically non-decreasing. Further, it is noted that $\Gamma(\Theta_{\text{MDEA}}(t),x(t))+N_c(t)$ remains constant within each time interval $[t_{l-1},t_l)$, as the assignment of defenders to active attackers is updated only at the time instants $t_l$. As a result, $\Gamma+N_c$ is non-decreasing. This indicates that the inequality \eqref{ineq:gamma} holds with $M_c=\Gamma(\Theta_{\text{MDEA}}^0,x_0)+N_c(0)$, which in turn, according to Lemma \ref{lem:oe}, leads to the satisfaction of the inequality \eqref{ineq:payoff} for the payoff function. 
\end{Proof}

Theorem \ref{thm:pgds} provides a solid assurance that our defense strategy \eqref{ds:m} is capable of achieving a performance level no worse than the bound of the payoff function specified in \eqref{ineq:payoff} for any initial joint state. Theorem \ref{thm:pgds} also establishes that the function $\Gamma+N_c$, representing the expected number of attackers that fail to reach the target set, is monotonically non-decreasing over time. This property suggests that the upper bound of the payoff function in \eqref{ineq:payoff} becomes progressively tighter as the game progresses, leading to a continuous improvement in the performance of the proposed strategy. In the next section, we will present empirical results to demonstrate performance enhancement in practical scenarios.

\section{Simulations}\label{sec:5}

In this section, we evaluate the performance and efficiency of our proposed single-attack defense coordination and allocation algorithms through a series of simulated experiments. All algorithms are written in Python, with convex programs and ILP implemented using the CVXPY solver \cite{diamond2016cvxpy}. We conduct the experiments on an Ubuntu 20.04 operating system equipped with an i9-10900 5.2 GHz CPU and 32 GB RAM, featuring 10 physical cores and 20 threads. To demonstrate the effectiveness of our approach, we benchmark our algorithms against current state-of-the-art algorithms in both 2D and 3D scenarios. Initially, we examine the Dual-Mode Switching Defense Coordination (DMSDC) algorithm (Algorithm \ref{alg:dmsdc}), which serves as the basis for subsequent experiments with the monotonic defense enhancement allocation (MDEA) algorithm (Algorithm \ref{alg:mdea}). Furthermore, to demonstrate the applicability of our algorithms in practical settings, we extend our evaluation to a more realistic simulation environment utilizing Gazebo with the robot operating system (ROS) \cite{koenig2004design}.

\subsection{Single-Attack Defense Coordination}\label{sec:V1}

We start with the DMSDC for the problem of single-attack defense coordination. Our approach is compared with two existing methods: the multi-agent pursuit-defense strategy presented in \cite{deng2020multi} for a 2D scene and the analytical HJI-based strategy introduced in \cite{garcia2020optimal} for a 3D scene. We conduct simulations in a 2D rectangular domain of $[-5,5]^2$ with a unit circle as the target set, and a 3D box domain of $[-5,5]^3$ with the origin as the target set. To examine the effectiveness of our approach in guaranteeing the coalition's win under various initial joint states, we perform 2000 randomized simulations with different numbers of defenders and capture radii for both the 2D and 3D scenarios. To ensure a fair comparison, we adopt the attack strategy given in \eqref{as} across all methods and set the maximum speed ratios to 1. The simulation outcomes are classified into four categories based on whether the coalition wins the game:
\begin{itemize}
    \item True Positive: The percentage of cases where both methods successfully capture the attacker.
    \item False Negative: The percentage of cases where our method succeeds while the comparison method fails to capture the attacker.
    \item False Positive: The percentage of cases where our method fails while the comparison method succeeds in capturing the attacker.
    \item True Negative: The percentage of cases where both methods fail in capturing the attacker.
\end{itemize}

Table \ref{table:1} showcases the 2D simulation results, which highlight the superior performance of our proposed algorithm over the multi-agent pursuit-defense strategy. Notably, our algorithm captures the attacker in 7.1\% to 22.4\% more instances than the multi-agent pursuit-defense strategy, without any occurrences where our algorithm fails while the multi-agent pursuit-defense strategy prevails. Interestingly, as the capture radius increases, our algorithm records a higher percentage of false negatives. This observation can be attributed to the fact that the Voronoi diagram employed in the multi-agent strategy does not inherently account for nonzero capture radii, whereas our algorithm exploits the safe-reachable set of the attacker to devise a more effective defense coordination strategy. Moreover, we note that when the number of defenders increases from 2 to 3, both capture radius scenarios exhibit a higher percentage of false negatives. This result arises from the multi-agent pursuit-defense strategy utilizing only one defender for defense, with the remaining defenders acting as pursuers. In contrast, our algorithm capitalizes on the collective strength of multiple defenders to achieve overall coordination.

\begin{table}[!ht]
\caption{Simulation results of a 2D multi-defender single-attacker reach-avoid game.}
\label{table:1}
\centering
\begin{tabular}{llllll}
\hline
Defender & Capture & True & False & False & True \\
Number & Radius & Positive & Negative & Positive & Negative \\ \hline
    ~~~~2 & ~~0.5 & 25.6\% & \textbf{7.1\%} & ~\textbf{0\%} & 67.3\%  \\
    ~~~~3 & ~~0.5 & 33.0\% & \textbf{10.0\%} & ~\textbf{0\%} & 57.0\%  \\
    \hline
    ~~~~2 & ~~3.0 & 33.9\% & \textbf{18.8\%} & ~\textbf{0\%} & 47.3\%  \\
    ~~~~3 & ~~3.0 & 41.6\% & \textbf{22.4\%} & ~\textbf{0\%} & 36.0\%  \\
    \hline
\end{tabular}
\end{table}

On the other hand, the 3D simulation results displayed in Table \ref{table:2} indicate that our proposed algorithm outperforms the analytical HJI-based strategy. Specifically, our algorithm captures the attacker in up to 13.6\% more instances than the analytical HJI-based strategy when the capture radius is 2, and there are no cases in which our algorithm fails while the analytical HJI-based strategy succeeds. It should be emphasized that the analytical HJI-based strategy is designed for cases with zero capture radii; consequently, the performance of our algorithm closely aligns with the analytical HJI-based strategy when the capture radius is near zero, which in turn demonstrates the optimality of our approach. Additionally, our algorithm exhibits better performance as the number of defenders increases, similar to our 2D comparison. This improvement is due to the analytical HJI-based strategy focusing solely on scenarios with one or two defenders, whereas our approach accommodates multiple defenders.

\begin{table}[!ht]
\caption{Simulation results of a 3D multi-defender single-attacker reach-avoid game.}
\label{table:2}
\centering
\begin{tabular}{llllll}
\hline
Defender & Capture & True & False & False & True \\
Number & Radius & Positive & Negative & Positive & Negative \\ \hline
    ~~~~2 & ~~0.1 & 66.5\% & ~\textbf{0.9\%} & ~\textbf{0\%} & 32.6\%\\
    ~~~~3 & ~~0.1 & 72.2\% & ~\textbf{1.6\%} & ~\textbf{0\%} & 26.2\%\\
    \hline
    ~~~~2 & ~~2.0 & 84.3\% & ~\textbf{7.3\%} & ~\textbf{0\%} & 8.4\%\\
    ~~~~3 & ~~2.0 & 81.6\% & ~\textbf{13.6\%} & ~\textbf{0\%} & 5.2\%\\
    \hline
\end{tabular}
\end{table}
 
\subsection{Multiple-Attack Defense Allocation}

\begin{figure}[tp!] 
	\centering
	\subfigure[]{
	\includegraphics[width=1\linewidth]{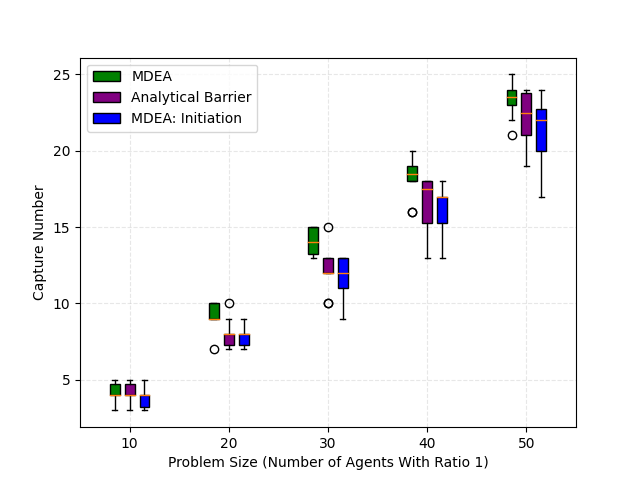}
	\label{fig:tro_1}}
	\subfigure[]{
	\includegraphics[width=1\linewidth]{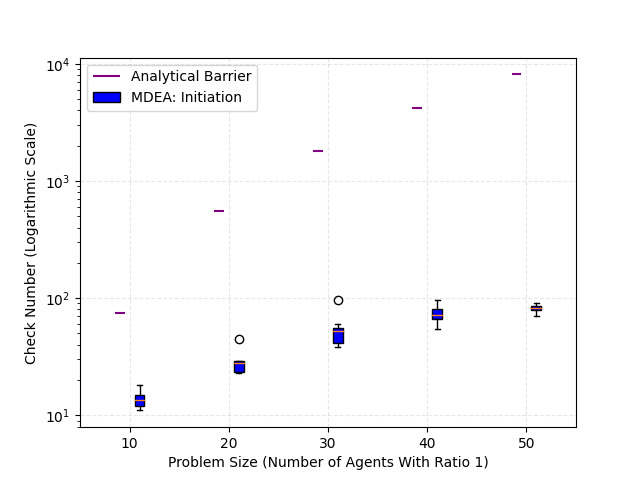}
	\label{fig:tro_2}}
	\caption{\footnotesize Performance comparison between our MDEA algorithm and the analytical barrier approach in a 2D scenario. (a): Capture number comparison. (b): Check number comparison.}
	\label{fig:boxplot_2d}
	\vspace{-5px}
\end{figure}

\begin{figure}[tp!] 
	\centering
	\subfigure[]{
	\includegraphics[width=1\linewidth]{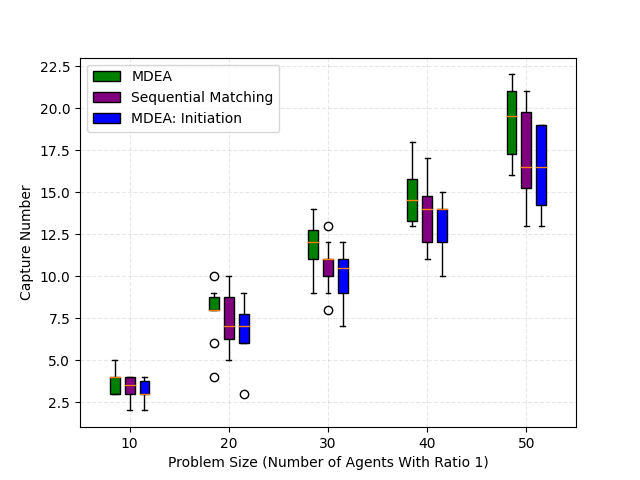}
	\label{fig:auto_1}}
	\subfigure[]{
	\includegraphics[width=1\linewidth]{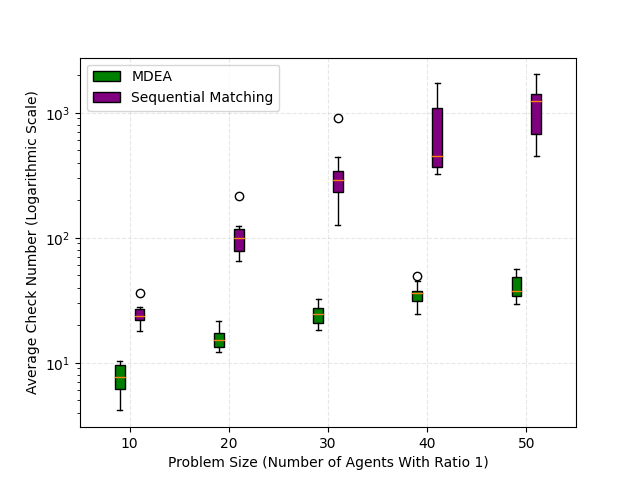}
	\label{fig:auto_2}}
	\caption{\footnotesize Performance comparison between our MDEA algorithm and the sequential matching approach in a 3D scenario. (a): Capture number comparison. (b): Average check number comparison.}
	\label{fig:boxplot_3d}
    \vspace{-5px}
\end{figure}

We next examine the MDEA algorithm for the multi-attack defense allocation problem. We compare our approach to two existing algorithms: the analytical barrier introduced in \cite{yan2019task} for a 2D scenario and the sequential matching presented in \cite{yan2022matching} for a 3D scenario. To conduct our experiments, we use five groups of agents, maintaining a 1:1 ratio between defenders and attackers. The number of agents in each group ranges from 10 to 50, and we perform simulations with 10 different initial joint states for each group. We employ the same domain as described in Section \ref{sec:V1}, with variations in the target sets for the different scenarios. Specifically, the 2D scenario has a rectangular domain of $[-5,5]^2$ with the line segment $y=5$ as the target set, while the 3D scenario has a box domain of $[-5,5]^3$ with the flat face $z=-5$ as the target set. 

During the simulation, we track two metrics to evaluate the effectiveness and computational efficiency of our proposed algorithm. The first metric is the capture number, representing the total number of attackers captured throughout the game, which is complementary to the payoff function. The second metric is the check number, denoting the count of convex programs of the form \eqref{co:d} computed within the multi-attack objective function defined in \eqref{fun:maof}. This metric plays a crucial role in computational complexity, particularly when the agent size increases, as witnessed by our empirical results. In these cases, it can be seen that our single-attack defense coordination strategy \eqref{ds:s} is the limiting case of that proposed in \cite{yan2019task} when the capture radii tend to zero, and it includes the one proposed in \cite{yan2022matching} as a special case. To ensure a fair comparison with other algorithms, we employ a low capture radius of $r_i=0.05\text{m}$ for the 2D scenario and $r_i=0.2\text{m}$ for the 3D scenario. Similarly, we adopt the attack strategy given in \eqref{as} and set the maximum speed ratios to 1, as described in Section \ref{sec:V1}.

We employ boxplots to display the results of our 2D simulation in Fig. \ref{fig:boxplot_2d}, indicating that our MDEA algorithm outperforms the analytical barrier approach. Three colors are used to represent three cases: 1) MDEA applied at each allocation time; 2) analytical barrier, which performs allocation only at the initial joint state; 3) MDEA implemented solely at the initial joint state, akin to the analytical barrier. As illustrated in Fig. \ref{fig:tro_1}, MDEA yields a considerable improvement over the analytical barrier, with its median values consistently exceeding those of the analytical barrier. Furthermore, the capture number achieved by MDEA surpasses that attained at the initial joint state, demonstrating the defense performance improvement of our approach as proven in Theorem \ref{thm:pgds}. In Fig. \ref{fig:tro_2}, it can be observed that the analytical barrier exhibits a constant check number for each case, equal to the dimension of the decision variables in the ILP \eqref{ilp}. Conversely, the check number of MDEA at the initial joint state is significantly lower than that of the analytical barrier, highlighting the superior computational efficiency of our method.

In the 3D simulation, our MDEA algorithm also outperforms the sequential matching method, as depicted in Fig. \ref{fig:boxplot_3d}. We employ the same three colors to represent three cases as in the 2D simulation, with the analytical barrier replaced by sequential matching. As illustrated in Fig. \ref{fig:auto_1}, MDEA surpasses the sequential matching with higher median values of capture number. Moreover, MDEA attains a final capture number that exceeds the expected capture number initially obtained by the algorithm, thereby demonstrating the effectiveness of our approach in enhancing defense performance. Besides, Fig. \ref{fig:auto_2} reveals that the average check number achieved by MDEA is significantly lower than that of the sequential matching, where the average check number is calculated by averaging the check number of each step. This showcases the computational efficiency of our approach.

\subsection{Gazebo Simulators with ROS}

\begin{figure*}[tp!]
	\centering
	\subfigure[]{
	\includegraphics[width=0.23\linewidth]{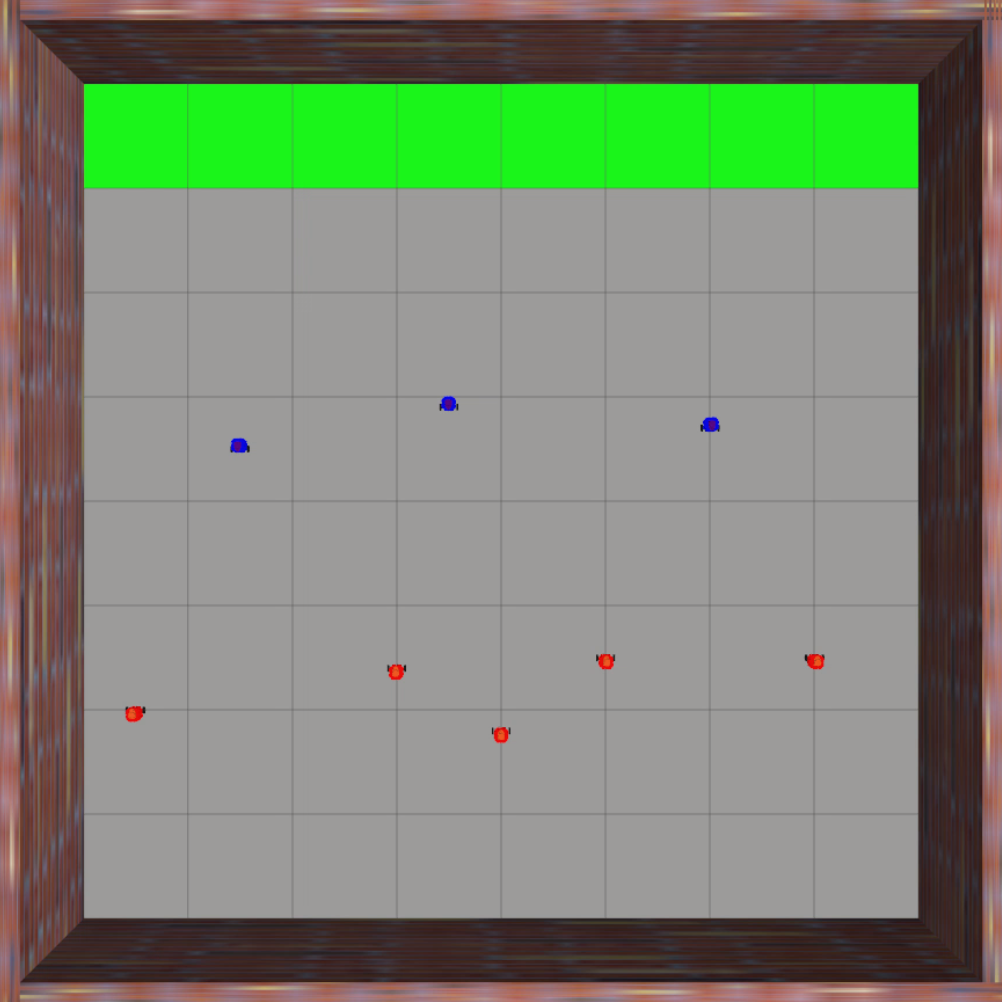}
	\label{fig:2d_aw_1}}
	\subfigure[]{
	\includegraphics[width=0.23\linewidth]{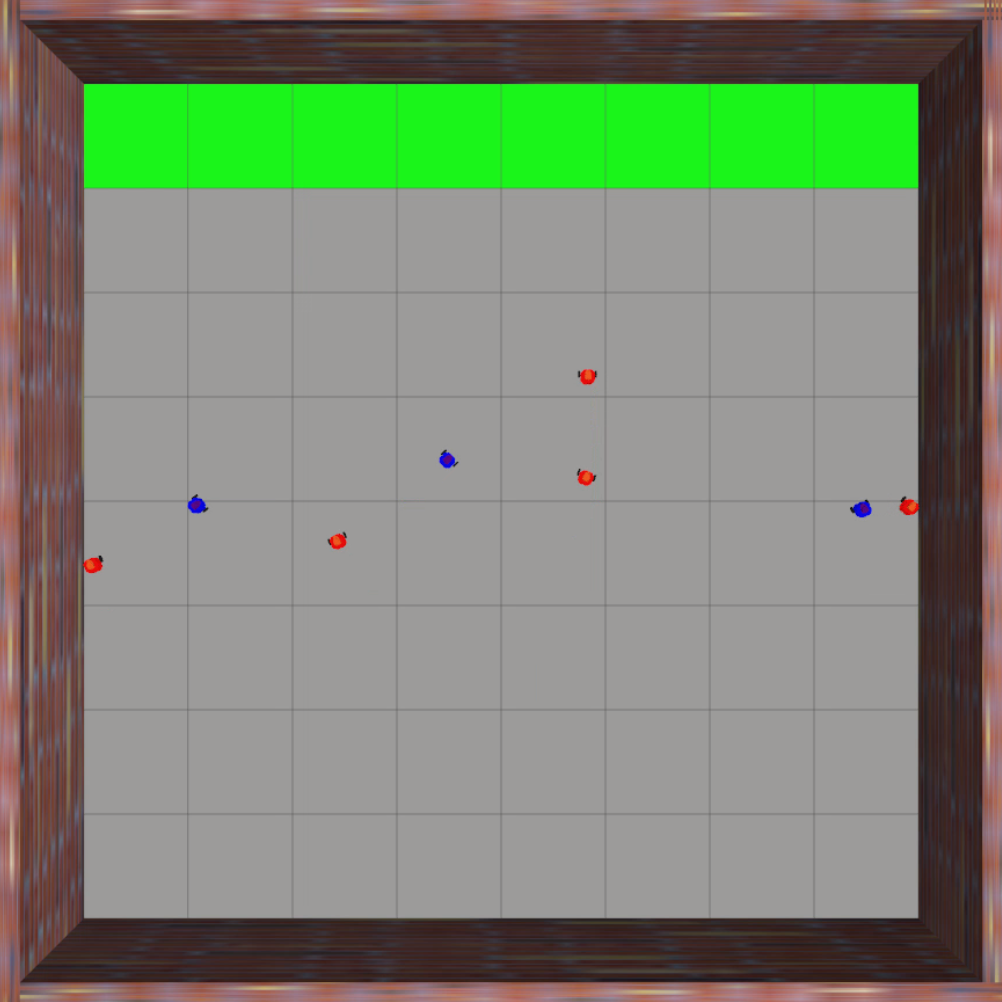}
	\label{fig:2d_aw_2}}
	\subfigure[]{
	\includegraphics[width=0.23\linewidth]{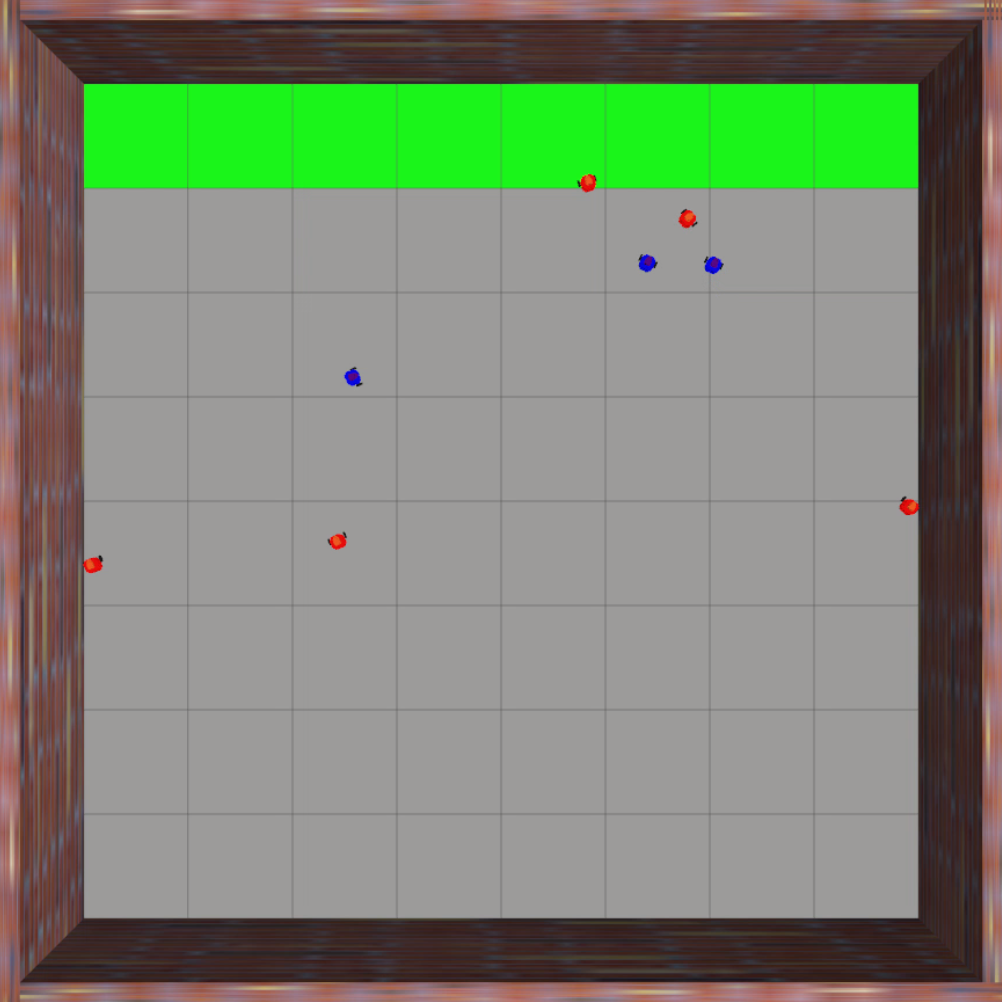}
	\label{fig:2d_aw_3}}
	\subfigure[]{
	\includegraphics[width=0.23\linewidth]{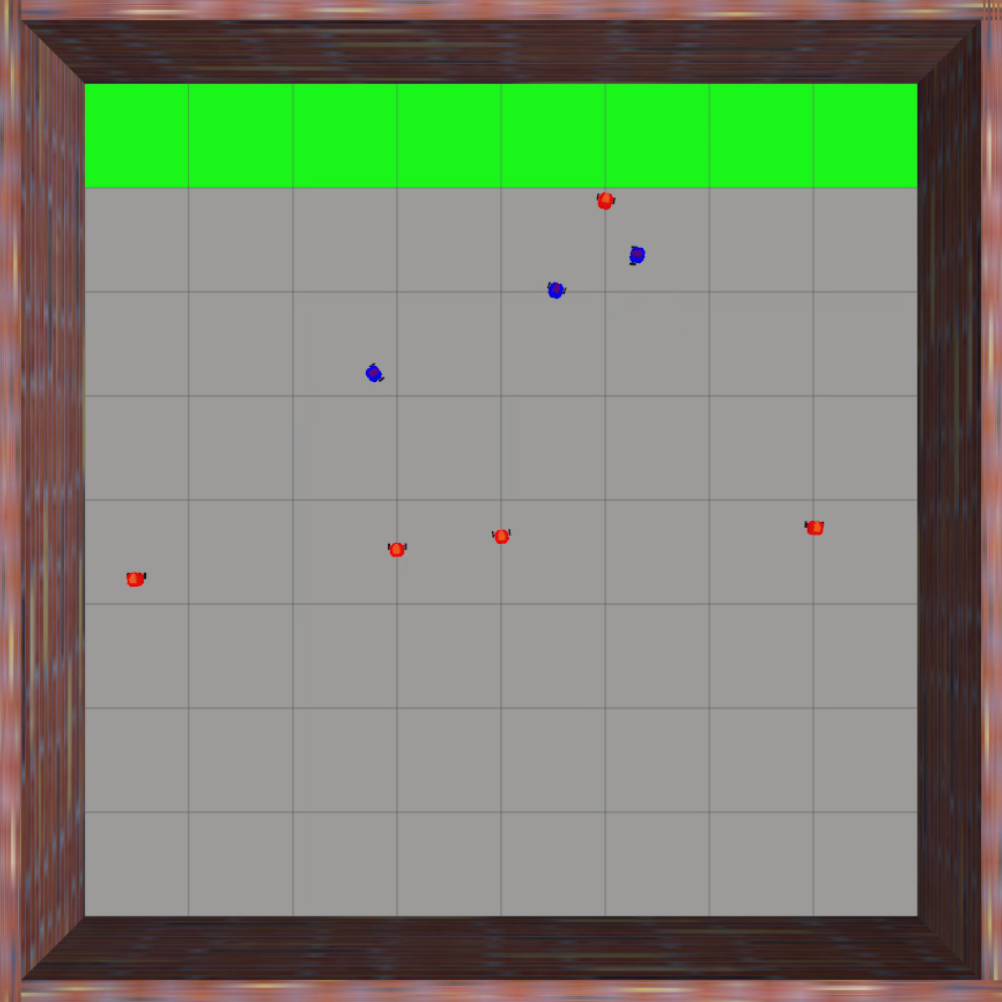}
	\label{fig:2d_aw_4}}
	\subfigure[]{
	\includegraphics[width=0.23\linewidth]{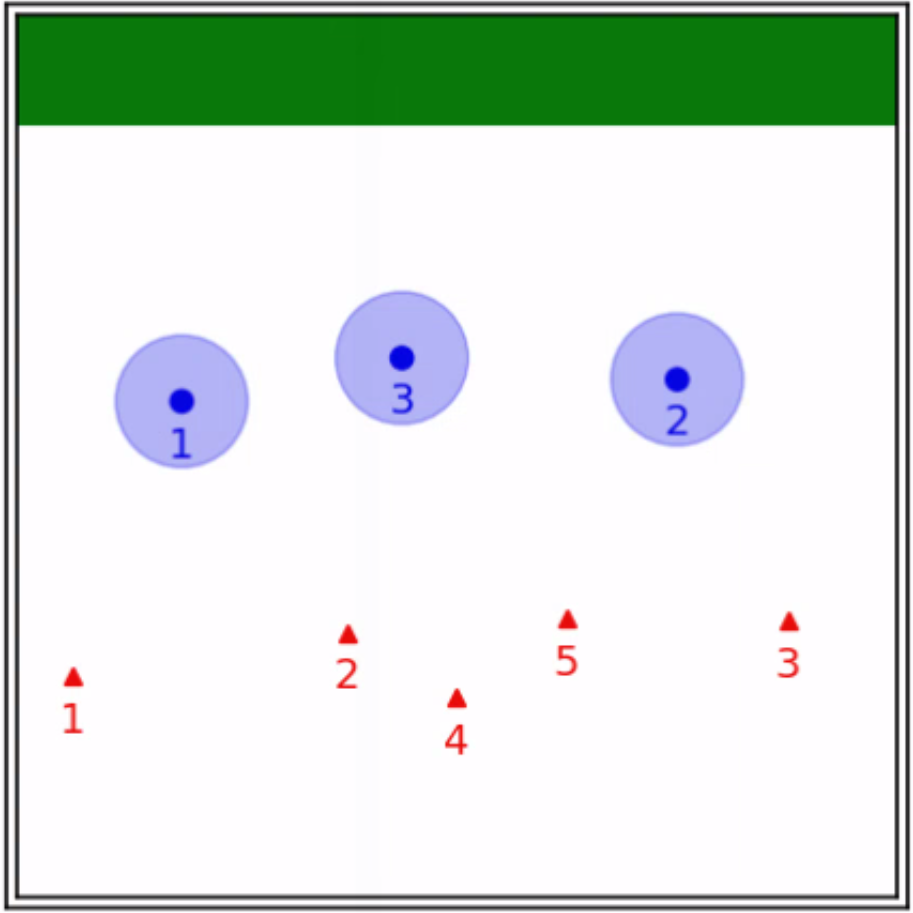}
	\label{fig:2d_dw_1}}
	\subfigure[]{
	\includegraphics[width=0.23\linewidth]{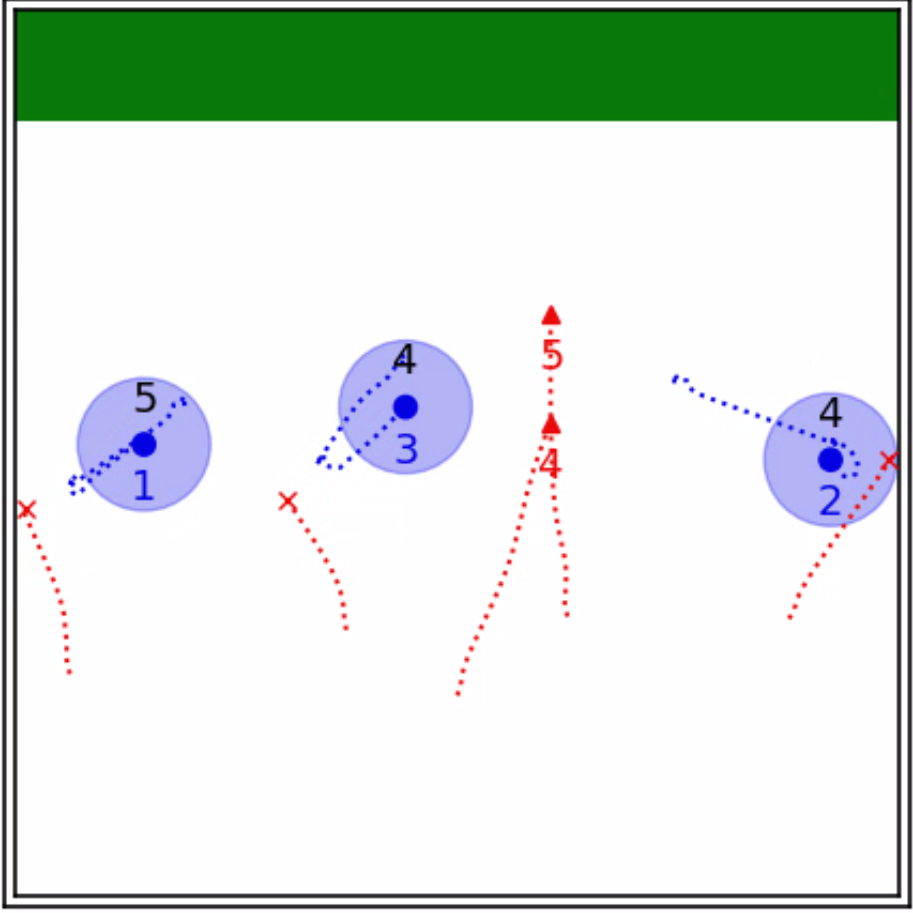}
	\label{fig:2d_dw_2}}
	\subfigure[]{
	\includegraphics[width=0.23\linewidth]{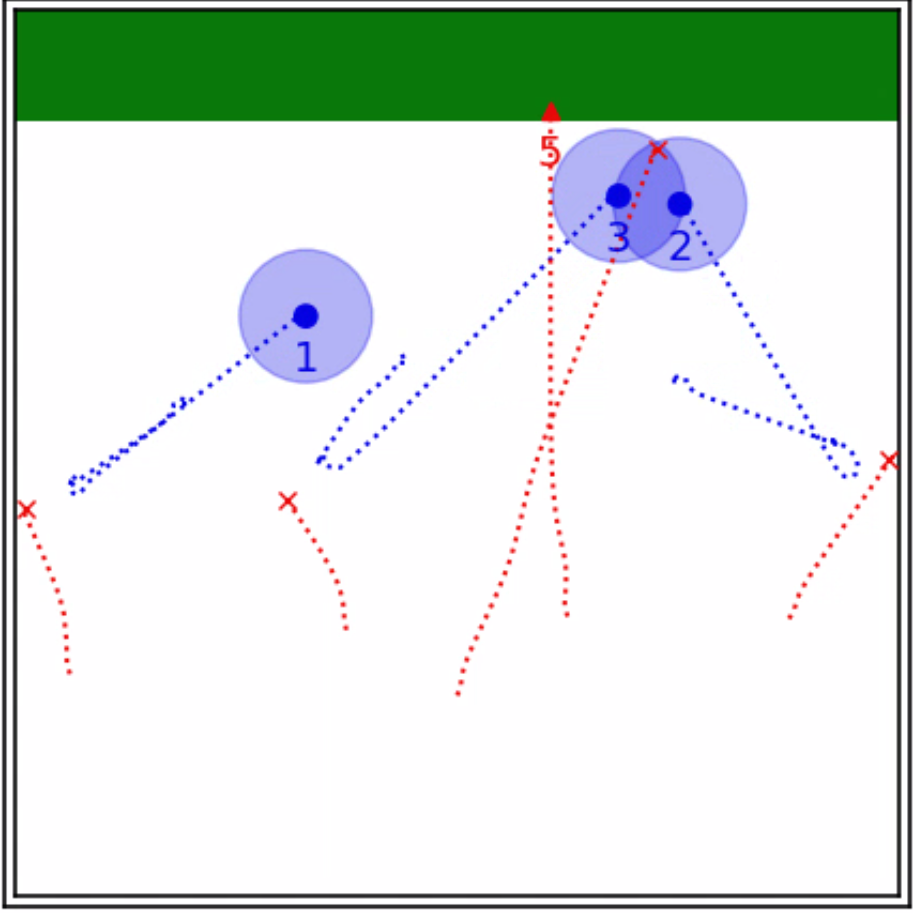}
	\label{fig:2d_dw_3}}
	\subfigure[]{
	\includegraphics[width=0.23\linewidth]{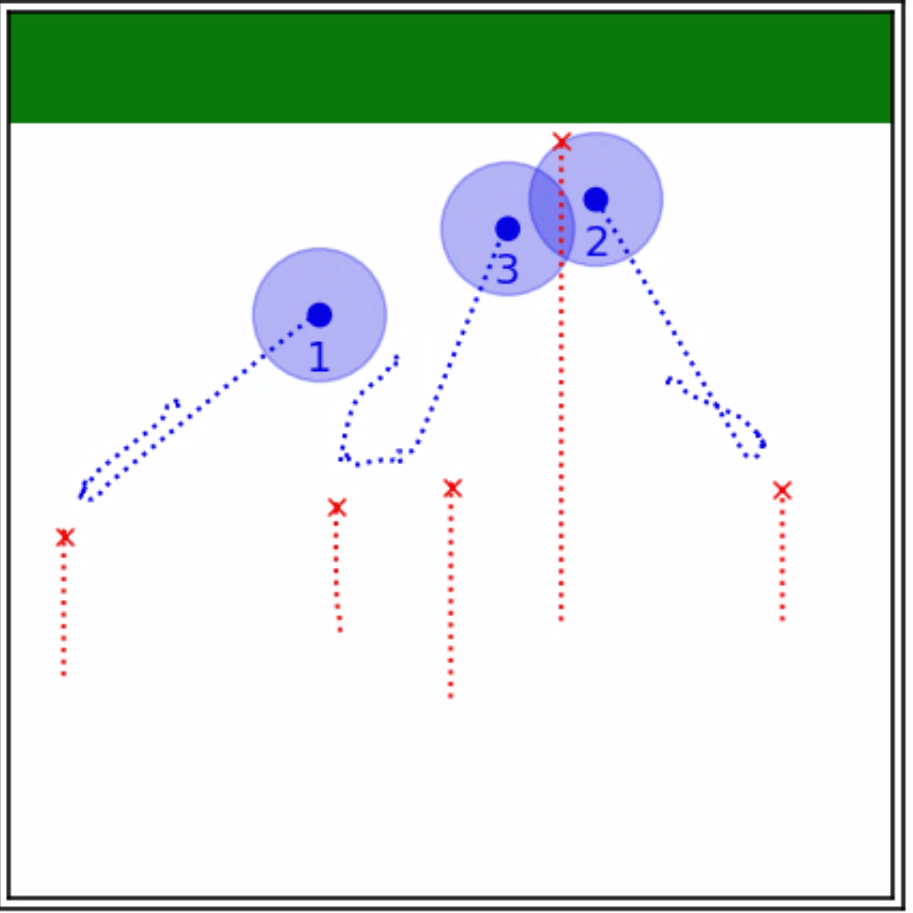}
	\label{fig:2d_dw_4}}
	\caption{\footnotesize 2D Gazebo simulation and visualization results for an indoor scenario with three defenders (blue) and five attackers (red). The green region denotes the target set, and the black numbers above the blue dots indicate the instantaneous assigned indices. (a) and (e) Initial configuration; (b) and (f) Configuration at 13s (optimal attack strategy case); (c) and (g) Final configuration (optimal attack strategy case); (d) and (h) Final configuration (straight-line attack strategy case).}
	\label{fig:2d}
	\vspace{-5px}
\end{figure*}

\begin{figure*}[tp!] 
	\centering
	\subfigure[]{
	\includegraphics[width=0.23\linewidth]{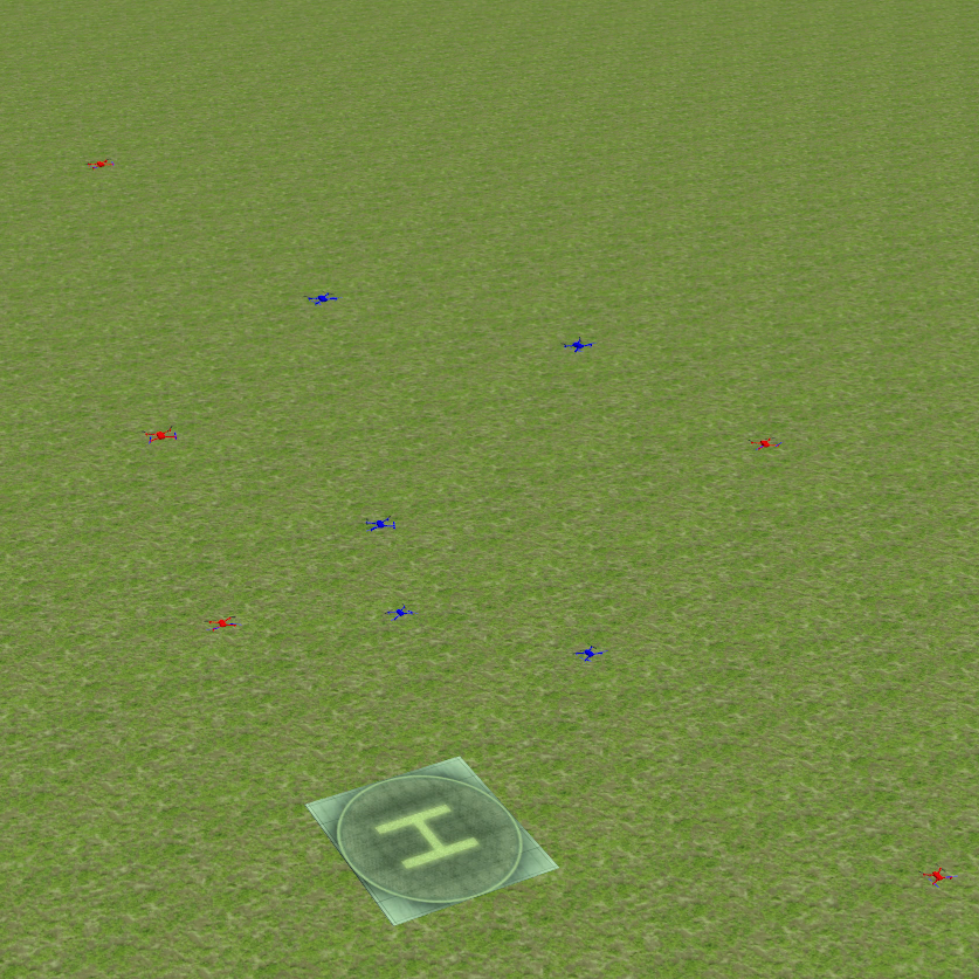}
	\label{fig:3d_aw_1}}
	\subfigure[]{
	\includegraphics[width=0.23\linewidth]{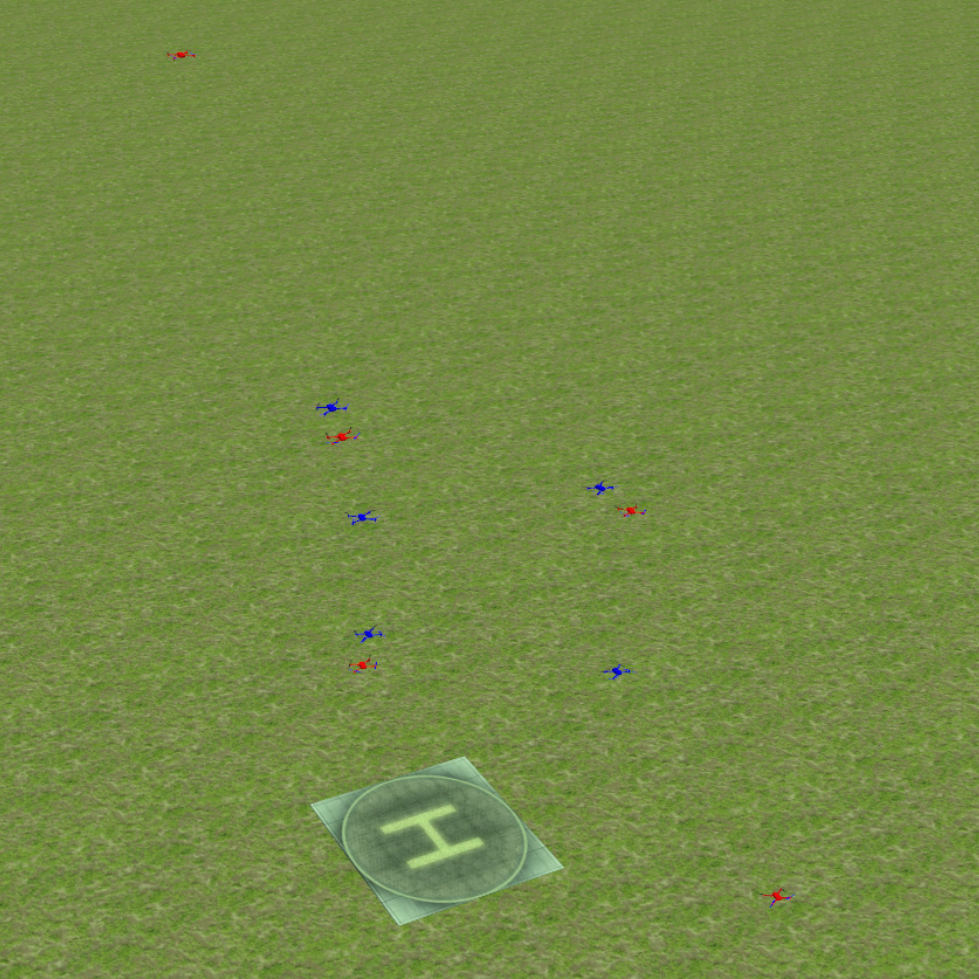}
	\label{fig:3d_aw_2}}
	\subfigure[]{
	\includegraphics[width=0.23\linewidth]{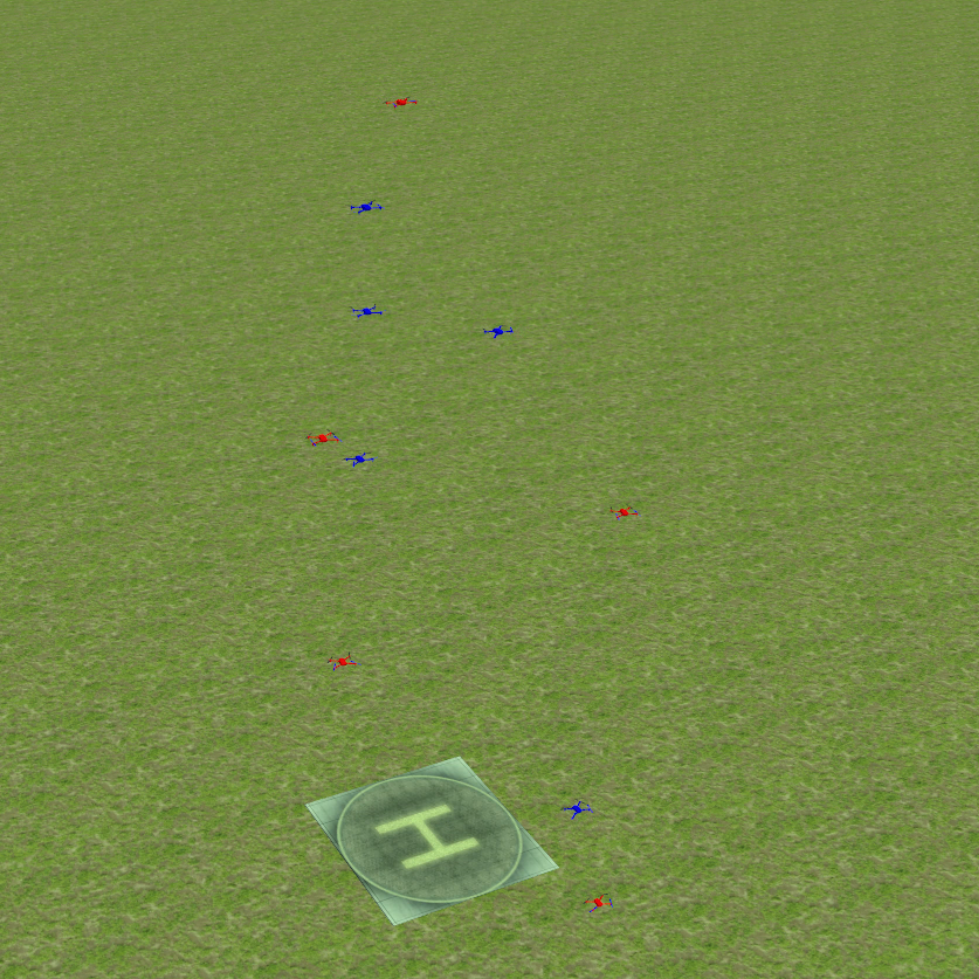}
	\label{fig:3d_aw_3}}
	\subfigure[]{
	\includegraphics[width=0.23\linewidth]{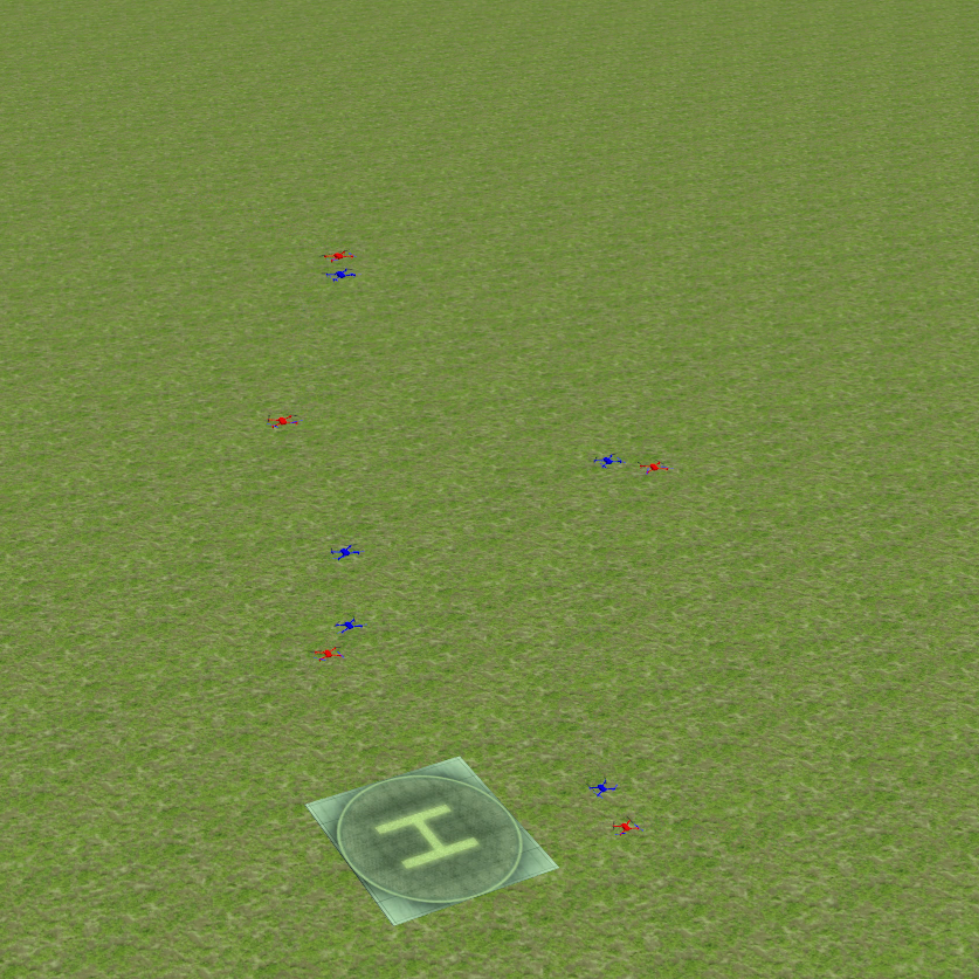}
	\label{fig:3d_aw_4}}
	\subfigure[]{
	\includegraphics[width=0.23\linewidth]{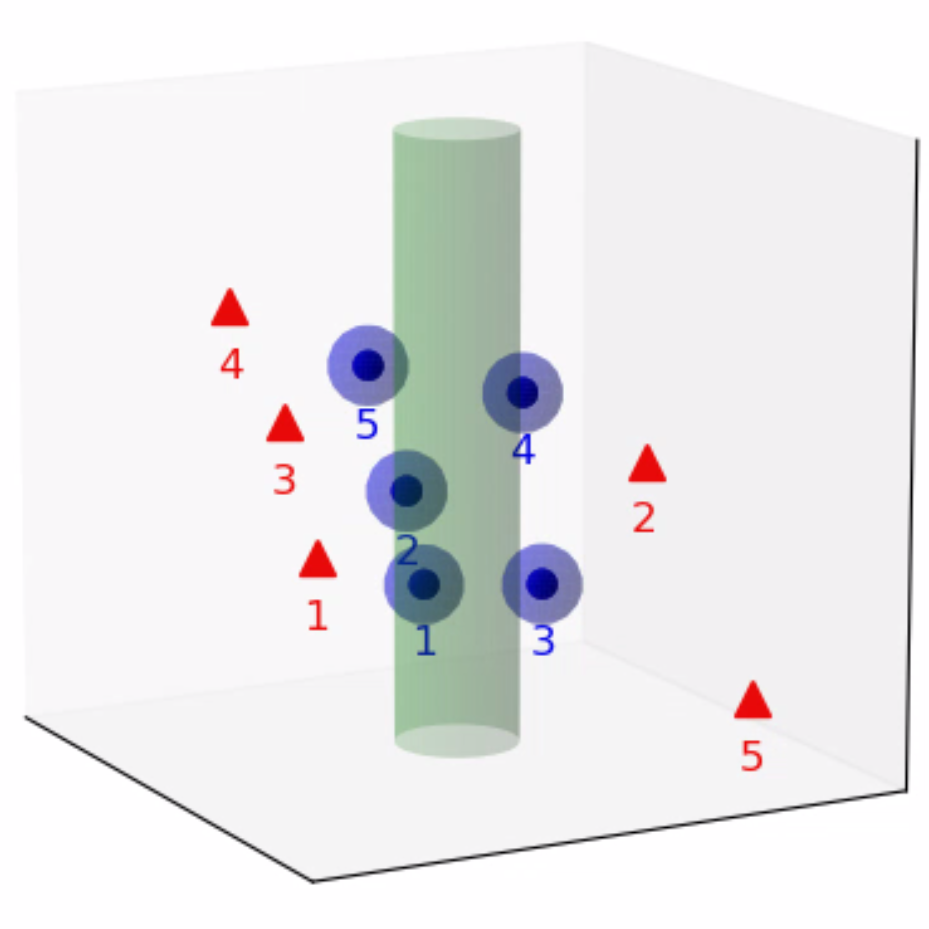}
	\label{fig:3d_dw_1}}
	\subfigure[]{
	\includegraphics[width=0.23\linewidth]{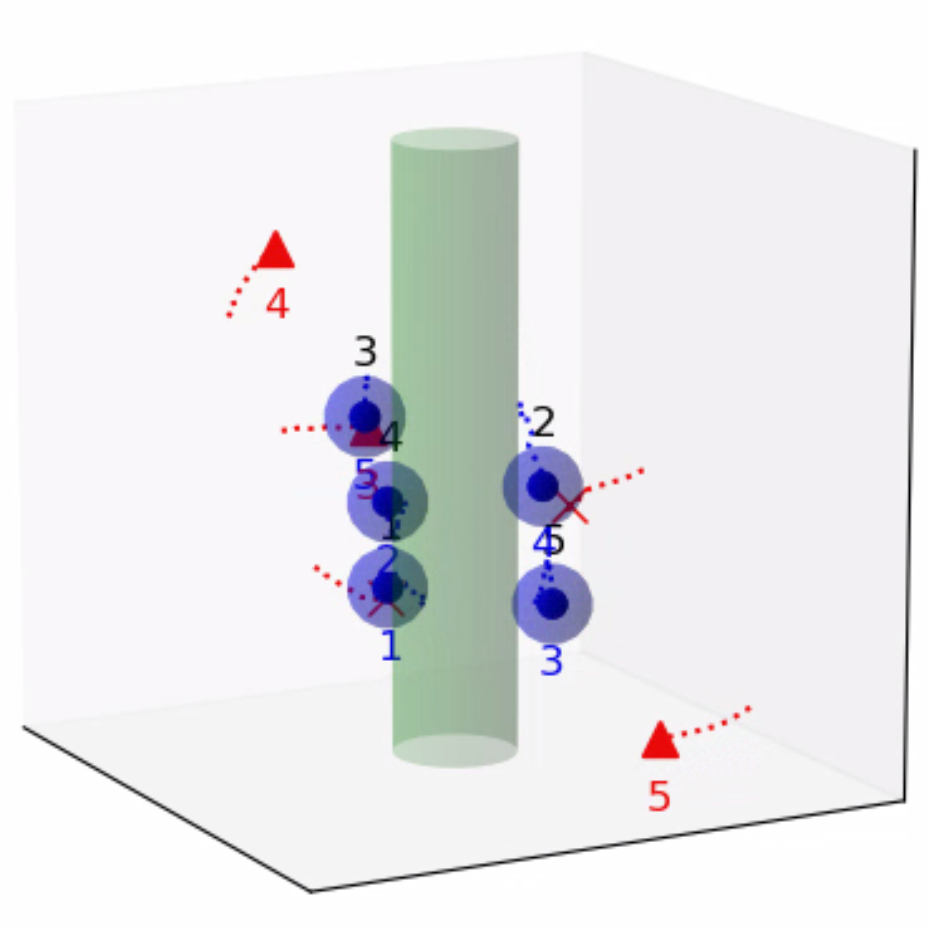}
	\label{fig:3d_dw_2}}
	\subfigure[]{
	\includegraphics[width=0.23\linewidth]{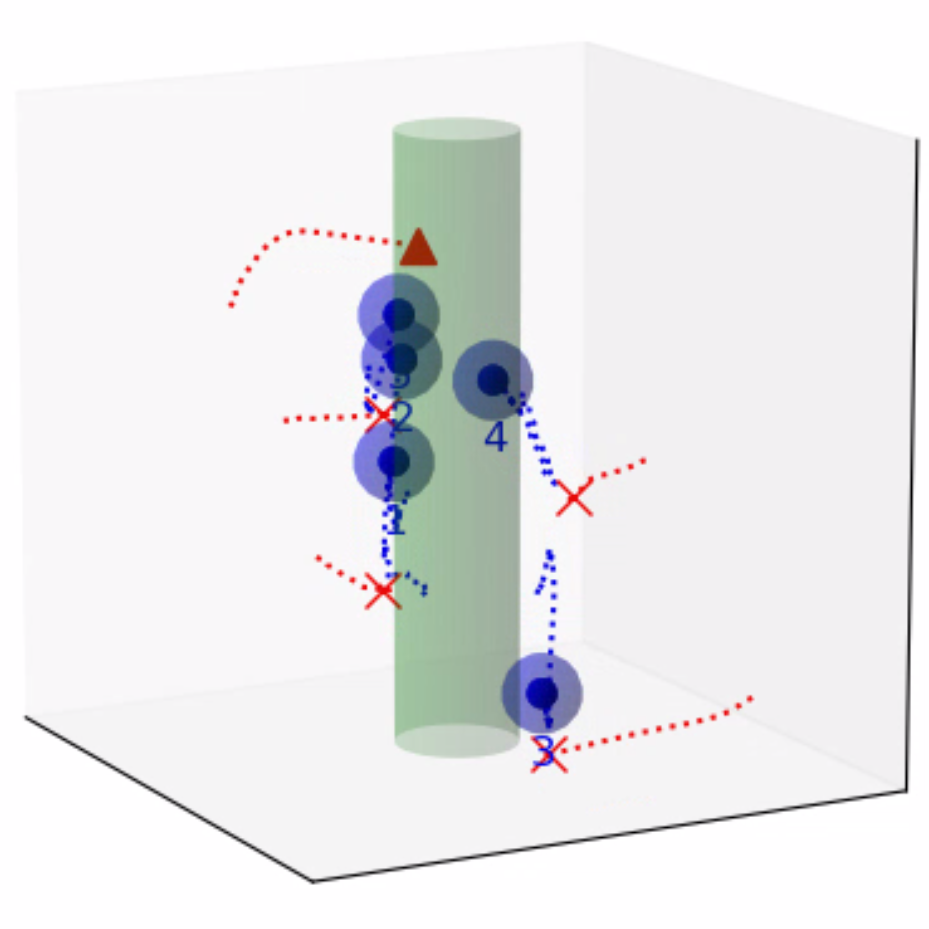}
	\label{fig:3d_dw_3}}
	\subfigure[]{
	\includegraphics[width=0.23\linewidth]{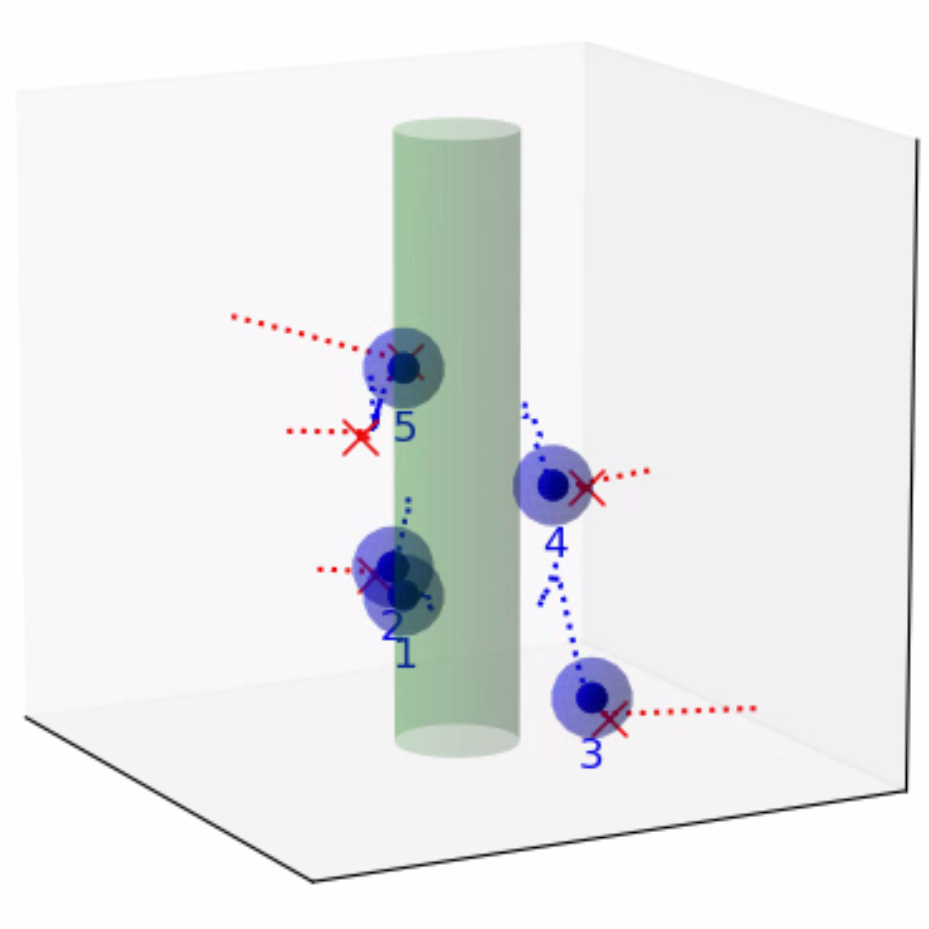}
	\label{fig:3d_dw_4}}
	\caption{\footnotesize 3D Gazebo simulation and visualization results for an outdoor scenario with five defenders (blue) and five attackers (red). The target set is represented by the cylindrical-shaped no-fly zone, as illustrated by the green cylinders in (e)-(h). The black numbers above the blue dots denote the instantaneous assigned indices. (a) and (e) Initial configuration; (b) and (f) Configuration at 5s (optimal attack strategy case); (c) and (g) Final configuration (optimal attack strategy case); (d) and (h) Final configuration (straight-line attack strategy case).}
	\label{fig:3d}
	\vspace{-5px}
\end{figure*}

Finally, we evaluate the practical applicability of our proposed algorithm using the Gazebo simulators in conjunction with ROS. To address real-world scenarios, we incorporate a multi-agent collision avoidance approach based on the control barrier function, as presented by Wang et al. \cite{wang2017safety}. Additionally, we utilize the mapping provided in \cite{lee2015multirobot} to convert the velocity vectors generated by our algorithm into corresponding linear and angular velocity commands. We conduct experiments in two different environments: a 2D indoor setting and a 3D outdoor setting. The indoor environment is designed as a square space surrounded by walls, with a rectangular region adjacent to one of the walls serving as the target set. In this scenario, TurtleBot3 Burger wheeled mobile robots are deployed as agents, featuring maximum linear and angular velocities of 0.22$\text{m/s}$ and 2.84$\text{rad/s}$, respectively. On the other hand, the outdoor environment is characterized by a box-shaped region elevated above the ground, with a cylindrical region designated as the target set to represent a no-fly zone in real-world scenarios. Here, Iris quadcopter robots are employed as agents. The computed velocity vector is utilized as a reference point to regulate the robot's velocity and heading. The linear and angular velocity references are set to 0.8$\text{m/s}$ and 0$\text{rad/s}$, respectively, enabling the Iris quadcopter robots to move swiftly towards the reference point while preserving safety. The safety radii for the TurtleBot3 Burger and Iris robots are 0.1$\text{m}$ and 0.5$\text{m}$, respectively, while their capture radii are set to 0.6$\text{m}$ and 1.5$\text{m}$, respectively. Both robotics platforms operate at a frequency of 50$\text{Hz}$, with a defense allocation frequency of 10$\text{Hz}$.

To test the effectiveness of our proposed algorithm, we use two different attack strategies and examine the defender team's adaptability in response to these variations. The first strategy is the optimal attack strategy given in \eqref{as}, where attacker $j$ moves towards the almost-optimal waypoint $\tilde{\xi}^{\N}_j$, as specified in \eqref{owp}. The second strategy termed the straight-line approach, entails each attacker heading toward the nearest point on the target set, with the objective of minimizing the time required to reach the target set while disregarding the risk of potential capture by the defenders. The simulation process can be seen in the attached video, and illustrative snapshots are provided in Fig. \ref{fig:2d} and Fig. \ref{fig:3d}. In the 2D indoor scenario, featuring a configuration of five attackers and three defenders, our proposed algorithm exhibits a high level of effectiveness, achieving a final capture number of four for the optimal attack strategy and five for the straight-line attack strategy. Likewise, in the 3D outdoor scenario, comprising five attackers and five defenders, the algorithm demonstrates its robustness by attaining a final capture number of four for the optimal attack strategy and five for the straight-line attack strategy. These results substantiate the effectiveness of our proposed algorithm in diverse environmental settings.

\section{Conclusion}

In this article, we presented a dual-layer online optimization strategy for defenders in multiplayer reach-avoid games, with the goal of minimizing the number of successful attackers entering a convex target set within a convex domain while facing unknown attack strategies. We devised a dual-mode switching defense algorithm for defender coalitions to counter individual attackers via online convex programming. This algorithm induces the maximum defense-winning region for initial joint states, sharpening the chances of winning while offering scalability. Additionally, we developed a monotonic defense enhancement allocation algorithm for determining coalition assignment matrices in integer linear programs (ILPs) by incorporating predicted outcomes from single-attack coordination. This algorithm employs a hierarchical iterative method to approximate ILPs and includes a monotonicity constraint, reducing the chosen coalition-attacker pairs while maintaining a non-decreasing expected number of unsuccessful attackers over time. We assessed our approach's performance across diverse multi-robot systems and environments, demonstrating its superiority in terms of optimality and efficiency. Moreover, Gazebo simulations confirmed the real-world applicability and potential for practical robotics implementation. In future work, extending our approach to encompass non-convex environments and integrating more complex agent models could enhance the performance and applicability of our proposed framework. In addition, accounting for attackers with different priorities may better reflect real-world scenarios and further refine the defense strategies employed in our approach.

\appendix

\subsection{Proof of Proposition \ref{prop:ssr}}\label{proof:ssr}
\begin{Proof}
To show the convexity of the SRS, we first exhibit that for each $i\in \C$, $c_{ij}(q,x_{ij})$ is convex with respect to $q$. Note that $c_{ij}$ can be decomposed into a sum of three terms: $c_{ij}(q,x_{ij})=(\gamma_{ij}^2-1)||q||^2+2r_i\gamma_{ij}||q-p^a_j||+[2(p^d_i-\gamma_{ij}^2p^a_j)^Tq+\gamma_{ij}^2||p^a_j||^2-||p^d_i||^2+r_i^2]$.
Owing to the maximum speed ratio constraint \eqref{con:ratio} and the positivity of $r_i$, the coefficients of the first two terms are non-negative. Consequently, these two terms are convex, as the (squared) Euclidean norm is a convex function. The convexity of $c_{ij}(q,x_{ij})$ then follows from the fact that the last term is linear, and the sum of convex functions is itself a convex function.

It remains to establish that if $q\in \D$ satisfies the inequality $\max_{i\in \C}c_{ij}(q,x_{ij})\leq 0$, then $q$ belongs to the SRS. Given any such $q$, consider the constant attack strategy $\pi^a_j=v^a_{j,\max}\n(q-p^a_j)$ over the time interval $[0,\tau]$, where $\tau=\frac{||q-p^a_j||}{v^a_{j,\max}}$.
The reachability property of the SRS holds by noting that attacker $j$ can reach $q$ from $p^a_j$ along the line segment $L$ connecting $p^a_j$ and $q$ using $\pi^a_j$. 
On the other hand, any point on $L$ can be expressed as $q_\lambda=(1-\lambda)p^a_j+\lambda q$ with parameter $\lambda\in [0,1]$, which is reached by attacker $j$ at time $\lambda\tau$ using $\pi^a_j$. 
Under the same time, the closest point to $q_\lambda$ that defender $i$ can reach from $p^d_i$, denoted by $q^{d*}_{i,\lambda}$, must lie on the straight line that connects $p^d_i$ and $q_\lambda$, as defender $i$ can move in any direction. This leads to $q^{d*}_{i,\lambda}=p^d_i+v^d_{i,\max}\frac{q_\lambda-p^d_i}{||q_\lambda-p^d_i||}\lambda\tau$. Besides, the convexity of $c_{ij}(q,x_{ij})$ implies that $c_{ij}(q_{\lambda},x_{ij})\leq (1-\lambda)c_{ij}(p^a_j,x_{ij})+\lambda c_{ij}(q,x_{ij})\leq 0$, which in turn indicates that $s_i(q^{d*}_{i,\lambda},q_\lambda)=r_i-\bigl\lvert ||q_\lambda-p^d_i||-\gamma_{ij}||q_\lambda-p^a_j||\bigr\rvert=\gamma_{ij}||q_\lambda-p^a_j||+r_i-||q_\lambda-p^d_i||\leq 0$.
As a result, for any admissible defense strategy $\pi^d_i$,
$s_i(\chi^d_i(\lambda\tau;p^d_i,\pi^d_i),q_\lambda)\leq s_i(q^{d*}_{i,\lambda},q_\lambda)\leq0$ for all $\lambda\in [0,1]$, yielding the safety property of the SRS.
\end{Proof}

\subsection{Proofs of Proposition \ref{prop:gc} and \eqref{eq:gc_a}}\label{proof:gc}

\begin{Proof}
Consider the Lagrangian function of \eqref{co:d}:
\begin{equation*}
	\begin{split}
		\L^d(\hat{q},\hat{\lambda})=&||q-\tilde{q}||^2+\sum_{i\in \C}\lambda_{ij} c_{ij}(q,x_{ij})\\
		&+\sum_{k\in\I_D}\bar{\lambda}_{k}d_k(q)+\sum_{l\in\I_G}\tilde{\lambda}_{l}g_l(\tilde{q})
	\end{split}
\end{equation*}
where $\hat{q}=(q,\tilde{q})$ and $\hat{\lambda}=(\lambda_{ij},i\in \C;\bar{\lambda}_k,k\in \I_D;\tilde{\lambda}_l,l\in\I_G)$ is the Lagrange multiplier. The KKT conditions asserts that for the optimal solution $\hat{q}^*=(\xi^\C_j,\tilde{\xi}^\C_j)$, there is a Lagrange multiplier $\hat{\lambda}^*$ with nonnegative components such that $\frac{\partial\L^d}{\partial \hat{q}}(\hat{q}^*,\hat{\lambda}^*)=0$, i.e., 
\begin{align}
	&2(\xi^\C_j-\tilde{\xi}^\C_j)^T+\sum_{i\in \C}\lambda_{ij}^*\frac{\partial c_{ij}}{\partial q}(\xi^\C_j,x_{ij})+\sum_{k\in\I_D}\bar{\lambda}_k^*\frac{\partial d_k}{\partial q}(\xi^\C_j)=0\label{eq:d1}\\
	&2(\xi^\C_j-\tilde{\xi}^\C_j)^T-\sum_{l\in\I_G}\tilde{\lambda}_{l}^*\frac{\partial g_l}{\partial \tilde{q}}(\tilde{\xi}^\C_j)=0.\label{eq:d2}
\end{align}
Additionally, for any interior point $x^\C_j\in \D'$ and any $i\in \C$, one of the following three cases occurs: 
i) $c_{ij}(\xi^\C_j,x_{ij})<0$; ii) $c_{ij}(\xi^\C_j,x_{ij})=0$ and $\dot{c}_{ij}(\xi^\C_j,x_{ij})=0$; iii) $c_{ij}(\xi^\C_j,x_{ij})=0$ and $\dot{c}_{ij}(\xi^\C_j,x_{ij})\neq 0$. In particular, since $c_{ij}$ is continuously differentiable, 
the set of interior points for which the last case holds has measure zero, according to the results from real analysis \cite{royden1988real}. Thus, the complementarity condition $\lambda_{ij}^*c_{ij}(\xi^\C_j,x_{ij})=0$ implies that for almost every $x^\C_j\in \D'$, $\lambda_{ij}^*\dot{c}_{ij}(\xi^\C_j,x_{ij})=0$, i.e., 
\begin{equation}\label{eq:d3}
	\begin{split}
		&\lambda_{ij}^*\left(\frac{\partial c_{ij}}{\partial q}(\xi^\C_j,x_{ij})\dot{\xi}^\C_j+\frac{\partial c_{ij}}{\partial p^d_i}(\xi^\C_j,x_{ij})u^d_i\right.\\
		&\left.+\frac{\partial c_{ij}}{\partial p^a_j}(\xi^\C_j,x_{ij})u^a_j\right)=0,~ \forall i\in \C.
	\end{split}
\end{equation}
Similar derivations can be respectively employed to the complementarity conditions $\bar{\lambda}_k^*d_k(\xi^\C_j)=0$, $k\in \I_D$ and $\tilde{\lambda}_l^*g_l(\tilde{\xi}^\C_j)=0$, $l\in\I_G$, yielding that for almost every $x^\C_j\in \D'$,
\begin{equation}\label{eq:d4}
	\bar{\lambda}_k^*\frac{\partial d_k}{\partial q}(\xi^\C_j)\dot{\xi}^\C_j=0,~\forall k\in \I_D
\end{equation}
and $\tilde{\lambda}_l^*\frac{\partial g_l}{\partial \tilde{q}} (\tilde{\xi}^\C_j)\dot{\tilde{\xi}}^\C_j=0$ for all $l\in \I_G$, where the latter equation together with \eqref{eq:d2} leads to
\begin{equation}\label{eq:d5}
	2(\xi^\C_j-\tilde{\xi}^\C_j)^T\dot{\tilde{\xi}}^\C_j=\sum_{l\in\I_G}\tilde{\lambda}_l^*\frac{\partial g_l}{\partial \tilde{q}}(\tilde{\xi}^\C_j)\dot{\tilde{\xi}}^\C_j=0.
\end{equation}
Therefore, 
\begin{equation*}
	\begin{split}
		\dot{\Phi}^\C_j\overset{\mathrm{def}}{=}&2(\xi^\C_j-\tilde{\xi}^\C_j)^T(\dot{\xi^\C_j}-\dot{\tilde{\xi}}^\C_j)\\
		\overset{\eqref{eq:d5}}{=}&2(\xi^\C_j-\tilde{\xi}^\C_j)^T\dot{\xi^\C_j}\\
		\overset{\eqref{eq:d1}}{=}&-\sum_{i\in \C}\lambda_{ij}^*\frac{\partial c_{ij}}{\partial q}(\xi^\C_j,x_{ij})\dot{\xi}^\C_j-\sum_{k\in\I_D}\bar{\lambda}_k^*\frac{\partial d_k}{\partial q}(\xi^\C_j)\dot{\xi}^\C_j\\
		\overset{\eqref{eq:d3}+\eqref{eq:d4}}{=}&\sum_{i\in \C}\lambda_{ij}^*\frac{\partial c_{ij}}{\partial p^d_i}(\xi^\C_j,x_{ij})u^d_i+\sum_{i\in \C}\lambda_{ij}^*\frac{\partial c_{ij}}{\partial p^a_j}(\xi^\C_j,x_{ij})u^a_j
	\end{split}
\end{equation*}
holds for almost every $x^\C_j\in \D'$. This completes the proof of Proposition \ref{prop:gc} by examining \eqref{eq:deriv_d}.

To verify the equality \eqref{eq:gc_a}, we consider the Lagrangian function of \eqref{co:a}:
\begin{equation*}
	\L^a(q,\hat{\mu})=||q-p^a_j||^2+\sum_{i\in \C}\mu_{ij}c_{ij}(q,x_{ij})+\sum_{l\in\I_G}\tilde{\mu}_{l}g_l(q)
\end{equation*}
with the Lagrange multiplier $\hat{\mu}=(\mu_{ij},i\in \C,\tilde{\mu}_l,l\in\I_G)$. 
Analogous to the proof of Proposition \ref{prop:gc}, the KKT conditions imply that there is a Lagrange multiplier $\hat{\mu}^*$ with nonnegative components such that 
\begin{equation}\label{eq:a1}
    2(\bar{\xi}^\C_j-p^a_j)^T+\sum_{i\in \C}\mu_{ij}^*\frac{\partial c_{ij}}{\partial q}(\bar{\xi}^\C_j,x_{ij})
	+\sum_{l\in \I_G}\tilde{\mu}_l^*\frac{\partial g_l}{\partial q}(\bar{\xi}^\C_j)=0
\end{equation}
and for almost every $x^\C_j\notin \D^\C_j$,
\begin{equation}\label{eq:a2}
    \begin{split}
        &\mu_{ij}^*\left(\frac{\partial c_{ij}}{\partial q}(\bar{\xi}^\C_j,x_{ij})\dot{\bar{\xi}}^\C_j+\frac{\partial c_{ij}}{\partial p^d_i}(\bar{\xi}^\C_j,x_{ij})u^d_i\right.\\
        &\left.+\frac{\partial c_{ij}}{\partial p^a_j}(\bar{\xi}^\C_j,x_{ij})u^a_j\right)=0,~\forall i\in \C
    \end{split}
\end{equation}
and
\begin{equation}\label{eq:a3}
	\tilde{\mu}_l^*\frac{\partial g_l}{\partial q}(\bar{\xi}^\C_j)\dot{\bar{\xi}}^\C_j=0,~\forall l\in \I_G.
\end{equation}
Consequently, the proof is done by noting that
\begin{equation*}
	\begin{split}
		&\dot{\bar{\Phi}}^\C_j+2(\bar{\xi}^\C_j-p^a_j)^Tu^a_j\\
		\overset{\mathrm{def}}{=}&2(\bar{\xi}^\C_j-p^a_j)^T\dot{\bar{\xi}}^\C_j\\
		\overset{\eqref{eq:a1}}{=}&-\sum_{i\in \C}\mu_{ij}^*\frac{\partial c_{ij}}{\partial q}(\bar{\xi}^\C_j,x_{ij})\dot{\bar{\xi}}^\C_j-\sum_{l\in \I_G}\tilde{\mu}_l^*\frac{\partial g_l}{\partial q}(\bar{\xi}^\C_j)\dot{\bar{\xi}}^\C_j\\
		\overset{\eqref{eq:a2}+\eqref{eq:a3}}{=}&\sum_{i\in \C}\mu_{ij}^*\frac{\partial c_{ij}}{\partial p^d_i}(\bar{\xi}^\C_j,x_{ij}) u^d_i+\sum_{i\in \C}\mu_{ij}^*\frac{\partial c_{ij}}{\partial p^a_j}(\bar{\xi}^\C_j,x_{ij})u^a_j.
	\end{split}
\end{equation*}
\end{Proof}

\subsection{Proof of Proposition \ref{prop:unique}}\label{proof:unique}
\begin{Proof}
The proof is conducted by contradiction. Suppose there are $q_1,q_2\in \Omega^\C_j(x^\C_j)$ such that $\Psi(q_1)=\Psi(q_2)$, where
\begin{equation}\label{fun:psi}
	\Psi(q)=\min_{\tilde{q}\in \G}||q-\tilde{q}||^2
\end{equation}
is the minimum squared distance from $q$ to the target set. We claim that the line segment connecting $q_1$ and $q_2$, denoted as $q_\lambda=\lambda q_1+(1-\lambda)q_2$ with parameter $\lambda\in [0,1]$, must lie on the zero level set defined by $c_{kj}(q,x_{kj})=0$ for some $k\in \N$. First, note that the line segment $q_\lambda$ must lie within the SRS due to its convexity. Next, by the Definition of $\Phi^\C_j$, we know that $\Phi^\C_j(x^\C_j)\leq \Psi(q)$ for all $q\in \Omega^\C_j(x^\C_j)$, and thus $\Phi^\C_j(x^\C_j)\leq \Psi(q_\lambda)$ for all $\lambda\in [0,1]$. Furthermore, since $\Psi$ is convex, we have $\Psi(q_\lambda)\leq \lambda \Psi(q_1)+(1-\lambda)\Psi(q_2)=\Phi^\C_j(x^\C_j)$. Therefore, $\Psi(q_\lambda)=\Phi^\C_j(x^\C_j)$ for all $\lambda\in [0,1]$, indicating that the line segment $q_\lambda$ is on the boundary of the SRS. Lastly, we verify the claim by noting that if $\max_{i\in \N}c_{ij}(q_{\lambda_0},x_{ij})<0$ for some $\lambda_0\in [0,1]$, then the point $q_{\lambda_0}$ lies strictly within the SRS, which would contradict the result that $q_{\lambda_0}$ lies on the boundary of the SRS. However, the claim violates the fact that the zero level set defined by $c_{kj}(q,x_{kj})=0$ contains no line segments. This is because $(q_1-q_2)^T\nabla_q c_{kj}=0$ can never have infinite solutions, where $\nabla_q c_{kj}=2(\gamma_{kj}^2-1)q+2r_k\gamma_{kj}\frac{q-p^a_j}{||q-p^a_j||}+2(p^d_k-\gamma_{kj}^2p^a_j)$ is the normal vector to the zero level set of $c_{kj}(q,x_{kj})=0$.
\end{Proof}

\subsection{Proof of Proposition \ref{prop:f_ads}}\label{proof:f_ads}
\begin{Proof}
It suffices to show by contradiction that $\Phi^{\C'}_j(x^{\C'}_j)=\Phi^{\C}_j(x^{\C}_j)$ with $\C'=\Lambda^\C_j$, or equivalently $\Psi(\xi^{\C'}_j)=\Psi(\xi^{\C}_j)$ in which $\Psi$ is defined in \eqref{fun:psi}. Assume that $\Psi(\xi^{\C'}_j)\neq\Psi(\xi^{\C}_j)$. In this case, we have $\Psi(\xi^{\C'}_j)<\Psi(\xi^{\C}_j)$ as $\xi^{\C}_j\in \Omega^{\C'}_j$. To reach a contradiction, we consider the line segment connecting $\xi^{\C'}_j$ and $\xi^{\C}_j$, which is parameterized by $\mu\in [0,1]$ as $\xi_{\mu}=\mu\xi^{\C'}_j+(1-\mu)\xi^{\C}_j$. Since $\C'$ is the ADS for $(\C,j)$, it follows that $c_{ij}(\xi^{\C}_j,x_{ij})<0$ for all $i\in \C\setminus\C'$. Due to the continuity of $c_{ij}(\xi_{\mu},x_{ij})$ with respect to $\mu$, there exists $\mu_i\in (0,1)$ for each $i\in \C\setminus\C'$ such that $c_{ij}(\xi_{\mu},x_{ij})\leq 0$ for all $\mu\in [0,\mu_i,]$. Let $\bar{\mu}$ be the minimum of all $\mu_i$. Then $c_{ij}(\xi_{\bar{\mu}},x_{ij})\leq 0$ for all $i\in \C\setminus\C'$. Moreover, since the line segment $\xi_{\mu}$ lies within $\Omega^{\C'}_j(x^{\C'}_j)$, we have $c_{ij}(\xi_{\bar{\mu}},x_{ij})\leq 0$ for all $i\in \C'$. As a result, $\xi_{\bar{\mu}}\in \Omega^{\C}_j$. However, it is noted that $\Psi(\xi_{\bar{\mu}})\leq \bar{\mu}\Psi(\xi^{\C}_j)+(1-\bar{\mu})\Psi(\xi^{\C'}_j)<\Psi(\xi^{\C}_j)$, which contradicts the fact that the minimum value of $\Psi$ occurs at $\xi^{\C}_j$.
\end{Proof}

\subsection{Proof of Proposition \ref{prop:ub}}\label{proof:ub}
\begin{Proof}
According to Proposition \ref{prop:f_ads}, it is sufficient to establish that for any feasible coalition-attacker pair $(\C,j)$ with $|\Lambda^\C_j|>n$, there exists a subset $\C'$ of $\Lambda^\C_j$ with $|\C'|\leq n$ such that $\Phi^{\C'}_j(x^{\C'}_j)=\Phi^{\C}_j(x^{\C}_j)$. For each $i\in \Lambda^\C_j$, the almost-optimal waypoint $\xi^\C_j$ lies on the tangent of the zero level set $c_{ij}(\xi^\C_j,x_{ij})=0$, and thus satisfies the linear equation $(q-\xi^\C_j)^T\nabla_qc_{ij}(\xi^\C_j,x_{ij})=0$, where $\nabla_qc_{ij}(\xi^\C_j,x_{ij})$ is the normal vector of $c_{ij}(q,x_{ij})=0$ at $q=\xi^\C_j$. Since an $n$-dimensional solution can be uniquely determined by at most $n$ independent linear equations, there is a subset $\C'$ of $\Lambda^\C_j$ with $|\C'|\leq n$ such that $\xi^\C_j$ is the unique solution to the linear equations $(q-\xi^\C_j)^T\nabla_qc_{ij}(\xi^\C_j,x_{ij})=0$ for all $i\in \C'$. Moreover, due to the sublevel set representation of $\Omega^\C_j$, $\C'$ can be chosen in a way that $U\cap \Omega^\C_j=U\cap \Omega^{\C'}_j$ for some neighborhood $U$ of $\xi^\C_j$. Consequently, by the convexity of the SRS, we can conclude that $\Phi^{\C'}_j(x^{\C'}_j)=\Phi^\C_j(x^\C_j)$.
\end{Proof}

\subsection{Proof of Proposition \ref{prop:oet}}\label{proof:oet}

\begin{Proof}
Optimality: Note that the maximum possible value for the multi-attack objective function $\Gamma$ is equal to the number of active attackers $j$ for which the coalition-attacker pair $(\N,j)$ is feasible. In view of Algorithm \ref{alg:hilp}, an index $j$ is removed from $\M_r$ after the first iteration if and only if $(\N,j)$ is infeasible or attacker $j$ is included in the suboptimal solution $\tilde{\Theta}_1$. Hence, the emptiness of $\M_r$ after the first iteration indicates that $\tilde{\Theta}$ achieves the maximum possible value of $\Gamma$, rendering it an optimal solution to the ILP \eqref{ilp}. 
Efficiency: In each iteration of Algorithm \ref{alg:hilp}, there are two steps that require the computation of the convex program of the form \eqref{co:d}: determining the feasibility of the coalition-attacker pair $(\N_r,j)$, and identifying all irreducible sub-pairs of $(\Lambda^{\N_r}_j,j)$. The former step requires only one calculation, while the latter step, performed using Algorithm \ref{alg:irre}, demands at most $\sum_{k=1}^{n-1}\binom{n}{k}=2^{n}-2$ computations given that the cardinality of the ADS does not exceed $n$. Thus, the total number of computations at this iteration for one attacker is at most $2^n-1$. Moreover, the number of iterations is constrained by the number of active attackers $M_a$, since at least one attacker is removed from the set $\M_r$ after each iteration. The total number of calculations for all iterations is therefore at most $\sum_{k=1}^{M_a}(2^n-1)k$, which is less than $2^{n-1}M_a(1+M_a)$.
\end{Proof}

\bibliographystyle{IEEEtran}
\bibliography{ref}

\begin{thebibliography}{10}
\providecommand{\url}[1]{#1}
\csname url@samestyle\endcsname
\providecommand{\newblock}{\relax}
\providecommand{\bibinfo}[2]{#2}
\providecommand{\BIBentrySTDinterwordspacing}{\spaceskip=0pt\relax}
\providecommand{\BIBentryALTinterwordstretchfactor}{4}
\providecommand{\BIBentryALTinterwordspacing}{\spaceskip=\fontdimen2\font plus
\BIBentryALTinterwordstretchfactor\fontdimen3\font minus
  \fontdimen4\font\relax}
\providecommand{\BIBforeignlanguage}[2]{{%
\expandafter\ifx\csname l@#1\endcsname\relax
\typeout{** WARNING: IEEEtran.bst: No hyphenation pattern has been}%
\typeout{** loaded for the language `#1'. Using the pattern for}%
\typeout{** the default language instead.}%
\else
\language=\csname l@#1\endcsname
\fi
#2}}
\providecommand{\BIBdecl}{\relax}
\BIBdecl

\bibitem{vidal2002probabilistic}
R.~Vidal, O.~Shakernia, H.~J. Kim, D.~H. Shim, and S.~Sastry, ``Probabilistic
  pursuit-evasion games: theory, implementation, and experimental evaluation,''
  \emph{IEEE transactions on robotics and automation}, vol.~18, no.~5, pp.
  662--669, 2002.

\bibitem{bansal2017hamilton}
S.~Bansal, M.~Chen, S.~Herbert, and C.~J. Tomlin, ``Hamilton-jacobi
  reachability: A brief overview and recent advances,'' in \emph{2017 IEEE 56th
  Annual Conference on Decision and Control (CDC)}, 2017, pp. 2242--2253.

\bibitem{robin2016multi}
C.~Robin and S.~Lacroix, ``Multi-robot target detection and tracking: taxonomy
  and survey,'' \emph{Autonomous Robots}, vol.~40, pp. 729--760, 2016.

\bibitem{lowe2017multi}
R.~Lowe, Y.~I. Wu, A.~Tamar, J.~Harb, O.~Pieter~Abbeel, and I.~Mordatch,
  ``Multi-agent actor-critic for mixed cooperative-competitive environments,''
  \emph{Advances in neural information processing systems}, vol.~30, 2017.

\bibitem{rashid2020monotonic}
T.~Rashid, M.~Samvelyan, C.~S. De~Witt, G.~Farquhar, J.~Foerster, and
  S.~Whiteson, ``Monotonic value function factorisation for deep multi-agent
  reinforcement learning,'' \emph{The Journal of Machine Learning Research},
  vol.~21, no.~1, pp. 7234--7284, 2020.

\bibitem{zhang2021multi}
K.~Zhang, Z.~Yang, and T.~Ba{\c{s}}ar, ``Multi-agent reinforcement learning: A
  selective overview of theories and algorithms,'' \emph{Handbook of
  reinforcement learning and control}, pp. 321--384, 2021.

\bibitem{chung2011search}
T.~H. Chung, G.~A. Hollinger, and V.~Isler, ``Search and pursuit-evasion in
  mobile robotics,'' \emph{Autonomous robots}, vol.~31, no.~4, pp. 299--316,
  2011.

\bibitem{bakolas2012relay}
E.~Bakolas and P.~Tsiotras, ``Relay pursuit of a maneuvering target using
  dynamic voronoi diagrams,'' \emph{Automatica}, vol.~48, no.~9, pp.
  2213--2220, 2012.

\bibitem{alexopoulos2015cooperative}
A.~Alexopoulos, T.~Schmidt, and E.~Badreddin, ``Cooperative pursue in
  pursuit-evasion games with unmanned aerial vehicles,'' in \emph{2015 IEEE/RSJ
  International Conference on Intelligent Robots and Systems (IROS)}.\hskip 1em
  plus 0.5em minus 0.4em\relax IEEE, 2015, pp. 4538--4543.

\bibitem{de2021decentralized}
C.~De~Souza, R.~Newbury, A.~Cosgun, P.~Castillo, B.~Vidolov, and D.~Kuli{\'c},
  ``Decentralized multi-agent pursuit using deep reinforcement learning,''
  \emph{IEEE Robotics and Automation Letters}, vol.~6, no.~3, pp. 4552--4559,
  2021.

\bibitem{zhang2022game}
R.~Zhang, Q.~Zong, X.~Zhang, L.~Dou, and B.~Tian, ``Game of drones: Multi-uav
  pursuit-evasion game with online motion planning by deep reinforcement
  learning,'' \emph{IEEE Transactions on Neural Networks and Learning Systems},
  2022.

\bibitem{huang2011differential}
H.~Huang, J.~Ding, W.~Zhang, and C.~J. Tomlin, ``A differential game approach
  to planning in adversarial scenarios: A case study on capture-the-flag,'' in
  \emph{2011 IEEE International Conference on Robotics and Automation}.\hskip
  1em plus 0.5em minus 0.4em\relax IEEE, 2011, pp. 1451--1456.

\bibitem{garcia2018capture}
E.~Garcia, D.~W. Casbeer, and M.~Pachter, ``The capture-the-flag differential
  game,'' in \emph{2018 IEEE conference on decision and control (CDC)}.\hskip
  1em plus 0.5em minus 0.4em\relax IEEE, 2018, pp. 4167--4172.

\bibitem{jaderberg2019human}
M.~Jaderberg, W.~M. Czarnecki, I.~Dunning, L.~Marris, G.~Lever, A.~G.
  Castaneda, C.~Beattie, N.~C. Rabinowitz, A.~S. Morcos, A.~Ruderman
  \emph{et~al.}, ``Human-level performance in 3d multiplayer games with
  population-based reinforcement learning,'' \emph{Science}, vol. 364, no.
  6443, pp. 859--865, 2019.

\bibitem{shishika2018local}
D.~Shishika and V.~Kumar, ``Local-game decomposition for multiplayer
  perimeter-defense problem,'' in \emph{2018 IEEE conference on decision and
  control (CDC)}.\hskip 1em plus 0.5em minus 0.4em\relax IEEE, 2018, pp.
  2093--2100.

\bibitem{guerrero2020perimeter}
L.~Guerrero-Bonilla and D.~V. Dimarogonas, ``Perimeter surveillance based on
  set-invariance,'' \emph{IEEE Robotics and Automation Letters}, vol.~6, no.~1,
  pp. 9--16, 2020.

\bibitem{velhal2022decentralized}
S.~Velhal, S.~Sundaram, and N.~Sundararajan, ``A decentralized multirobot
  spatiotemporal multitask assignment approach for perimeter defense,''
  \emph{IEEE Transactions on Robotics}, 2022.

\bibitem{margellos2011hamilton}
K.~Margellos and J.~Lygeros, ``Hamilton--jacobi formulation for reach--avoid
  differential games,'' \emph{IEEE Transactions on automatic control}, vol.~56,
  no.~8, pp. 1849--1861, 2011.

\bibitem{fisac2015reach}
J.~F. Fisac, M.~Chen, C.~J. Tomlin, and S.~S. Sastry, ``Reach-avoid problems
  with time-varying dynamics, targets and constraints,'' in \emph{Proceedings
  of the 18th international conference on hybrid systems: computation and
  control}, 2015, pp. 11--20.

\bibitem{mitchell2005time}
I.~M. Mitchell, A.~M. Bayen, and C.~J. Tomlin, ``A time-dependent
  hamilton-jacobi formulation of reachable sets for continuous dynamic games,''
  \emph{IEEE Transactions on automatic control}, vol.~50, no.~7, pp. 947--957,
  2005.

\bibitem{bacsar1998dynamic}
T.~Ba{\c{s}}ar and G.~J. Olsder, \emph{Dynamic noncooperative game
  theory}.\hskip 1em plus 0.5em minus 0.4em\relax SIAM, 1998.

\bibitem{landry2018reach}
B.~Landry, M.~Chen, S.~Hemley, and M.~Pavone, ``Reach-avoid problems via
  sum-or-squares optimization and dynamic programming,'' in \emph{2018 IEEE/RSJ
  International Conference on Intelligent Robots and Systems (IROS)}.\hskip 1em
  plus 0.5em minus 0.4em\relax IEEE, 2018, pp. 4325--4332.

\bibitem{zhou2018efficient}
Z.~Zhou, J.~Ding, H.~Huang, R.~Takei, and C.~Tomlin, ``Efficient path planning
  algorithms in reach-avoid problems,'' \emph{Automatica}, vol.~89, pp. 28--36,
  2018.

\bibitem{garcia2020optimal}
E.~Garcia, D.~W. Casbeer, and M.~Pachter, ``Optimal strategies for a class of
  multi-player reach-avoid differential games in 3d space,'' \emph{IEEE
  Robotics and Automation Letters}, vol.~5, no.~3, pp. 4257--4264, 2020.

\bibitem{fu2023justification}
H.~Fu and H.~H.-T. Liu, ``Justification of the geometric solution of a target
  defense game with faster defenders and a convex target area using the hji
  equation,'' \emph{Automatica}, vol. 149, p. 110811, 2023.

\bibitem{pan2012pursuit}
S.~Pan, H.~Huang, J.~Ding, W.~Zhang, C.~J. Tomlin \emph{et~al.}, ``Pursuit,
  evasion and defense in the plane,'' in \emph{2012 American Control Conference
  (ACC)}.\hskip 1em plus 0.5em minus 0.4em\relax IEEE, 2012, pp. 4167--4173.

\bibitem{deng2020multi}
Z.~Deng and Z.~Kong, ``Multi-agent cooperative pursuit-defense strategy against
  one single attacker,'' \emph{IEEE Robotics and Automation Letters}, vol.~5,
  no.~4, pp. 5772--5778, 2020.

\bibitem{huang2011guaranteed}
H.~Huang, W.~Zhang, J.~Ding, D.~M. Stipanovi{\'c}, and C.~J. Tomlin,
  ``Guaranteed decentralized pursuit-evasion in the plane with multiple
  pursuers,'' in \emph{2011 50th IEEE Conference on Decision and Control and
  European Control Conference}.\hskip 1em plus 0.5em minus 0.4em\relax IEEE,
  2011, pp. 4835--4840.

\bibitem{pierson2016intercepting}
A.~Pierson, Z.~Wang, and M.~Schwager, ``Intercepting rogue robots: An algorithm
  for capturing multiple evaders with multiple pursuers,'' \emph{IEEE Robotics
  and Automation Letters}, vol.~2, no.~2, pp. 530--537, 2016.

\bibitem{yan2018reach}
R.~Yan, Z.~Shi, and Y.~Zhong, ``Reach-avoid games with two defenders and one
  attacker: An analytical approach,'' \emph{IEEE transactions on cybernetics},
  vol.~49, no.~3, pp. 1035--1046, 2018.

\bibitem{yan2019task}
------, ``Task assignment for multiplayer reach--avoid games in convex domains
  via analytical barriers,'' \emph{IEEE Transactions on Robotics}, vol.~36,
  no.~1, pp. 107--124, 2019.

\bibitem{yan2020guarding}
------, ``Guarding a subspace in high-dimensional space with two defenders and
  one attacker,'' \emph{IEEE Transactions on Cybernetics}, vol.~52, no.~5, pp.
  3998--4011, 2020.

\bibitem{yan2022matching}
R.~Yan, X.~Duan, Z.~Shi, Y.~Zhong, and F.~Bullo, ``Matching-based capture
  strategies for 3d heterogeneous multiplayer reach-avoid differential games,''
  \emph{Automatica}, vol. 140, p. 110207, 2022.

\bibitem{isaacs1999differential}
R.~Isaacs, \emph{Differential games: a mathematical theory with applications to
  warfare and pursuit, control and optimization}.\hskip 1em plus 0.5em minus
  0.4em\relax Courier Corporation, 1999.

\bibitem{zhou2016cooperative}
Z.~Zhou, W.~Zhang, J.~Ding, H.~Huang, D.~M. Stipanovi{\'c}, and C.~J. Tomlin,
  ``Cooperative pursuit with voronoi partitions,'' \emph{Automatica}, vol.~72,
  pp. 64--72, 2016.

\bibitem{gerkey2004formal}
B.~P. Gerkey and M.~J. Matari{\'c}, ``A formal analysis and taxonomy of task
  allocation in multi-robot systems,'' \emph{The International journal of
  robotics research}, vol.~23, no.~9, pp. 939--954, 2004.

\bibitem{korsah2013comprehensive}
G.~A. Korsah, A.~Stentz, and M.~B. Dias, ``A comprehensive taxonomy for
  multi-robot task allocation,'' \emph{The International Journal of Robotics
  Research}, vol.~32, no.~12, pp. 1495--1512, 2013.

\bibitem{chen2016multiplayer}
M.~Chen, Z.~Zhou, and C.~J. Tomlin, ``Multiplayer reach-avoid games via
  pairwise outcomes,'' \emph{IEEE Transactions on Automatic Control}, vol.~62,
  no.~3, pp. 1451--1457, 2017.

\bibitem{ford1956maximal}
L.~R. Ford and D.~R. Fulkerson, ``Maximal flow through a network,''
  \emph{Canadian journal of Mathematics}, vol.~8, pp. 399--404, 1956.

\bibitem{hopcroft1973n}
J.~E. Hopcroft and R.~M. Karp, ``An n\^{}5/2 algorithm for maximum matchings in
  bipartite graphs,'' \emph{SIAM Journal on computing}, vol.~2, no.~4, pp.
  225--231, 1973.

\bibitem{garcia2020multiple}
E.~Garcia, D.~W. Casbeer, A.~Von~Moll, and M.~Pachter, ``Multiple pursuer
  multiple evader differential games,'' \emph{IEEE Transactions on Automatic
  Control}, vol.~66, no.~5, pp. 2345--2350, 2020.

\bibitem{mitchell2007toolbox}
I.~M. Mitchell \emph{et~al.}, ``A toolbox of level set methods,'' \emph{UBC
  Department of Computer Science Technical Report TR-2007-11}, p.~31, 2007.

\bibitem{blanchini1999set}
F.~Blanchini, ``Set invariance in control,'' \emph{Automatica}, vol.~35,
  no.~11, pp. 1747--1767, 1999.

\bibitem{nocedal2006numerical}
J.~Nocedal and S.~Wright, \emph{Numerical optimization}.\hskip 1em plus 0.5em
  minus 0.4em\relax Springer Science \& Business Media, 2006.

\bibitem{boyd2004convex}
S.~Boyd, S.~P. Boyd, and L.~Vandenberghe, \emph{Convex optimization}.\hskip 1em
  plus 0.5em minus 0.4em\relax Cambridge university press, 2004.

\bibitem{diamond2016cvxpy}
S.~Diamond and S.~Boyd, ``Cvxpy: A python-embedded modeling language for convex
  optimization,'' \emph{The Journal of Machine Learning Research}, vol.~17,
  no.~1, pp. 2909--2913, 2016.

\bibitem{koenig2004design}
N.~Koenig and A.~Howard, ``Design and use paradigms for gazebo, an open-source
  multi-robot simulator,'' in \emph{2004 IEEE/RSJ International Conference on
  Intelligent Robots and Systems (IROS)(IEEE Cat. No. 04CH37566)},
  vol.~3.\hskip 1em plus 0.5em minus 0.4em\relax IEEE, 2004, pp. 2149--2154.

\bibitem{wang2017safety}
L.~Wang, A.~D. Ames, and M.~Egerstedt, ``Safety barrier certificates for
  collisions-free multirobot systems,'' \emph{IEEE Transactions on Robotics},
  vol.~33, no.~3, pp. 661--674, 2017.

\bibitem{lee2015multirobot}
S.~G. Lee, Y.~Diaz-Mercado, and M.~Egerstedt, ``Multirobot control using
  time-varying density functions,'' \emph{IEEE Transactions on robotics},
  vol.~31, no.~2, pp. 489--493, 2015.

\bibitem{royden1988real}
H.~L. Royden and P.~Fitzpatrick, \emph{Real analysis}.\hskip 1em plus 0.5em
  minus 0.4em\relax Macmillan New York, 1988, vol.~32.

\end{thebibliography}

\end{document}